%% file: main.tex
\documentclass{article}

\usepackage[accepted]{icml2022}

\input{preamble}
\input{commands}

\begin{document}

\twocolumn[
\icmltitle{Model-Value Inconsistency as a Signal for Epistemic Uncertainty}

\icmlsetsymbol{equal}{*}

\begin{icmlauthorlist}
\icmlauthor{Angelos Filos}{equal,dm,ox}
\icmlauthor{Eszter V\'ertes}{equal,dm}
\icmlauthor{Zita Marinho}{equal,dm}
\icmlauthor{Gregory Farquhar}{dm}
\icmlauthor{Diana Borsa}{dm}
\icmlauthor{Abram Friesen}{dm}
\icmlauthor{Feryal Behbahani}{dm}
\icmlauthor{Tom Schaul}{dm}
\icmlauthor{Andr\'e Barreto}{dm}
\icmlauthor{Simon Osindero}{dm}
\end{icmlauthorlist}

\icmlaffiliation{dm}{DeepMind}
\icmlaffiliation{ox}{University of Oxford}

\icmlcorrespondingauthor{Angelos Filos}{angelos.filos@cs.ox.ac.uk}

\icmlkeywords{
  Machine Learning,
  Reinforcement Learning,
  Planning,
  Epistemic Uncertainty,
}

\vskip 0.3in
]

\printAffiliationsAndNotice{\icmlEqualContribution}

\input{sections/0_abstract}
\input{sections/1_introduction}
\input{sections/2_background}
\input{sections/3_method}
\input{sections/4_experiments}
\input{sections/5_related_work}
\input{sections/6_discussion}

\bibliography{references}
\bibliographystyle{icml2022}

\clearpage
\appendix
\onecolumn

\input{sections/A_experimental_details}
\input{sections/B_implementation_details}
\input{sections/C_extensions}

\input{sections/D_ablations}
\clearpage
\input{sections/E_muesli_and_ive}

\end{document}

%% file: preamble.tex
\usepackage{titlesec}
\titlespacing*{\paragraph}{0pt}{0.5em}{0.5em}

\usepackage{microtype}
\usepackage{graphicx}
\usepackage{booktabs} 
\usepackage{caption}
\usepackage{subcaption}
\usepackage{makecell}
\usepackage{multirow}

\usepackage{amsmath}
\usepackage{amssymb}
\usepackage{amsfonts}
\usepackage{amsthm}
\usepackage{mathtools}

\usepackage{color}
\usepackage{xcolor,colortbl}
\usepackage{ml-colors}
\definecolor{ourmethod}{gray}{0.93}
\newcommand{\bftab}{\fontseries{b}\selectfont}

\usepackage[framemethod=TikZ]{mdframed}
\mdfsetup{
  backgroundcolor=black!10,
  roundcorner=10pt,
  skipabove=0.5em,
  skipbelow=0.5em
}
\usepackage[most]{tcolorbox}

\usepackage{tikz}
\usetikzlibrary{bayesnet}
\usetikzlibrary{arrows}
\usetikzlibrary{automata}
\usetikzlibrary{matrix}
\usetikzlibrary{calc}
\usetikzlibrary{positioning}
\usetikzlibrary{graphs}
\usetikzlibrary{shapes.geometric}
\usetikzlibrary{backgrounds}

\usepackage{enumitem}

\usepackage{alphalph}
\usepackage{pgffor}

\newtheoremstyle{thm}
  {2pt} 
  {2pt} 
  {\itshape} 
  {} 
  {\bfseries} 
  {.} 
  {.5em} 
  {} 
\theoremstyle{thm}

\usepackage{thmtools} 
\usepackage{thm-restate}

\newtheorem{definition}{Definition}

\usepackage[colorlinks]{hyperref}
\definecolor{mydarkblue}{rgb}{0,0.08,0.45}
\colorlet{hypercolor}{mydarkblue}
\hypersetup{%
  colorlinks=true,
  linkcolor=hypercolor,
  citecolor=hypercolor,
  filecolor=hypercolor,
  urlcolor=hypercolor}
\usepackage{textcomp}

\usepackage{etoolbox}

\usepackage[normalem]{ulem}

%% file: commands.tex
\newcommand{\note}[1]{\begin{mdframed}#1\end{mdframed}}

\DeclareMathOperator*{\argmax}{arg\,max}
\DeclareMathOperator*{\argmin}{arg\,min}
\newcommand{\underdescribe}[3][0pt]{\hspace*{.12em}\underbracket[0.5pt][1pt]{#2\hspace*{#1}}_{#3}}

\newcommand{\E}[0]{\mathbb{E}}

\newcommand{\B}[0]{\mathcal{B}}

\newcommand{\T}[0]{\mathcal{T}}
\newcommand{\mstar}[0]{m^{*}}
\newcommand{\mhat}[0]{\hat{m}}
\newcommand{\vhat}[0]{\hat{v}}
\newcommand{\rhat}[0]{\hat{r}}
\newcommand{\phat}[0]{\hat{p}}
\newcommand{\hhat}[0]{\hat{h}}
\newcommand{\pihat}[0]{\hat{\pi}}

\newcommand{\procgen}[0]{\texttt{procgen}}
\newcommand{\minatar}[0]{\texttt{minatar}}
\newcommand{\gridworld}[0]{\texttt{gridworld}}
\newcommand{\walker}[0]{\texttt{walker}}
\newcommand{\levels}[0]{\texttt{\#levels}}


%% file: sections/0_abstract.tex
\begin{abstract}
  Using a model of the environment and a value function, an agent can construct many estimates of a state’s value, by unrolling the model for different lengths and bootstrapping with its value function.
  Our key insight is that one can treat this set of value estimates as a type of ensemble, which we call an \emph{implicit value ensemble} (IVE).
  Consequently, the discrepancy between these estimates can be used as a proxy for the agent’s epistemic uncertainty; we term this signal \emph{model-value inconsistency} or \emph{self-inconsistency} for short.
  Unlike prior work which estimates uncertainty by training an ensemble of many models and/or value functions, this approach requires only the single model and value function which are already being learned in most model-based reinforcement learning algorithms.
  We provide empirical evidence in both tabular and function approximation settings from pixels that self-inconsistency is useful (i) as a signal for exploration, (ii) for acting safely under distribution shifts, and (iii) for robustifying value-based planning with a learned model.
\end{abstract}

%% file: sections/1_introduction.tex
\section{Introduction}
\label{sec:introduction}

\input{figures/ive-cg-gp}
\input{figures/nets}

Agents that employ learning to improve their decision making should be equipped with mechanisms for representing and using their acquired knowledge effectively.
Learned models of the environment~\citep{sutton1991dyna} and value functions~\citep{sutton1988learning} are explicit ways that reinforcement learning~\citep[RL,][]{sutton2018reinforcement} agents use to represent their knowledge about the environment.
\looseness=-1

Equally important is the agents' ability to reason about their \emph{ignorance}~\citep[i.e., epistemic uncertainty,][]{strens2000bayesian} and factor it in their decisions~\citep{milnor1951games}.
In tabular settings, exact Bayesian inference can be used for quantifying the agents' uncertainty in both model-free~\citep{dearden1998bayesian} and model-based~\citep{dearden2013model} RL approaches.
However, in complex RL problems, since exact Bayesian inference is intractable, proxy signals are often used instead, including prediction error~\citep{lopes2012exploration,pathak2017curiosity}, approximate state visitation counts~\citep{bellemare2016unifying} and disagreement of samples from either approximate posterior distributions over learned parameters~\citep{blundell2015weight} or explicit ensembles of value functions~\citep{osband2016deep} or world models~\citep{chua2018deep}.
\looseness=-1

In this work, we introduce a novel signal for capturing RL agents' ignorance, termed \emph{model-value inconsistency} or \emph{self-inconsistency} for short.
A $k$-step self-inconsistency signal is constructed by applying the model-induced Bellman operator $\mathcal{T}_{\mhat}$ to learned value function $\vhat$, $k$ times.
This produces $k+1$ different estimates of the state value: $\{ \vhat, \mathcal{T}_{\mhat} \vhat, (\mathcal{T}_{\mhat})^{2} \vhat, \ldots, (\mathcal{T}_{\mhat})^{k} \vhat \}$, as illustrated in Figures~\ref{fig:ive-cg-gp} and~\ref{subfig:ive-net}.
Our key insight is that these can be thought of as predictions from an ensemble of value functions, which we call the \emph{implicit value ensemble} (IVE).

Consequently, the disagreement of these predictions can tell us about the agent’s uncertainty in the value of a state.
The intuition behind this is based on the fact that the true model and value are by definition Bellman-consistent.
As a result, for regions of the state space where the learned model and value are accurate, we expect the self-inconsistency to be low.
Conversely,  high self-inconsistency can signal that the learned model-value pair is inaccurate.
\looseness=-1

In contrast to prior work that requires explicit ensembles of learned value functions~\citep{osband2016deep,lowrey2018plan} or ensembles of world models~\citep{chua2018deep,sekar2020planning}, self-inconsistency can be efficiently calculated by \emph{any} RL agent that has a learned (approximate) model of the environment and value function, see Figure~\ref{fig:nets}.
Moreover, unlike model-ensembles, self-inconsistency captures the agents' ignorance about behaviourally-relevant quantities, i.e., rewards and values, and hence is robust to irrelevant information for control noise~\citep{schmidhuber2010formal}.

We provide empirical evidence that self-inconsistency provides a proxy of epistemic uncertainty (Section~\ref{subsec:detecting-out-of-distribution-regimes-with-self-inconsistency}), and that this information can be used to guide exploration or act safely  (Section~\ref{subsec:self-inconsistency-as-a-signal-for-exploration}), and to robustify planning (Section~\ref{subsec:planning-with-ensembled-values}).

%% file: figures/ive-cg-gp.tex
\begin{figure}[t]
  \centering
  \begin{subfigure}[b]{0.6\linewidth}
    \centering
    \resizebox{\linewidth}{!}{\input{assets/tikz/ive-cg}}
    \caption{Implicit Value Ensemble (IVE)}
    \label{subfig:ive-cg}
  \end{subfigure}
  ~
  \begin{subfigure}[b]{0.37\linewidth}
    \centering
    \begin{subfigure}[b]{\linewidth}
      \centering
      \includegraphics[width=\linewidth]{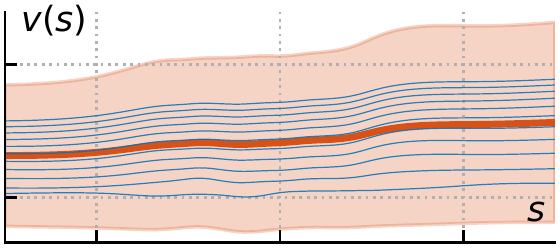}
      \caption{At initialisation}
      \label{subfig:ive-prior}
    \end{subfigure}
    \\[0.5em]
    \begin{subfigure}[b]{\linewidth}
      \centering
      \includegraphics[width=\linewidth]{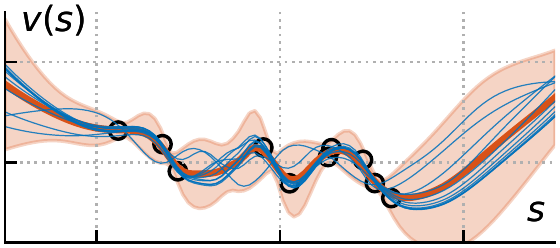}
      \caption{After training}
      \label{subfig:ive-posterior}
    \end{subfigure}
  \end{subfigure}
  \caption{
    \emph{Implicit value ensemble} (IVE) estimated from a single learned model $\mhat$ and value function $\vhat$.\
    (a) Computation graph. The model-induced Bellman operator $\T_{\mhat}$ is repeatedly applied $k$ times on the approximate value function $\vhat$, i.e., $\vhat_{\mhat}^{k}(s) \triangleq (\T_{\mhat})^{k} \vhat (s)$.
    (b-c) Didactic example with 1D state space:
    Value predictions (in \textcolor{mlblue}{blue}) for different values of $k$, i.e., $\{ \vhat_{\mhat}^{k} \}_{k=0}^{10}$, along with the ensemble mean $\mu$-IVE(10) and standard deviation $\sigma$-IVE(10) (in \textcolor{mlorange}{orange}), before (b) and after (c) training with value targets (black circles).
    The ensemble standard deviation is non-trivial at out-of-distribution (OOD) states and zero at in-distribution states.
    \looseness=-1
  }
  \label{fig:ive-cg-gp}
  \vspace{-1.5em}
\end{figure}

%% file: assets/tikz/ive-cg.tex
\begin{tikzpicture}[every node/.style={inner sep=0,outer sep=0, minimum size=2.5em}]
  \node[draw=none] (s0) {$s$};
  \node[right = 0.75em of s0, draw, circle, fill=mlyellow!35] (vhat0) {$\hat{v}$};
  \node[right = 0.75em of vhat0] (v0) {$\vhat_{\mhat}^{0}(s)$};
  \path (s0) edge [>=latex, ->] (vhat0);
  \path (vhat0) edge [>=latex, ->] (v0);
  \node[below = 1.35em of s0, draw=none] (s1) {$s$};
  \node[right = 0.75em of s1, draw, rectangle, fill=mlblue!35] (That1) {$\T_{\mhat}$};
  \node[right = 0.75em of That1, draw, circle, fill=mlyellow!35] (vhat1) {$\hat{v}$};
  \node[right = 0.75em of vhat1] (v1) {$\vhat_{\mhat}^{1}(s)$};
  \path (s1) edge [>=latex, ->] (That1);
  \path (That1) edge [>=latex, ->] (vhat1);
  \path (vhat1) edge [>=latex, ->] (v1);
  \node[below = 1.35em of s1, draw=none] (s2) {$s$};
  \node[right = 0.75em of s2, draw, rectangle, fill=mlblue!35] (That21) {$\T_{\mhat}$};
  \node[right = 0.75em of That21, draw, rectangle, fill=mlblue!35] (That22) {$\T_{\mhat}$};
  \node[right = 0.75em of That22, draw, circle, fill=mlyellow!35] (vhat2) {$\hat{v}$};
  \node[right = 0.75em of vhat2] (v2) {$\vhat_{\mhat}^{2}(s)$};
  \path (s2) edge [>=latex, ->] (That21);
  \path (That21) edge [>=latex, ->] (That22);
  \path (That22) edge [>=latex, ->] (vhat2);
  \path (vhat2) edge [>=latex, ->] (v2);
  \node[below = 1.35em of s2, draw=none] (sk) {$s$};
  \node[right = 0.75em of sk, draw, rectangle, fill=mlblue!35] (Thatk1) {$\T_{\mhat}$};
  \node[right = 0.75em of Thatk1] (Thatkd) {$\cdots$};
  \node[right = 0.75em of Thatkd, draw, rectangle, fill=mlblue!35] (Thatkk) {$\T_{\mhat}$};
  \node[right = 0.75em of Thatkk, draw, circle, fill=mlyellow!35] (vhatk) {$\hat{v}$};
  \node[right = 0.75em of vhatk] (vk) {$\vhat_{\mhat}^{k}(s)$};
  \path (sk) edge [>=latex, ->] (Thatk1);
  \path (Thatk1) edge [>=latex, ->] (Thatkd);
  \path (Thatkd) edge [>=latex, ->] (Thatkk);
  \path (Thatkk) edge [>=latex, ->] (vhatk);
  \path (vhatk) edge [>=latex, ->] (vk);
  
\end{tikzpicture}

%% file: figures/nets.tex
\begin{figure*}[t]
  \centering
  \begin{subfigure}[b]{0.25\linewidth}
    \centering
    \resizebox{!}{4cm}{\input{assets/tikz/vensemble-net}}
    \caption{Value Ensemble (EVE)}
    \label{subfig:vensemble-net}
  \end{subfigure}
  ~
  \begin{subfigure}[b]{0.31\linewidth}
    \centering
    \resizebox{!}{4cm}{\input{assets/tikz/mensemble-net}}
    \caption{Model Ensemble (EMVE)}
    \label{subfig:mensemble-net}
  \end{subfigure}
  ~
  \begin{subfigure}[b]{0.39\linewidth}
    \centering
    \resizebox{!}{4cm}{\input{assets/tikz/ive-net}}
    \caption{Implicit Value Ensemble (IVE)}
    \label{subfig:ive-net}
  \end{subfigure}
  \caption{
    Value computation in scalable epistemic uncertainty-aware RL agents.
    (a-b) Explicit ensemble of value functions~\citep{osband2016deep} and world models~\citep{chua2018deep}, approximating samples from $p(v | \B)$ and $p(m | \B)$, respectively.
    The number of parameters grows linearly with the ensemble size.
    (c) Implicit value ensemble (IVE) make ensemble value predictions using a single learned value function and world model by exploiting the model-induced Bellman operator $\T_{\mhat}$ and the Bellman consistency of the ``true'' model $\mstar$ and value function $v$, keeping the number of parameters constant.
    \looseness=-1
  }
  \label{fig:nets}
\end{figure*}
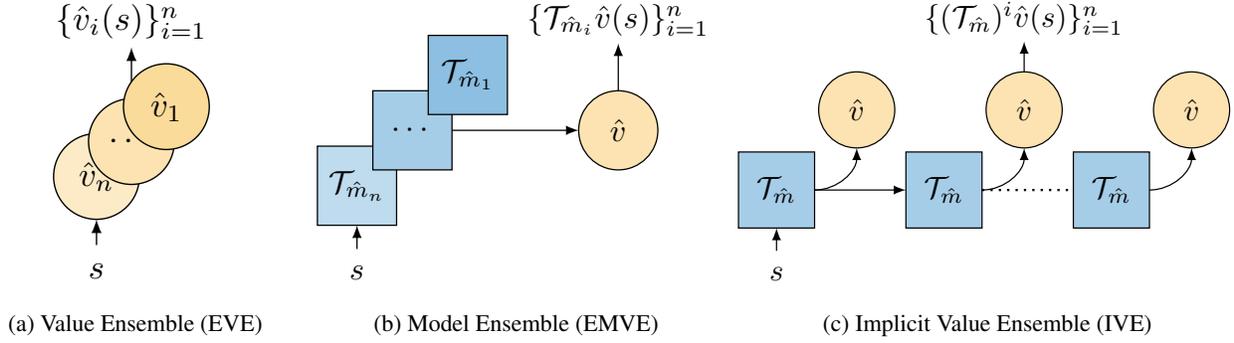

%% file: assets/tikz/vensemble-net.tex
\begin{tikzpicture}[every node/.style={inner sep=0,outer sep=0, minimum size=2.5em}]
  \node[draw=none, minimum size=1.5em] (s) {$s$};
  \node[above = 0.75em of s, draw, circle, fill=mlyellow!25] (vhat3) {$\vhat_{n}$};
  \node[above right = -0.75em and -0.75em of vhat3, draw, circle, fill=mlyellow!35] (vhat2) {$\cdots$};
  \node[above = 1.5em of vhat2, minimum size=1.5em] (v) {$\{\vhat_{i}(s)\}_{i=1}^{n}$};
  \path (s) edge [>=latex, ->] (vhat3);
  \path (vhat2) edge [>=latex, ->] (v);
  \node[above right = -0.75em and -0.75em of vhat2, draw, circle, fill=mlyellow!45] (vhat1) {$\vhat_{1}$};
\end{tikzpicture}

%% file: assets/tikz/mensemble-net.tex
\begin{tikzpicture}[every node/.style={inner sep=0,outer sep=0, minimum size=2.5em}]
  \node[draw=none, minimum size=1.5em] (s) {$s$};
  \node[above = 0.75em of s, draw, rectangle, fill=mlblue!25] (mhat3) {$\T_{\mhat_{n}}$};
  \node[above right = -0.75em and -0.75em of mhat3, draw, rectangle, fill=mlblue!35] (mhat2) {$\cdots$};
  \node[right = 4em of mhat2, draw, circle, fill=mlyellow!35] (vhat) {$\vhat$};
  \node[above = 1.5em of vhat, minimum size=1.5em] (v) {$\{\T_{\mhat_{i}} \vhat (s)\}_{i=1}^{n}$};
  \path (s) edge [>=latex, ->] (mhat3);
  \path (mhat2) edge [>=latex, ->] (vhat);
  \path (vhat) edge [>=latex, ->] (v);
  \node[above right = -0.75em and -0.75em of mhat2, draw, rectangle, fill=mlblue!45] (mhat1) {$\T_{\mhat_{1}}$};
\end{tikzpicture}

%% file: assets/tikz/ive-net.tex
\begin{tikzpicture}[every node/.style={inner sep=0,outer sep=0, minimum size=2.5em}]
  \node[draw=none, minimum size=1.5em] (s) {$s$};
  \node[above = 0.75em of s, draw, rectangle, fill=mlblue!35] (mhat3) {$\T_{\mhat}$};
  \node[above right = 0.5em and 0.5em of mhat3, draw, circle, fill=mlyellow!35] (vhat3) {$\vhat$};
  \node[right = 3em of mhat3, draw, rectangle, fill=mlblue!35] (mhat2) {$\T_{\mhat}$};
  \node[above right = 0.5em and 0.5em of mhat2, draw, circle, fill=mlyellow!35] (vhat2) {$\vhat$};
  \node[right = 3em of mhat2, draw, rectangle, fill=mlblue!35] (mhat1) {$\T_{\mhat}$};
  \node[above right = 0.5em and 0.5em of mhat1, draw, circle, fill=mlyellow!35] (vhat1) {$\vhat$};
  \node[above = 1.0em of vhat2, minimum size=1.5em] (v) {$\{(\T_{\mhat})^{i} \vhat (s)\}_{i=1}^{n}$};
  \path (s) edge [>=latex, ->] (mhat3);
  \path (mhat3) edge [>=latex, ->] (mhat2);
  \path (mhat3) edge [>=latex, ->, bend right = 45] (vhat3);
  \path (mhat2) edge [>=latex, dotted, line width=0.075em] (mhat1);
  \path (mhat2) edge [>=latex, ->, bend right = 45] (vhat2);
  \path (mhat1) edge [>=latex, ->, bend right = 45] (vhat1);
  \path (vhat2) edge [>=latex, ->] (v);
\end{tikzpicture}

%% file: sections/2_background.tex
\section{Background}
\label{sec:background}

We model the agent's interaction with the environment as a Markov decision process~\citep[MDP,][]{puterman2014markov}, i.e., $\mathcal{M} \triangleq \left( \mathcal{S}, \mathcal{A}, p, r \right)$.
At any discrete time step $t \geq 0$, the agent is in state $s_{t} \in \mathcal{S}$, takes an action $a_{t} \in \mathcal{A}$, according to a policy $\pi: \mathcal{S} \rightarrow \Delta(\mathcal{A})$, then receives reward $R_{t+1} \sim r(\cdot | s_t, a_{t}) \in \mathbb{R}$ and transitions to the state $S_{t+1} \sim p(\cdot | s_{t}, a_{t})$.
For brevity, the ``true'' model is denoted by $\mstar \triangleq (p, r)$ and we write $S_{t+1}, R_{t+1} \sim m^{*}(\cdot, \cdot | s_{t}, a_{t})$.
The agent's goal is to find the policy that maximises the \emph{value} of each state, for a discount factor $\gamma \in [0, 1)$, $v^{\pi}(s) \triangleq \E_{\pi, \mstar} [ \sum_{t \geq 0} \gamma^{t} R_{t} | S_{0} = s ]$,
where $\E_{\pi, \mstar} \left[ \cdot \right]$ denotes the expectation\footnote{In this work, we only construct estimates of the \emph{mean} of the returns distribution~\citep[a.k.a value distribution][]{bellemare2017distributional} and hence environment and policy stochasticity is integrated out.} over the trajectories induced by running policy $\pi$ in the environment $\mstar$, starting from state $s$.
\looseness=-1

The computation of the value of a policy $\pi$, i.e., $v^{\pi}$, is termed \emph{policy evaluation} and can be concisely formulated using Bellman evaluation operators~\citep{bellman1957markovian}.
Next, we define the \emph{one-step} Bellman evaluation operator, applied on a state-(to-scalar) function $v \in \mathbb{V} \triangleq \{ f: \mathcal{S} \rightarrow \mathbb{R} \}$.

\begin{definition}[Bellman evaluation operator]
  Given the model $\mstar$ and policy $\pi$ the one-step Bellman evaluation operator $\T^{\pi}: 
  \mathbb{V} \rightarrow \mathbb{V}$ is induced, and its application on a state-function $v \in \mathbb{V}$, for all $s \in \mathcal{S}$, is given by
  \begin{align}
    \T^{\pi} v(s) \triangleq \E_{\pi, \mstar} \left[ R_{1} + \gamma v(S_{1}) \mid S_{0}=s \right].
  \label{eq:bellman-evaluation-one-step}
  \end{align}
\end{definition}

The $k$-times repeated application of an one-step Bellman operator gives rise to the $k$-steps Bellman operator,
\begin{align}
  (\T^{\pi})^{k} v \triangleq \underdescribe{\T^{\pi} \cdots \T^{\pi}}{k\text{-times}} v .
\label{eq:bellman-evaluation-k-steps}
\end{align}
The Bellman evaluation operator, $\T^{\pi}$, is a contraction mapping~\citep{puterman2014markov}, and its fixed point is the value of the policy $\pi$, i.e., $\lim_{n \rightarrow \infty} (\T^{\pi})^{n} v = v^{\pi}$, for any $v \in \mathbb{V}$.

\subsection{Model-Based Reinforcement Learning}
\label{subsec:model-based-reinforcement-learning}

In the general RL formulation, it is assumed that the environment model $\mstar$ is unknown to the agent~\citep{sutton2018reinforcement} which thus cannot directly compute Eqn.~(\ref{eq:bellman-evaluation-one-step}).
\emph{Model-free} RL agents resolve this by estimating these expectations through sampling.
\emph{Model-based} RL agents, the focus of this paper, learn an approximate model $\mhat \approx \mstar$, possibly together with a learned value function $\vhat \approx v^{\pi}$~\citep{sutton1991dyna}, and use them to compute an estimate of the value, by replacing model and function $\mstar$, $v$ with $\mhat$, $\vhat$ in Eqn.~(\ref{eq:bellman-evaluation-one-step}).

\paragraph{Model-induced Bellman operator.}
A model $\mhat$ and policy $\pi$ induce a Bellman evaluation operator $\T_{\mhat}^{\pi}$ with a fixed point $v_{\mhat}^{\pi}$.
Similar to Eqn.~(\ref{eq:bellman-evaluation-k-steps}), a $k$-steps model-induced Bellman operator is given by $(\T_{\mhat}^{\pi})^{k} v = \underdescribe{\T_{\mhat}^{\pi}\cdots\T_{\mhat}^{\pi}}{k\text{-times}} v$.


\paragraph{Model learning principles.}
The agent interacts with the environment, generating a sequence of states, actions and rewards, which we denote with $\B \triangleq \{(s_{t}, a_{t}, r_{t}) \}_{t \geq 0}$.

Maximum likelihood estimation~\citep[MLE,][]{kumar2015stochastic,sutton1991dyna} can be used for learning the model parameters, given experience tuples $(s, a, r', s') \sim \B$,
\begin{align}
  \mhat_{\text{MLE}} = \argmax_{m} \E_{\B} \left[ \log m(r', s' \mid s, a) \right].
\label{eq:theta-mle}
\end{align}
Action-conditioned hidden Markov models have been used to scale MLE methods to high-dimensional environments~\citep{watter2015embed}, e.g., with pixel observations.
\looseness=-1

Value equivalence~\citep[VE,][]{grimm2021proper} is an alternative principle for model learning.
It selects the model that induces the ``best'' approximation to the $k$-th order Bellman operator of the environment, applied on state-functions $\mathcal{V}$, policies $\Pi$ and state $s$, trained via samples $(s, a, r', s') \sim \B$,
\looseness=-1
\begin{align}
  \mhat_{\text{VE}} = \argmin_{m} \E_{\B}
  \!\!\!\!\!\sum_{\pi \in \Pi, v \in \mathcal{V}}\!\!\!\!\!
  \left| (\T_{m}^{\pi})^{k} v(s) - (\T^{\pi})^{k} v(s) \right|.
\label{eq:theta-ve}
\end{align}

\subsection{Epistemic-Uncertainty-Aware Agents}
\label{subsec:epistemic-uncertainty-aware-agents}

We refer to learning agents that can quantify their uncertainty about their learned components, e.g., value function or model, as \emph{epistemic uncertainty-aware} (a.k.a. ignorance-aware) agents.
While \emph{aleatoric uncertainty} captures the inherent and irreducible stochasticity of the agents' environment, epistemic uncertainty is agent-centric~\citep[i.e., \emph{subjective},][]{savage1972foundations} and reducible~\citep{hutter2004universal}.

\paragraph{Bayesian agents.}
A principled approach to quantifying epistemic uncertainty is by treating learned quantities as random variables and perform Bayesian inference given the observed data.
Bayesian RL agents maintain beliefs over value functions~\citep{dearden1998bayesian} or world models~\citep{dearden2013model}, which are updated upon interactions with the environment.
Exact inference is intractable for most interesting problems and thus ensemble-based approximations are used instead~\citep{lu2021reinforcement}.

\paragraph{Explicit ensemble methods.}
In deep RL, neural networks (NNs) are used to approximate the value function~\citep{mnih2013playing} or the model~\citep{watter2015embed}.
A popular approach to epistemic uncertainty quantification for NNs is \emph{deep ensembles}~\citep{lakshminarayanan2016simple}.
Under certain assumptions~\citep{pearce2020uncertainty}, the ensemble components can be seen as samples from the posterior distribution over NN parameters.
It has been argued that the diversity (i.e., de-correlation) of the ensemble components is important for better capturing epistemic uncertainty~\citep{wilson2020bayesian} and various methods have been used to achieve this, all of which inject noise into the learning algorithm, such as: (i) data bootstrapping~\citep{tibshirani1996comparison,osband2016deep}; (ii) different loss function (iii) function form~\citep{wenzel2020hyperparameter} or (iv) structured noise per ensemble component~\citep[e.g., priors,][]{osband2018randomized}.

RL agents with an ensemble of value functions or models have been used to quantify their epistemic uncertainty e.g. \citep{osband2016deep, kurutach2018model}, see Figure~\ref{subfig:vensemble-net} and~\ref{subfig:mensemble-net}, respectively.
We call these methods \emph{explicit} ensemble methods and their number of parameters grows linearly with the ensemble size.
In contrast, \emph{implicit} ensembles escape this linear scaling by sharing parameters between the ensemble members but without sacrificing diversity.

%% file: sections/3_method.tex
\section{Your Model-Based Agent is Secretly an Ensemble of Value Functions}
\label{sec:your-model-based-agent-is-secretly-an-ensemble-of-value-functions}

We now present a proxy signal for epistemic uncertainty, computable by any model-based RL agent with a single (point) estimate of a world model and a value function\footnote{In this section, we define everything in terms of the Bellman evaluation operator and an approximate on-policy value function. 
The Bellman \emph{optimality} operator and an approximate optimal value function could be used instead. For completeness, see Appendix~\ref{app:extensions}.}.

\input{figures/std-value-ensembles}

\subsection{Implicit Value Ensemble}
\label{subsec:implicit-value-esemble}

A key component of our method is the value estimated by a $k$-step application of the model-induced Bellman operator on the learned value function, which we call \emph{$k$-steps model-predicted value}\footnote{Similar quantities have been used in prior work, e.g., $k$-preturn~\citep{silver2017predictron} and MVE~\citep{feinberg2018model}. We discuss them and their differences in more detail in Section~\ref{sec:related-work}.} ($k$-MPV), given by
\begin{align}
  \vhat_{\mhat}^{k} \triangleq (\T_{\mhat}^{\pi})^{k} \vhat.
\label{eq:k-mpv}
\end{align}

\note{%
The $k$-MPV is a value estimator that interpolates between (i) a model-free value estimator, i.e., $k=0$ and (ii) a purely model-based value estimator, i.e., $k \rightarrow \infty$.}

\paragraph{$k$-MPV and $n$-step returns.}
The $k$-MPV should \emph{not} be confused with the $n$-step returns used in temporal difference~\citep[TD,][]{sutton1988learning} learning.
The former is an agent's estimate about its value%
, i.e., $\vhat_{\mhat}^{k} \approx v^{\pi}$ that uses both the learned value function and model.
The latter is a stochastic estimate of the \emph{environment's} $n$-step Bellman operator that can be used for constructing value target estimators in TD learning with reduced bias.
\looseness=-1

An ensemble of $k$-MPV predictions can be made by varying $k$.
We call this an \emph{implicit value ensemble}\footnote{Non-successive values of $k$ can be used in the construction of an IVE, e.g., $k \in \{ 1, 7, 13 \}$, but in practice this is less computationally efficient, see Section~\ref{subsec:practical-implementation}.} (IVE), depicted in Figure~\ref{subfig:ive-cg} and~\ref{subfig:ive-net} and denoted by
\begin{align}
 \{ \vhat_{\mhat}^{i} \}_{i=0}^{n} \triangleq \underdescribe{ \{ \vhat, \T_{\mhat}^{\pi} \vhat, \ldots, (\T_{\mhat}^{\pi})^{n} \vhat \} }{n+1 \text{ value estimates}}.
\label{eq:ive}
\end{align}

\note{%
Any agent with a model and value function is, in effect, also equipped with an ensemble of value functions.}

\subsection{Model-Value Inconsistency}
\label{subsec:model-value-inconsistency}

We term the disagreement of the IVE~components as \emph{model-value inconsistency} or just \emph{self-inconsistency}, for short, since it quantifies the Bellman-inconsistency~\citep{farquhar2021self} of the learned model and value function.

As our learned model and value function better approximate their ``true'' counterparts, the self-inconsistency reduces since the ``true'' model and value function are Bellman consistent, i.e., $(\T_{\mstar}^{\pi})^{n} v^{\pi} = (\T_{\mstar}^{\pi})^{l} v^{\pi}, \forall n, l \in \mathbb{N}$.
If the true model and value function are contained in the hypothesis classes of our approximators and the respective learning algorithms converge to the ``true'' solutions, then the self-inconsistency reduces to zero.

\note{%
In regions of state space where the learned model and value function are accurate, they are also self-consistent.
With high self-inconsistency the learned model or/and value should be inaccurate.}

Various metrics can be used to quantify the disagreement between the IVE components.
Since the {$k$-MPV}s are scalars, we can use any measure of disagreement of its components, e.g., the standard deviation across the IVE members, denoted by $\sigma$-IVE($n$) for $n$ members.
Similarly, we define $\mu$-IVE($n$) as the value prediction, given by the ensemble mean, and $\mu+\beta \cdot \sigma$-IVE($n$) as the weighted sum of the IVE mean and standard deviation, where $\beta \in \mathbb{R}$.
We can induce a self-inconsistency- (i) seeking; (ii) averse or (iii) neutral policy when $\beta>0$, $\beta<0$ and $\beta = 0$, respectively.
\looseness=-1

\subsection{Practical Implementation}
\label{subsec:practical-implementation}

We use parametric function approximators, in particular neural networks, to approximate the model and value function: $\theta$ are the model and $\phi$ are the value function parameters, from hypotheses classes $\Theta$ and $\Phi$, respectively, i.e., $\mhat(\cdot, \cdot | s, a; \theta) \approx \mstar(\cdot, \cdot | s, a)$ and $\vhat(s; \phi) \approx \vhat(s)$.

With small tabular models, such as the ones used for the gridworld in Figure~\ref{fig:std-value-ensembles}, we can calculate the {$k$-MPV}s exactly.
With neural network models, the calculation of the expectation in Eqn.~(\ref{eq:bellman-evaluation-one-step}) is generally intractable and hence we can only approximate it, e.g., in the case of stochastic models, via Monte Carlo (MC) sampling.
An MC sample of the $k$-MPV of state $s \in \mathcal{S}$ is given by:
\begin{align}
  \mathbf{\vhat_{\mhat}^{k}}(s) = \sum_{i=1}^{k-1} \gamma^{i-1}\mathbf{r_{\mhat}^{i+1}} + \gamma^{k}\vhat(\mathbf{s_{\mhat}^{k}}),
\label{eq:mc-estimator}
\end{align}
where $\mathbf{s_{\mhat}^{0}} = s$ and the samples from the model and policy are in \textbf{bold} and subscripted with $\mhat$ and $\pi$, i.e., $\mathbf{r_{\mhat}^{i+1}}, \mathbf{s_{\mhat}^{i+1}} \sim \mhat(\cdot, \cdot | \mathbf{s_{\mhat}^{i}}, \mathbf{a_{\pi}^{i}})$ and $\mathbf{a_{\pi}^{i}} \sim \pi(\cdot | \mathbf{s_{\mhat}^{i}})$.

In practice, to minimise the number of samples required to calculate an IVE prediction, we reuse the samples used for estimating the different components of the ensemble.
In particular, for every MC sample $\mathbf{\vhat_{\mhat}^{n}}(s)$, we use the sampled rewards, states and actions trajectories $\{(\mathbf{r_{\mhat}^{i+1}}, \mathbf{s_{\mhat}^{i+1}}, \mathbf{a_{\pi}^{i}}) \}_{i=0}^{n-1}$ to also estimate the ``preceeding'' ensemble components $\{\mathbf{\vhat_{\mhat}^{i}}(s)\}_{i=0}^{n-1}$.
This makes the computation of IVE no more expensive than online sample-based planning methods~\citep{hafner2019learning,schrittwieser2020mastering}.

\paragraph{Expectation models.}
Deterministic multi-step expectation models, e.g., the MuZero/Muesli model~\citep{schrittwieser2020mastering}, are especially well-suited for calculating IVEs in stochastic environments: they learn to predict \emph{expected} rewards and values conditioned on a sequence of actions, thereby implicitly averaging over stochastic state transitions.
To estimate the $k$-MPV in Eqn.~(\ref{eq:mc-estimator}), only policy samples are needed.
Empirically, in Appendix~\ref{app:ablations}, we found after an ablation that one sample from the policy sufficed.
\looseness=-1

\subsection{Diversity in the Implicit Value Ensemble}
\label{subsec:heterogeneous-ensemble-of-value-functions-from-a-learned-model}

The components of the IVE form a \emph{heterogeneous} ensemble~\citep{wichard2003building} since they differ in (i) functional form, and (ii) learning algorithm.
Next, we elaborate on how these can impact the diversity of the IVE predictions.

\paragraph{Functional form.}
While the ensemble components share the same model and value parameters, $\theta$ and $\phi$, respectively, they make predictions by composing these parameters differently.
For $k=0$, only the parameters of the value functions are used for making predictions.
As $k$ grows, the contribution of the model parameters to the prediction increases.
For instance, the $1$-MPV and $5$-MPV, i.e., $\vhat_{\mhat}^{1}$ and $\vhat_{\mhat}^{5}$, are both functions parametrised by $\theta$ and $\phi$ but their functional dependence on $\theta$ and $\phi$ is generally different.
This introduces diversity in the ensemble since different functions will have different generalisation properties and their predictions in out-of-distribution states are expected to differ, for an illustration, see Figure~\ref{fig:ive-cg-gp} and Appendix~\ref{app:experimental-details} for an exposition.

Variability between IVE members is also introduced by the training procedure.
The exact details depend on the algorithm used to learning algorithm.
Next, we analyse the Muesli model and value learning algorithms~\citep{hessel2021muesli} and their impact on the diversity on the IVE members.

\paragraph{Muesli learning algorithm.}
In training from a sequence of interactions, the deterministic expectation model is unrolled from each state for $K$ steps, following the actions that were taken in the environment.
The bootstrap target used to update the $i$'th resulting value estimate $\vhat(s_{t+i}), i \in \{0, \dots, K\}$ uses the environment samples from $t+i$ to $t+i+n$.
This receding horizon means that each value estimate, and therefore each member of the IVE, is regressed against a different target, furthering the diversity among their predictions.
See Appendix~\ref{app:muesli-and-ive} for more details.

%% file: figures/std-value-ensembles.tex
\begin{figure*}[t]
  \centering
  \begin{subfigure}[b]{0.14\linewidth}
    \centering
    \includegraphics[width=\linewidth]{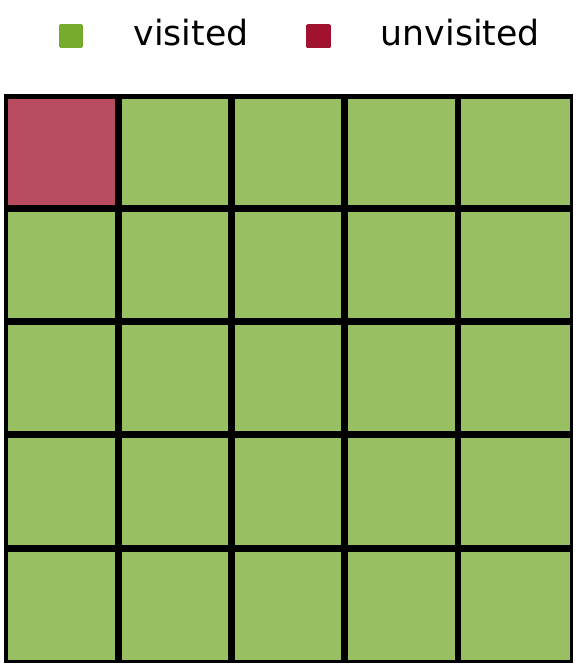}
    \caption{Dataset}
    \label{subfig:dataset}
  \end{subfigure}
  ~
  \begin{subfigure}[b]{0.14\linewidth}
    \centering
    \includegraphics[width=\linewidth]{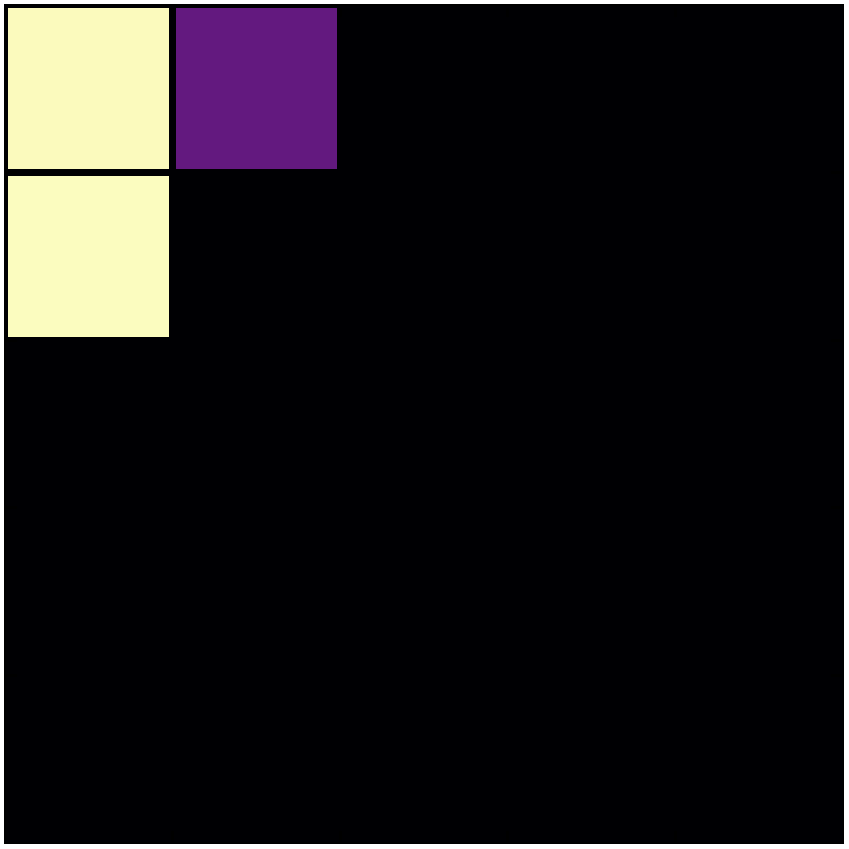}
    \caption{$\sigma$-IVE(1)}
    \label{subfig:sigma-ive-1}
  \end{subfigure}
  ~
  \begin{subfigure}[b]{0.14\linewidth}
    \centering
    \includegraphics[width=\linewidth]{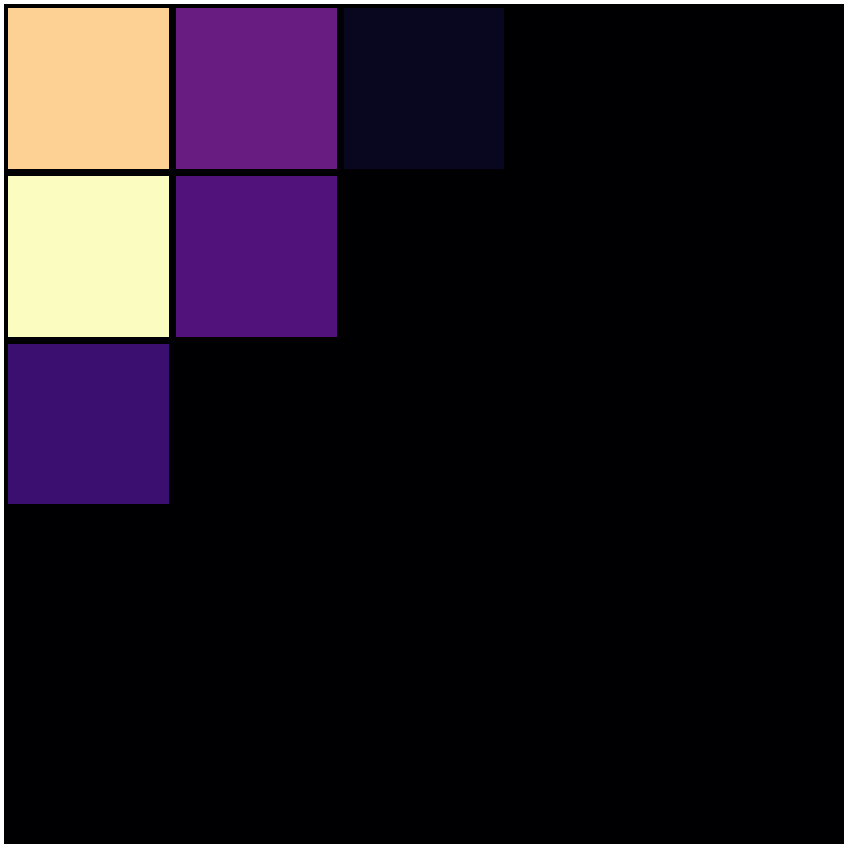}
    \caption{$\sigma$-IVE(2)}
    \label{subfig:sigma-ive-2}
  \end{subfigure}
  ~
  \begin{subfigure}[b]{0.14\linewidth}
    \centering
    \includegraphics[width=\linewidth]{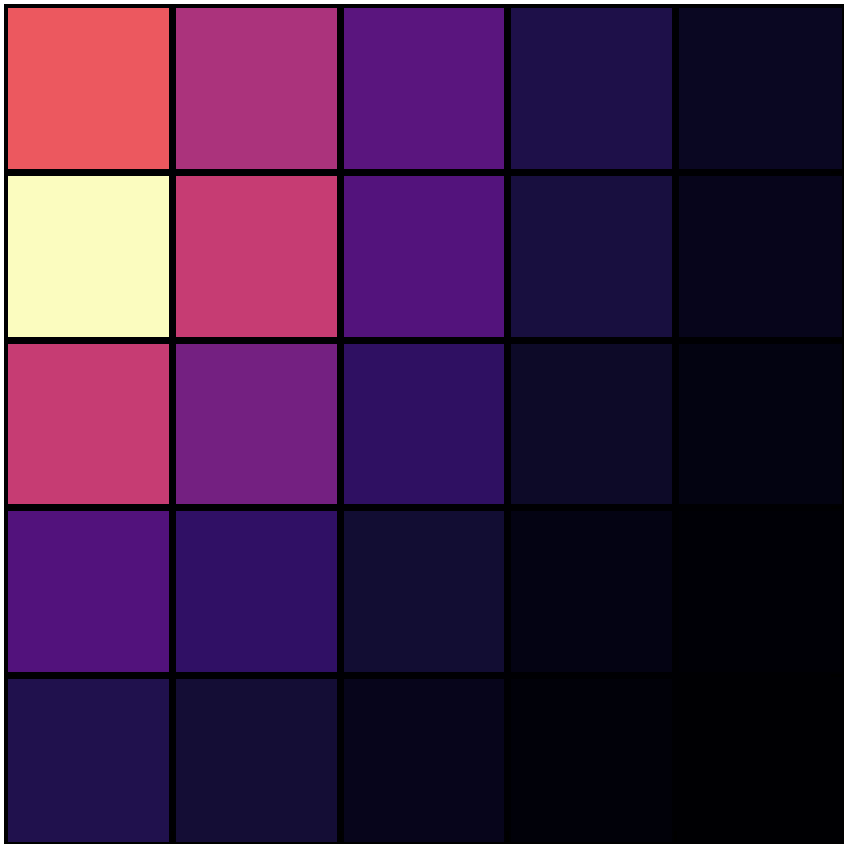}
    \caption{$\sigma$-IVE(20)}
    \label{subfig:sigma-ive-20}
  \end{subfigure}
  ~
  \begin{subfigure}[b]{0.14\linewidth}
    \centering
    \includegraphics[width=\linewidth]{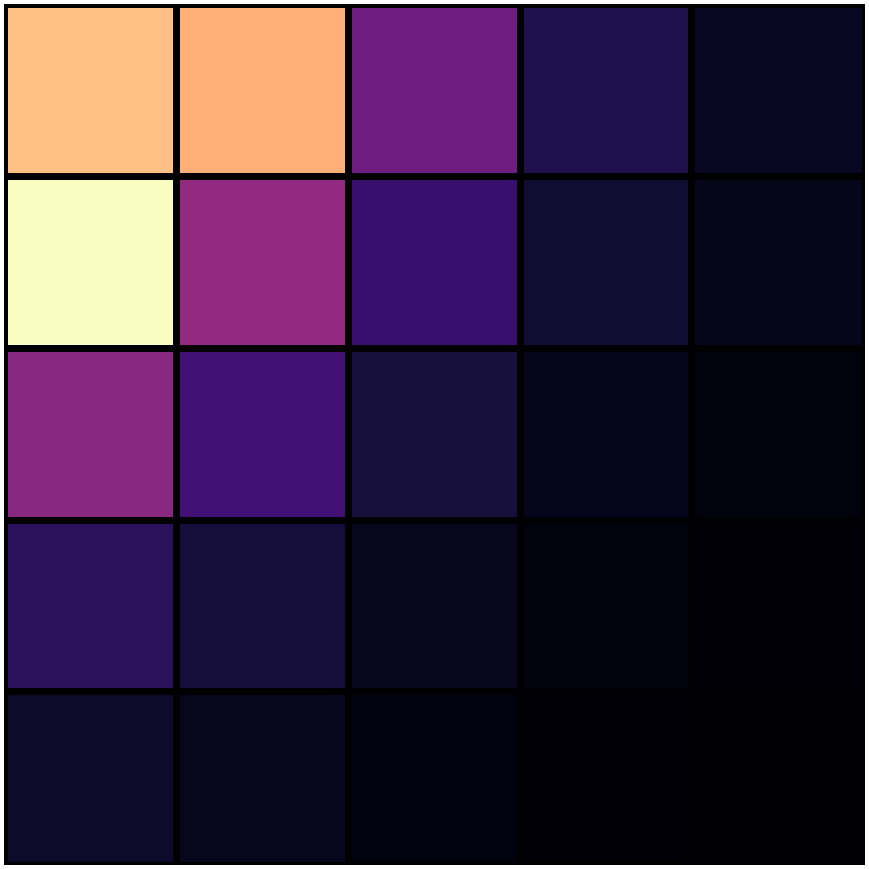}
    \caption{$\sigma$-EVE(20)}
    \label{subfig:sigma-eve-20}
  \end{subfigure}
  ~
  \begin{subfigure}[b]{0.14\linewidth}
    \centering
    \includegraphics[width=\linewidth]{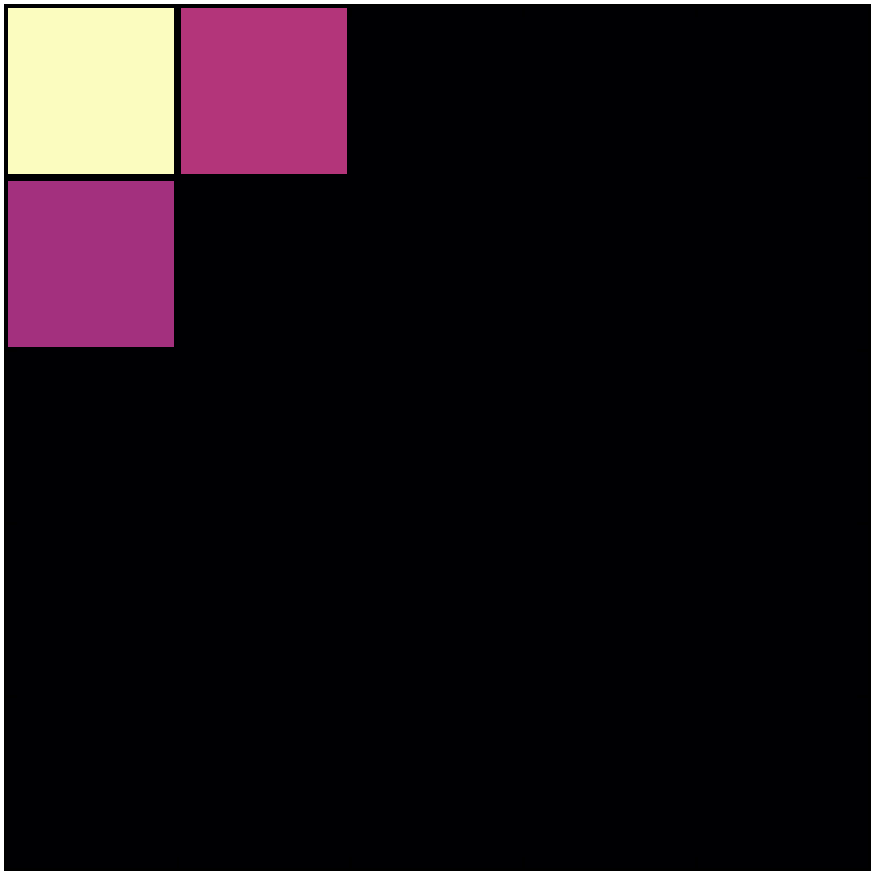}
    \caption{$\sigma$-EMVE(20)}
    \label{subfig:sigma-emve-20}
  \end{subfigure}
  ~
  \begin{subfigure}[t]{0.02\linewidth}
    \centering
    \includegraphics[height=9em]{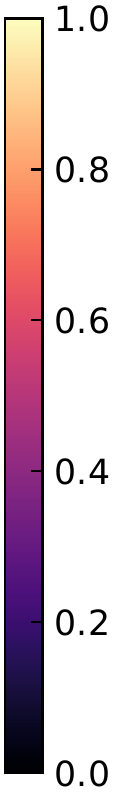}
  \end{subfigure}
  \caption{
    Model-value inconsistency ($\sigma$-IVE, see Section~\ref{subsec:model-value-inconsistency}) as the standard deviation across the implicit value ensemble (IVE, see Section~\ref{subsec:implicit-value-esemble}) for different numbers of ensemble components $n$.
    (a) The top left state of the \gridworld~is excluded from the data used to train the model $\mhat$ and value function $\vhat$.
    (b-e) The disagreement between the IVE predictions diffuse for (b) 1-step; (c) 2-steps and (d) 20-steps model unrolls.
    The same disagreement across explicit ensembles (e) $\sigma$-EVE and (f) $\sigma$-EMVE built from different initialisation parameters.
    The standard deviation $\sigma$ is normalised in range $[0, 1]$ per figure.
    \looseness=-1
  }
  \label{fig:std-value-ensembles}
  \vspace{-0.5em}
\end{figure*}

%% file: sections/4_experiments.tex
\section{Experiments}
\label{sec:experiments}

We conduct a series of tabular and deep RL experiments\footnote{Further experiments, details on the experimental protocol and implementations can be found in Appendices~\ref{app:ablations},~\ref{app:experimental-details} and~\ref{app:implementation-details}.} to determine how effective  model-value inconsistency is as a signal for epistemic uncertainty.
Our goal is \emph{not} to show that the IVE is better than explicit ensembles. Instead, since 
IVE is present in any model-based RL agent, we want to empirically study its properties and validate its usefulness.
\looseness=-1

\paragraph{Baselines.}
In the tabular experiments, we learn value functions with expected SARSA~\citep{van2009theoretical} and use maximum likelihood estimation for model learning (see Section~\ref{sec:background}).
The explicit ensemble components are trained independently, using exactly the same data.
The only sources of variability are random initialisation of parameters and stochastic gradient descent.\looseness=-1

In the deep RL experiments, we built on the following model-based agents, that use either the MLE or VE model learning principles, described in Section~\ref{sec:background}:
(i) \textbf{Muesli}~\citep{hessel2021muesli} is a policy optimisation method with a learned multi-step expectation model.
Muesli also learns a state-value function, using Retrace~\citep{munos2016safe} to correct for the off-policiness of the replayed experience.
The learned model is used for representation learning and for constructing action-value estimates, by one-step model unroll, used for policy improvement.
The model parameters are trained to predict reward and value $k$-steps into the future (corresponding to the individual terms in the $k$-MPV);
(ii) \textbf{Dreamer}~\citep{hafner2019dream} is a policy optimisation method with an MLE model.
The model is an action-conditioned hidden Markov model, trained to maximise (a lower bound on) the likelihood of the reward and observation sequences.
Dreamer learns a value function using \emph{only} rollouts from the learned model and its parameters are learned
such that the learned value function becomes (self-)consistent with the model;
(iii) \textbf{VPN}~\citep{oh2017value} is a value-based planning method with a multi-step expectation model.
The action-value function and model are trained simultaneously with $n$-steps Q-learning~\citep{watkins1992q}.
In this case, the $k$-MPV is the value estimate after applying $k$ times the model-induced Bellman \emph{optimality} operator on the learned value function (see Appendix~\ref{app:extensions} for a formal exposition).
\looseness=-1

\paragraph{Environments.}
In the tabular experiments, we use an empty $5 \times 5$ \gridworld, and collect data by rolling out a uniformly random policy, initialised at the bottom right cell.
We exclude from the dataset any transitions to the top left cell, as illustrated in Figure~\ref{subfig:dataset}, in order to control for visited (in-distribution) and unvisited (out-of-distribution) states.
\looseness=-1

In the deep RL experiments, we use a selection of 5 tasks from the \procgen~suite~\citep{procgen19} to (i) control the number of distinct levels used for training the agent (i.e., \levels) and (ii) hold out a set of test levels that are not seen during training.
We also use a modification of the \walker~walk task from the DeepMind Control suite~\citep{tunyasuvunakool2020dm_control}.
The original \walker~task has a per-step reward $r_t$ bounded in $[0,1]$ which is computed based on the agent's torso height and forward velocity.
To parameterise exploration difficulty, we modify the reward function to set any reward less than $\eta$ to zero: $\tilde{r}_{t} = \mathcal{H}(r_{t} - \eta)r_{t}$, where $\mathcal{H}$ is the Heaviside step function.
For large $\eta$, agents that rely on naive exploration methods will struggle to find rewards and solve the task.
Lastly, we use the original \minatar~\citep{young2019minatar} suite for fast experimentation with value-based agents~\citep{mnih2013playing}.
\looseness=-1

\subsection{Detecting Out-Of-Distribution Regimes with Self-Inconsistency}
\label{subsec:detecting-out-of-distribution-regimes-with-self-inconsistency}

Based on the proposed role of self-inconsistency as a signal for epistemic uncertainty, and how epistemic uncertainty changes between in- and out-of-distribution regimes, we expect the following hypotheses to hold.
\textbf{H1}: Self-inconsistency is low in in-distribution regions of the state-action space.
\textbf{H2:} Self-inconsistency is high in out-of-distribution (OOD) regions.
\textbf{H3:} Self-inconsistency in an OOD test distribution is reduced by bringing the training distribution closer to it.
\looseness=-1

\paragraph{Tabular.}
Figures~\ref{subfig:sigma-ive-1}-\ref{subfig:sigma-ive-20} show the self-inconsistency, measured as $\sigma$-IVE($n$) for different values of $n$, in the tabular \gridworld.
As $n$ grows from $1$ to $20$, the standard deviation across the IVE is qualitatively similar to the explicit value ensemble's (EVE) in Figure~\ref{subfig:sigma-eve-20}.
We observe that the self-inconsistency is lower for visited states (\textbf{H1}) than unvisited (OOD) ones (\textbf{H2}).
\paragraph{Deep RL.}
Figure~\ref{fig:procgen-probing} shows the Muesli agent's performance (left) and its self-inconsistency (right)---calculated after training as the $\sigma$-IVE($5$)---for the different \procgen~tasks and for varying training \levels, after 100M environment steps.
The self-inconsistency for the training (in-distribution) levels is always low, regardless of the \levels~used for training the agent (\textbf{H1}).
We also observe that the self-inconsistency in the test (OOD) levels is higher than the train ones (\textbf{H2}).
Importantly, as the number of training levels increases the self-inconsistency on the test levels decreases, which confirms \textbf{H3}.
Also as expected, this reduced self-inconsistency correlates with improved test performance.
\looseness=-1

\input{figures/procgen-probing}

\subsection{Optimism and Pessimism in the Face of Self-Inconsistency}
\label{subsec:self-inconsistency-as-a-signal-for-exploration}

Epistemic uncertainty has been (i) sought to drive exploration~\citep{sekar2020planning} and (ii) avoided for acting safely~\citep{filos2020can}.
This section addresses two hypotheses.
\textbf{H4}: Self-inconsistency is an effective signal for exploration.
\textbf{H5:} Avoiding self-inconsistency leads to robustness to distribution shifts.

\paragraph{Tabular.}
Figure~\ref{subfig:prob_reaching_ood_state_ofu} shows the probability of reaching the novel state in \gridworld~when a self-inconsistency-seeking policy is followed (+$\sigma$-IVE). 
Seeking self-inconsistency improves upon a uniformly random or greedy policy and is on par with an explicit ensemble of values (EVE) method (\textbf{H4}).
For the experiment in Figure~\ref{subfig:prob_reaching_ood_state_pfu}, a distribution shift is performed by raising the environment stochasticity from $\delta = 0.1$ to $\delta = 0.5$, and the probability of a self-inconsistency-avoiding policy (-$\sigma$-IVE) is illustrated.
We observe that the self-inconsistency-avoiding policy is robust to the drift of the environment dynamics (\textbf{H5}).

\input{figures/reaching-ood-states}

\paragraph{Deep RL.}
Table~\ref{tab:dreamer-mvsic-walker-walk} gives the performance of the Dreamer agent and variants that use the model for online planning~\citep{ma2020contrastive} as we increase reward sparsity for the \walker~task, e.g., $\eta=0$ is the original task and $\eta=0.5$ sets rewards below $0.5$ to zero.
We used the mean of IVE components in place of the learned policy for acting ($\mu$-IVE(5)), and  combined the mean with the self-inconsistency signal for acting optimistically in the face of uncertainty ($\mu+\sigma$-IVE(5)).  
The self-inconsistency-seeking Dreamer-variant, i.e., $\mu+\sigma$-IVE($5$), is performing well for $\eta=0.3$ and $\eta=0.5$ while the base agent fails, corroborating \textbf{H4}.
Similar to the tabular experiment results, the IVE is on par with the the explicit value ensemble (EVE, Figure~\ref{subfig:vensemble-net}) and outperforms the explicit model value ensemble (EMVE, Figure~\ref{subfig:mensemble-net}).\looseness=-1

\input{tables/dreamer-mvsic-walker-walk}

\subsection{Planning with Averaged Model-Predicted Values}
\label{subsec:planning-with-ensembled-values}

Bayesian model averaging (BMA), i.e., integrating over epistemic uncertainty for making predictions, has been used to boost performance~\citep{wilson2020bayesian}.
The interpretation of the IVE as an ensemble allows to justify prior methods in the literature that have argued for averaging {MPV}s~\citep{oh2017value,byravan2020imagined} in order to robustify value-based planning, casting them as approximate BMA methods.
This section addresses one hypothesis:
\textbf{H6:} Ensemble averaging of the IVE members is \emph{in general} more robust for value prediction than any component individually.

\paragraph{Deep RL.}
Table~\ref{tab:vpn-ive-minatar} shows the final performance of a VPN(5) agent that uses $\mu$-IVE($5$) value targets and its $\vhat_{\mhat}^{1}$ and $\vhat_{\mhat}^{5}$ variants' on the \minatar~tasks.
The ensembled $\mu$-IVE($5$) value predictor is consistently better than the single value predictors, supporting \textbf{H6}.
\looseness=-1

\input{tables/vpn-ive-minatar}

%% file: figures/procgen-probing.tex
\begin{figure*}
    \centering
    \includegraphics[height=3.5cm]{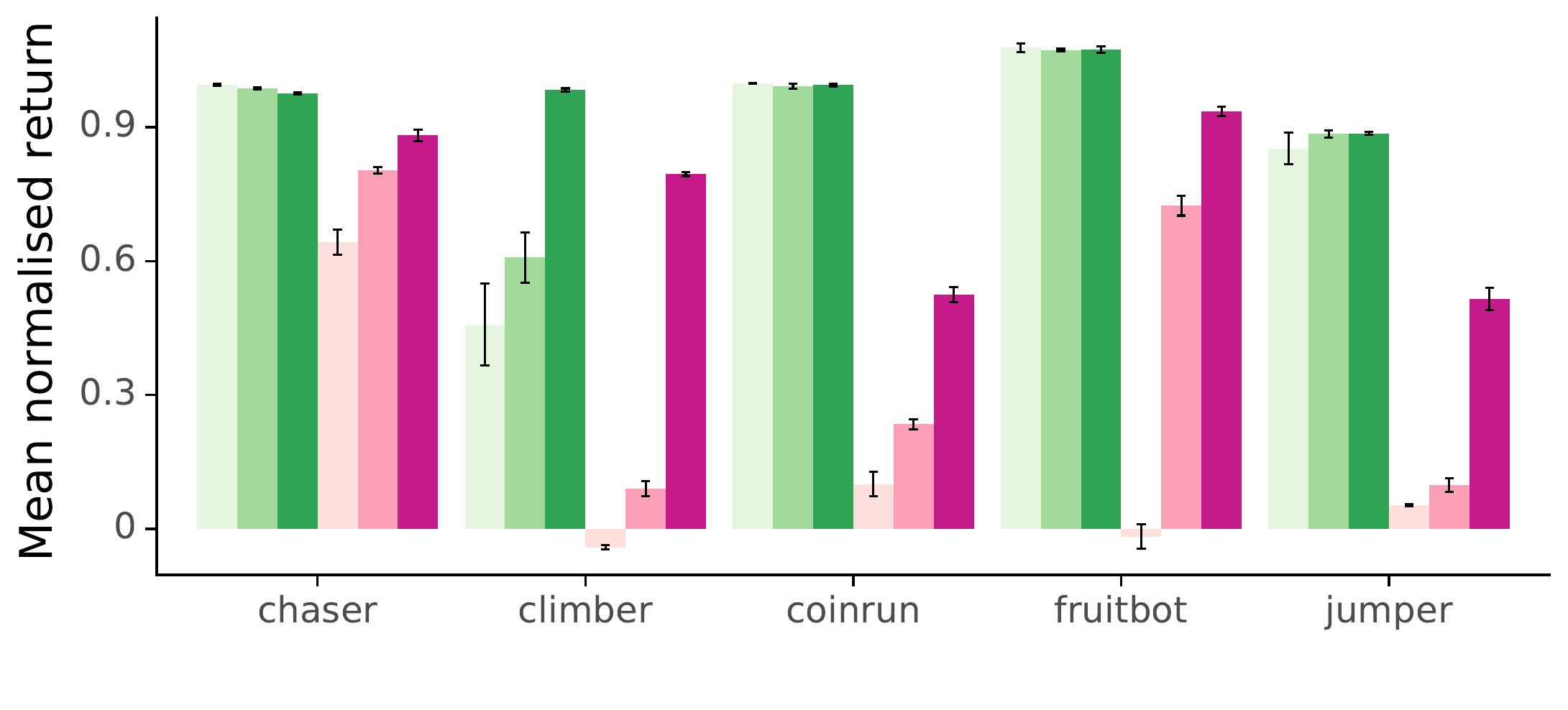}
    \includegraphics[height=3.5cm]{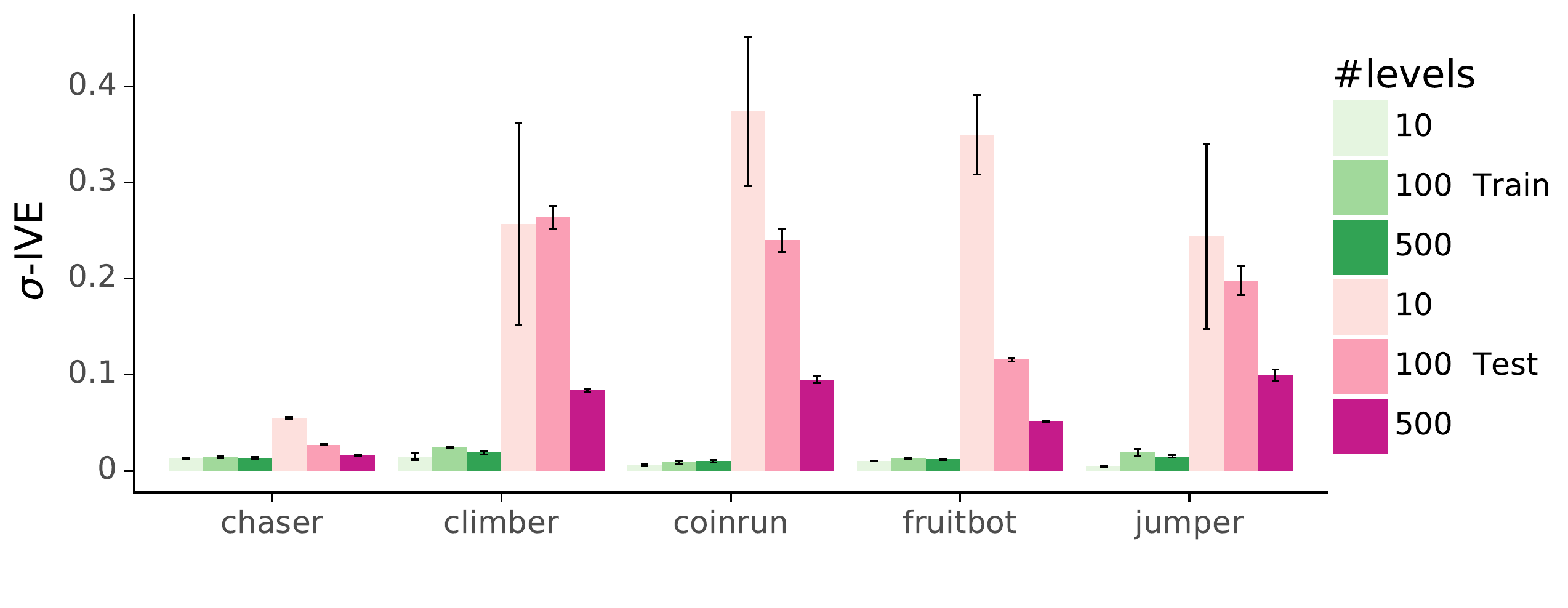}
    \vspace{-1.5em}
    \caption{
    Left: Normalised training and test performance for a Muesli agent evaluated on both training and unseen test levels of 5 \procgen~games after 100M environment frames, for different numbers of unique levels seen during training.
    Values are normalised by the min and max scores for each game. 
    Right: $\sigma$-IVE($5$) computed using the model of the Muesli agent while evaluating on both training and unseen test levels, for different numbers of unique levels seen during training. Bars, error-bars show mean and standard error across 3 seeds, respectively.}
    \label{fig:procgen-probing}
\end{figure*}

%% file: figures/reaching-ood-states.tex
\begin{figure}[h]
  \centering
  \includegraphics[width=\linewidth]{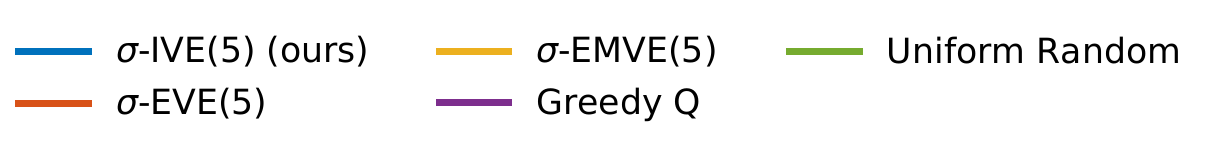}
    \begin{subfigure}[b]{0.48\linewidth}
      \centering
      \includegraphics[width=\linewidth]{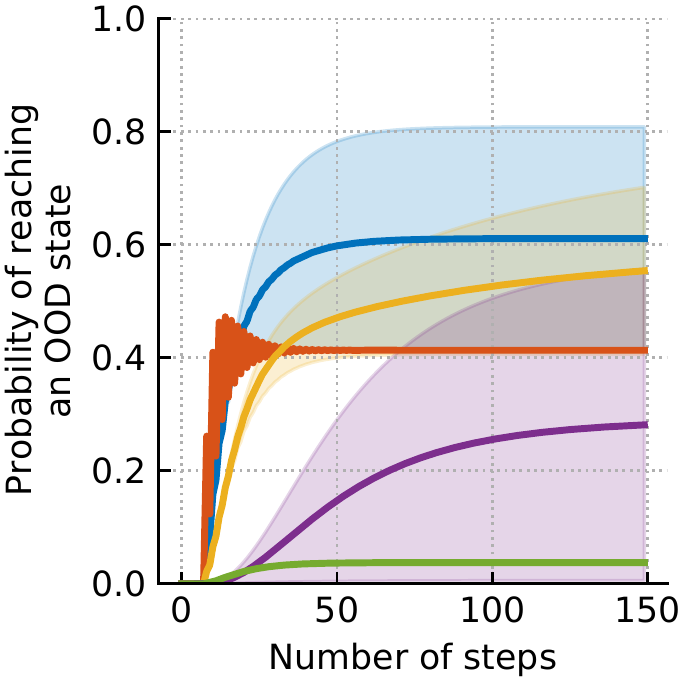}
      \caption{Optimism}
      \label{subfig:prob_reaching_ood_state_ofu}
    \end{subfigure}
    ~ 
    \begin{subfigure}[b]{0.48\linewidth}
      \centering
      \includegraphics[width=\linewidth]{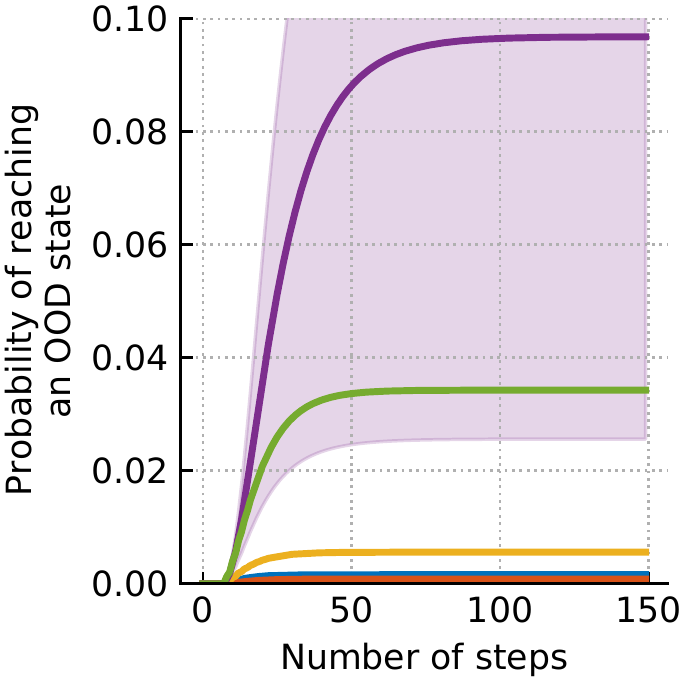}
      \caption{Pessimism}
      \label{subfig:prob_reaching_ood_state_pfu}
    \end{subfigure}
  \caption{
    Probability of reaching the out-of-distribution state in a tabular \gridworld, starting from the bottom right cell (Figure~\ref{subfig:dataset}) by (a) seeking or (b) avoiding self-inconsistency ($\sigma$-IVE, see Section~\ref{subsec:model-value-inconsistency}) or explicit value or model ensemble (EVE, EMVE) standard deviation.
    Error bars show standard error over 100 seeds.
    \looseness=-1
  }
  \label{fig:reaching-ood-states}
\end{figure}

%% file: tables/dreamer-mvsic-walker-walk.tex
\begin{table}[!h]
  \centering
  \caption{
    Pixel-based continuous control experiments.
    Results for the Dreamer agent and IVE variants on a modified version of the Walker Walk task with varying degrees of reward sparsity controlled by $\eta$, where higher $\eta$ corresponds to harder exploration.
    A ``$\diamondsuit$'' indicates methods that use online-planning for acting.
    We report mean and standard error of episodic returns (rounded to the nearest tenth) over 3 runs after 1M steps.
    Higher-is-better and the performance is upper bounded by 1000.
    The \textbf{best performing} method, per-task, is in bold.
  }
  \label{tab:dreamer-mvsic-walker-walk}
  \resizebox{\linewidth}{!}{
  \begin{tabular}{lrrrrr}
  \toprule
  \textbf{Methods} &
  \multicolumn{1}{c}{$\eta = 0.0$}      &
  \multicolumn{1}{c}{$\eta = 0.2$}      &
  \multicolumn{1}{c}{$\eta = 0.3$}      &
  \multicolumn{1}{c}{$\eta = 0.5$}      \\
  \midrule
  Dreamer                               &
  \bftab 1000${\color{black!50}\pm00}$  &
  720${\color{black!50}\pm10}$          &
  570${\color{black!50}\pm60}$          &
  80${\color{black!50}\pm50}$           \\
  %
  %
  
  %
  \rowcolor{ourmethod} $\mu$-IVE(5)$^{\diamondsuit}$ &
  \bftab 1000${\color{black!50}\pm00}$      &
  860${\color{black!50}\pm40}$              &
  690${\color{black!50}\pm70}$              &
  210${\color{black!50}\pm60}$              \\
  \midrule
  $\mu+\sigma$-EVE(5)$^{\diamondsuit}$      &
  \bftab 1000${\color{black!50}\pm00}$      &
  \bftab 1000${\color{black!50}\pm00}$      &
  \bftab 980${\color{black!50}\pm10}$       &
  \bftab 280${\color{black!50}\pm50}$       \\
  $\mu+\sigma$-EMVE(5)$^{\diamondsuit}$     &
  \bftab 1000${\color{black!50}\pm00}$      &
  910${\color{black!50}\pm20}$              &
  730${\color{black!50}\pm40}$              &
  210${\color{black!50}\pm60}$              \\
  \rowcolor{ourmethod} $\mu+\sigma$-IVE(5)$^{\diamondsuit}$ &
  \bftab 1000${\color{black!50}\pm00}$      &
  \bftab 1000${\color{black!50}\pm00}$      &
  \bftab 1000${\color{black!50}\pm00}$      &
  \bftab 330${\color{black!50}\pm70}$       \\
  \bottomrule
  \end{tabular}
  }
\vspace{-1em}
\end{table}

%% file: tables/vpn-ive-minatar.tex
\begin{table}[!h]
  \centering
  \caption{
    Value-based planning experiments on \minatar~tasks, testing the impact of planning with the IVE ensembled mean.
    The original VPN(5) is the same with our $\mu$-IVE(5).
    Non-ensembled value targets ($\vhat_{\mhat}^{1}$, $\vhat_{\mhat}^{5}$) lead to significant deterioration in final performance.
    We report mean and standard error of episodic returns over 3 runs after 2M steps, higher-is-better.
    The \textbf{best performing} method, per-task, is in bold.\looseness=-1
  }
  \label{tab:vpn-ive-minatar}
  \resizebox{\linewidth}{!}{
  \begin{tabular}{lrrrrrr}
  \toprule
  \textbf{Methods}          &
  \texttt{Asterix}          &
  \texttt{Breakout}         &
  \texttt{Freeway}          &
  \texttt{Seaquest}         &
  \texttt{S. Inv.}       \\
  \midrule
  DQN                                       &
  14.7${\color{black!50}\pm0.4}$            &
  12.1${\color{black!50}\pm1.2}$            &
  \bftab 49.6${\color{black!50}\pm0.3}$     &
  2.3${\color{black!50}\pm0.6}$             &
  47.2${\color{black!50}\pm1.3}$            \\
  \midrule
  VPN+$\vhat_{\mhat}^{1}$                   &
  15.1${\color{black!50}\pm0.6}$            &
  13.8${\color{black!50}\pm0.8}$            &
  \bftab 49.1${\color{black!50}\pm0.7}$     &
  4.7${\color{black!50}\pm0.9}$             &
  53.9${\color{black!50}\pm1.8}$            \\
  VPN+$\vhat_{\mhat}^{5}$                   &
  7.1${\color{black!50}\pm2.3}$             &
  4.2${\color{black!50}\pm2.3}$             &
  24.3${\color{black!50}\pm4.2}$            &
  1.2${\color{black!50}\pm1.4}$             &
  28.6${\color{black!50}\pm8.3}$            \\
  \midrule
  \rowcolor{ourmethod}
  $\mu$-IVE(5)                              &
  \bftab 18.3${\color{black!50}\pm0.2}$     &
  \bftab 22.0${\color{black!50}\pm0.7}$     &
  \bftab 49.4${\color{black!50}\pm0.5}$     &
  \bftab 8.6${\color{black!50}\pm0.3}$      &
  \bftab 97.3${\color{black!50}\pm9.6}$    \\
  \bottomrule
  \end{tabular}
  }
\end{table}

%% file: sections/5_related_work.tex
\section{Related Work}
\label{sec:related-work}

\paragraph{Ensemble RL methods.}
Ensembles of deep neural networks have been used in value and model-based online RL methods for (i) stabilising learning~\citep{fausser2015neural,anschel2017averaged,kalweit2017uncertainty,kurutach2018model,chua2018deep}; (ii) exploration by seeking epistemic uncertainty~\citep{osband2016deep,shyam2019model,pathak2019self,flennerhag2020temporal,ball2020ready,sekar2020planning}; (iii) tackling distribution shifts~\citep{lowrey2018plan,kenton2019generalizing,agarwal2020optimistic} and (iv) representation learning~\citep{fedus2019hyperbolic,dabney2020value,lyle2021effect}.
All of the above consider explicit ensemble methods (see Section~\ref{sec:background}) which can be graphically represented by Figures~\ref{subfig:vensemble-net} and~\ref{subfig:mensemble-net} or some combination of them. 
In contrast, IVE is an implicit ensemble method that does not rely on an ensemble of either value functions or models but uses a single (point) estimate.
IVE could be combined with explicit ensembles, this would break the correlation between its ensemble components since parameters would not be shared, at the expense of growing the model size.\looseness=-1

\paragraph{Model-based RL.}
Learned models can be useful to RL agents in various ways, such as: (i) action selection via planning~\citep{richalet1978model,hafner2019learning}; (ii) representation learning~\citep{schmidhuber1990line,jaderberg2016reinforcement,lee2019stochastic,guez2020value,hessel2021muesli}; (iii) planning for policy optimisation or value learning~\citep{werbos1987learning,sutton1991dyna,hafner2019learning,byravan2020imagined}; or (iv) a combination of all of them~\citep{schrittwieser2020mastering}.
In this work, we use the learned model-induced Bellman operator and value function to construct an ensemble of value estimators and interpret the disagreement of their predictions as a proxy of epistemic uncertainty.

\paragraph{Model-value expansion.}
Alternative methods predict values by unrolling the learned model for $k$-steps and bootstrapping from the model-free learned value function, see Figures~\ref{subfig:ive-cg} and~\ref{subfig:ive-net}.
~\citet{feinberg2018model,buckman2018sample,byravan2020imagined} follow a two-steps process: (i) they learn a model by maximum likelihood (Section~\ref{subsec:model-based-reinforcement-learning}) and then (ii) learn the value function by regressing it to MPV predictions/targets.
~\citet{oh2017value,silver2017predictron,farquhar2017treeqn,gregor2019shaping,schrittwieser2020mastering,nikishin2021control} train the model and value function jointly, with a direct regression loss on the MPV.
Both the IVE and self-inconsistency signal are compatible with these learning approaches.

\paragraph{Adapting $k$.}
With varying $k$, MPV interpolates between the learned model and value function.
In particular, for (i) $k=0$ the value predictions are based only on the learned value function and for (ii) $k \rightarrow \infty$ only the learned model contributes to the value predictions.
The $\lambda$-predictron~\citep{silver2017predictron} uses a learned and adaptive mechanism for mixing the predictions for different $k$s.
STEVE~\citep{buckman2018sample} is an epistemic-uncertainty-informed mechanism for weighting the different MPVs.
It learns an explicit ensemble of models and value functions and weights the MPV using an inverse variance weighting of the means, calculated across the \emph{explicit} ensemble.
This should not be confused with our $\sigma$-IVE(n) signal, which is the variance across the MPVs and cannot be used for selecting the ``best'' $k$-th element but quantifies the model-value disagreement.

\paragraph{Novelty signals.}
Non-explicit ensemble methods have been proposed for estimating the model prediction error and use this as a proxy signal for novelty.
Most of these methods make novelty predictions for a state $s_{t}$, \emph{after} observing a transition $s_{t} \overset{a_{t}}{\longrightarrow} s_{t+1}$~\citep{stadie2015incentivizing,pathak2017curiosity,raileanu2020ride} and therefore are termed \emph{retrospective novelty predictors} in the literature~\citep{sekar2020planning}.
~\citet{lopes2012exploration} assume that the agent's learning progress is a predictable process and fit a model to it.
While $(s_{t}, a_{t}, s_{t+1})$ triplets are necessary for training the novelty predictor, after training, the signal can be calculated \emph{before} observing $s_{t+1}$ and hence can be used for planning purposes, which we term a \emph{plannable novelty predictor}.
The $\sigma$-IVE signal can be interpreted as a prediction error estimate that quantifies how the learned value function and model disagree in their predictions and hence we can use it as a plannable novelty signal.

\paragraph{Self-consistency regularisation.}
~\citet{silver2017predictron} and~\citet{farquhar2021self} regularised their learned value and model pairs to be self-consistent for prediction and control tasks, respectively.
Self-consistency regularisation has been used for learned world models by matching the predictions of a forward dynamics model with a backward dynamics model~\citep{yu2021playvirtual}.
Similar regularisation ideas have been used in other areas of machine learning, including offline multi-task inverse RL~\citep{filos2021psiphi}, natural language processing~\citep{bojar2011improving,edunov2018understanding} and generative modelling~\citep{zhu2017unpaired}.
All prior work directly ``forces'' self-consistency on modelled quantities as a form of regularisation, e.g., applied on imagined data~\citep{farquhar2021self}.
Instead, we treat self-inconsistency as a proxy for epistemic uncertainty and, e.g., indirectly promote self-consistency by actively guiding data collection/exploration with a self-inconsistency-seeking policy (see Section~\ref{subsec:self-inconsistency-as-a-signal-for-exploration}).
Consequently, this avoids degenerate but self-consistent solutions since the learned model and value functions are trained on real data (i.e., external consistency).\looseness=-1

\paragraph{Implicit NN ensembles.}
Ensembles from a single NN have been proposed and successfully used in supervised learning but they require modifications to the learning algorithm~\citep{huang2017snapshot,maddox2019simple,antoran2020depth} or architecture~\citep{huang2016deep,dusenberry2020efficient}.
In contrast, IVE relies on the structure of the RL problem and leverages the Bellman consistency~\citep{farquhar2021self} that the ``true'' model and value function satisfy and hence their learned counterparts should also do.

%% file: sections/6_discussion.tex
\section{Discussion}
\label{sec:discussion}

We have introduced  model-value self-inconsistency as a signal for capturing RL agents' epistemic uncertainty.
Our key insight is that a \emph{single} (point) estimate of a world model and value function can be used to generate multiple estimates of the state value, which can be combined to form an implicit value ensemble (IVE).
We showed empirically that self-inconsistency of the IVE---i.e., the disagreement amongst its members--- is an effective signal for epistemic uncertainty in tabular and pixel-based deep RL settings.
We then demonstrated that self-inconsistency can be used to guide exploration, increase an agent's ability  to handle distribution shifts, and robustify value-based planning methods.

\paragraph{Future work.}
We want to explore ways to:
(i) Modify the model, value-learning algorithms, or network architecture to increase  diversity in the IVE while keeping the model size unchanged, such as using different sub-samples of the data to train each IVE member or injecting $k$-dependent structured noise~\citep{osband2018randomized}. 
(ii) Integrate the self-inconsistency signal into more complex online planning methods~\citep[e.g., MCTS,][]{coulom2006efficient} since they already compute some ``modification'' of the IVE components.

\paragraph{Acknowledgements.}
We thank Mark Rowland, Lo\"ic Matthey, Hado van Hasselt, Theophane Weber, Ioannis Antonoglou, Amin Barekatain, Junhyuk Oh, Abhinav Gupta, Panagiotis Tigas, Yarin Gal, David Silver and Satinder Singh for helpful discussions and feedback.

%% file: sections/A_experimental_details.tex
\section{Experimental Details}
\label{app:experimental-details}

In this section, we describe the environments used in our experiments (see Section~\ref{sec:experiments}) and the experiment design.

\subsection{Environments}
\label{app:rl-environments}

In this section, we provide details on the specification of each task used in our experiments.

\subsubsection{Tabular Environment}
\label{app:tabular-environment}

\begin{minipage}{0.75\textwidth}
  We use an empty $5 \times 5$ gridworld (\gridworld) environment for our tabular experiments.
  The task is specified by:
  \begin{enumerate}[noitemsep]
    \item \textbf{State space, $\mathcal{S}$:}
      A finite discrete state space, i.e., $s \in \{0, 1, \ldots, 24\}$.
    \item \textbf{Action space, $\mathcal{A}$:}
      A finite discrete action space for moving the agent in the four cardinal directions (\texttt{N}, \texttt{W}, \texttt{S}, \texttt{E}), i.e., $s \in \{0, 1, 2, 3\}$.
    \item \textbf{Reward function, $r(s, a)$:}
      The zero function, i.e., $r(s, a) = 0, \forall (s, a) \in \mathcal{S} \times \mathcal{A}$.
    \item \textbf{Transition dynamics, $p(s' | s, a)$:}
      We consider the episodic setting, i.e., \texttt{episode\_length = 20}, and the dynamics are (optionally) stochastic.
      In particular, we use a single parameter that controls the stochasticity, called \texttt{wind\_prob} $\in [0, 1]$ and implement stochastic dynamics as actuator noise, i.e., there is a \texttt{wind\_prob} probability that the agent action is ignores and an other action is applied to the environment by sampling randomly from the action space.
  \end{enumerate}
\end{minipage}
\hspace{1.0em}
\begin{minipage}{0.20\textwidth}
  \centering
  \includegraphics[width=\linewidth]{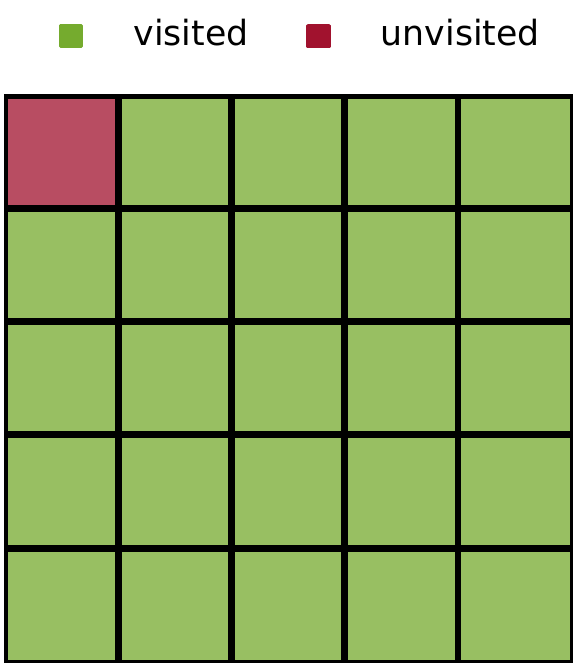}
  \captionof{figure}{\gridworld}
\end{minipage}

\subsubsection{Procgen~\citep{procgen19}}
\label{subsubapp:procgen}

\begin{figure*}[!h]
  \centering
  \begin{subfigure}[b]{0.18\linewidth}
    \centering
    \includegraphics[width=\linewidth]{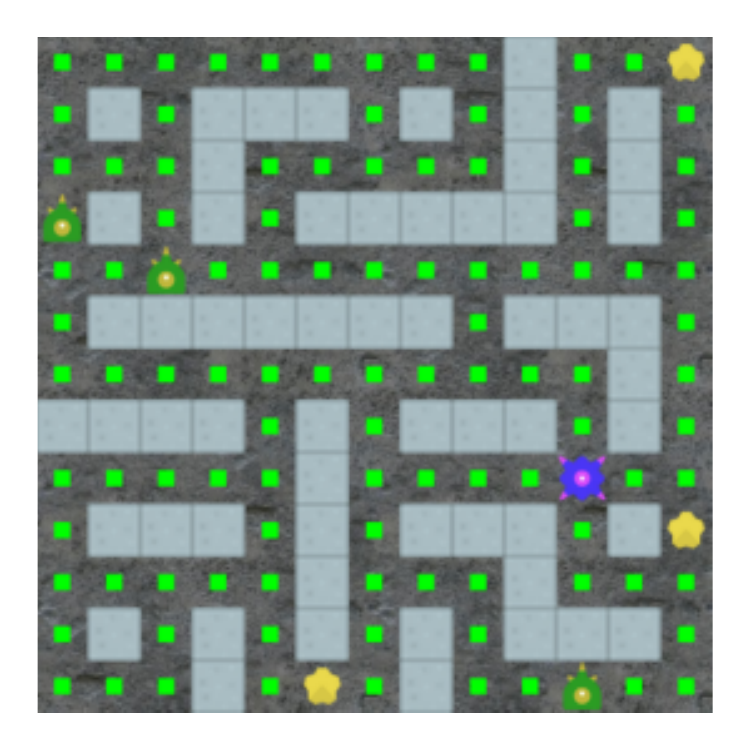}
    \caption{\texttt{chaser}}
    \label{subfig:chaser}
  \end{subfigure}
  ~
  \begin{subfigure}[b]{0.18\linewidth}
    \centering
    \includegraphics[width=\linewidth]{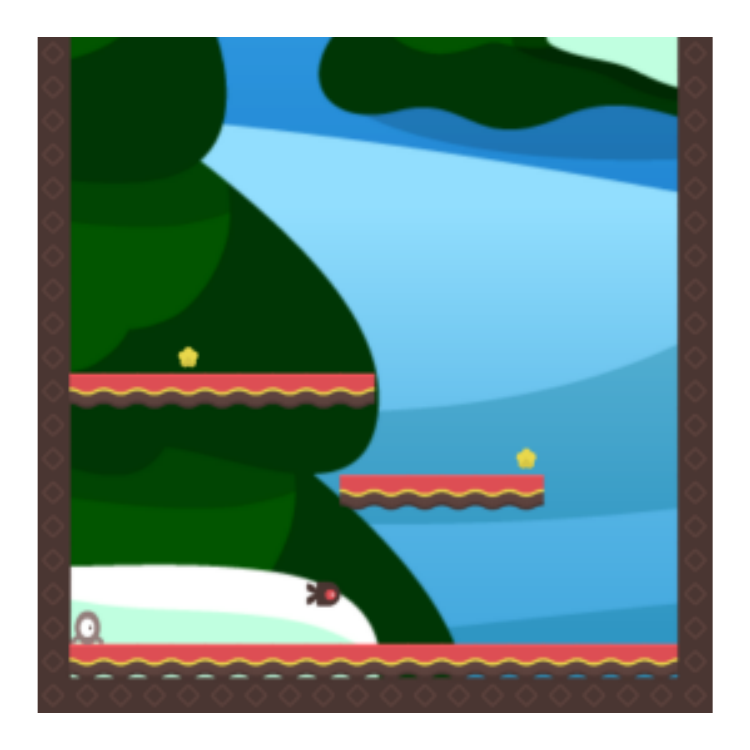}
    \caption{\texttt{climber}}
    \label{subfig:climber}
  \end{subfigure}
  ~
  \begin{subfigure}[b]{0.18\linewidth}
    \centering
    \includegraphics[width=\linewidth]{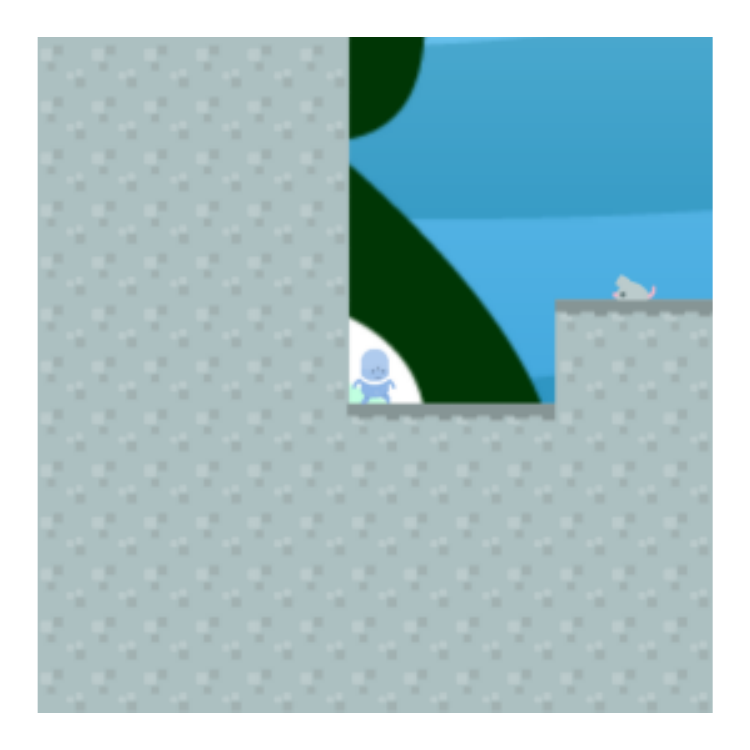}
    \caption{\texttt{coinrun}}
    \label{subfig:coinrun}
  \end{subfigure}
  ~
  \begin{subfigure}[b]{0.18\linewidth}
    \centering
    \includegraphics[width=\linewidth]{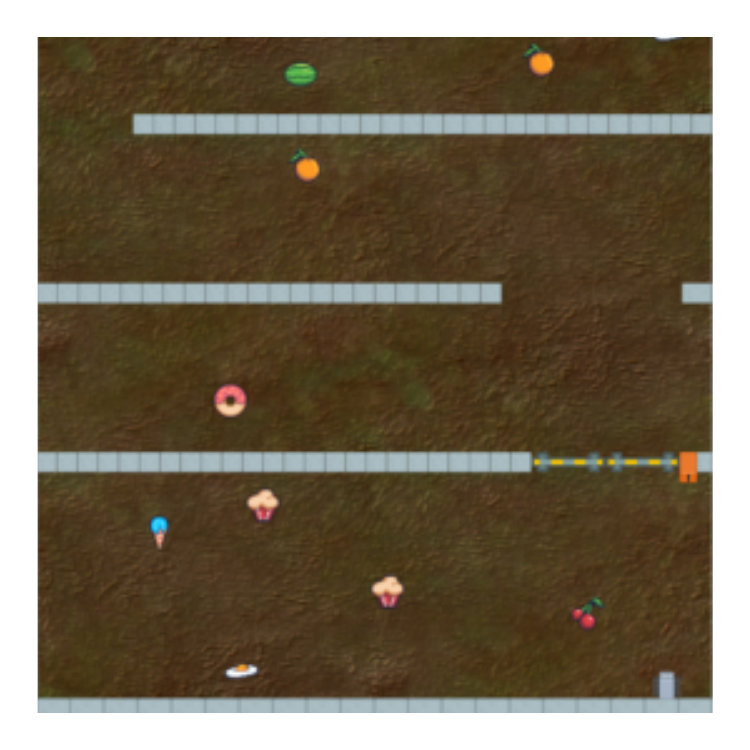}
    \caption{\texttt{fruitbot}}
    \label{subfig:fruitbot}
  \end{subfigure}
  ~
  \begin{subfigure}[b]{0.18\linewidth}
    \centering
    \includegraphics[width=\linewidth]{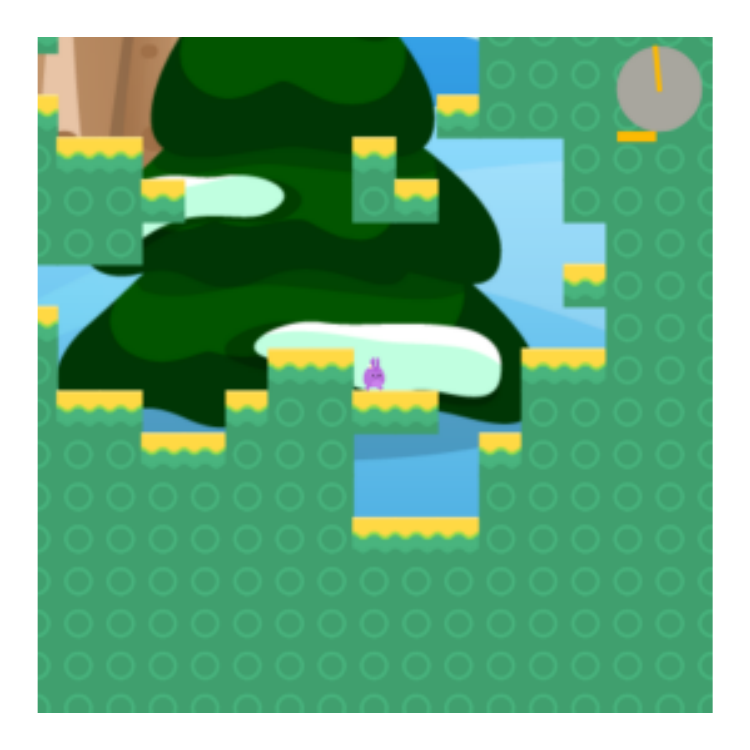}
    \caption{\texttt{jumper}}
    \label{subfig:jumper}
  \end{subfigure}
  \caption{
    \procgen~tasks.
  }
  \label{fig:procgen-envs}
\end{figure*}

We used 5 tasks from the Procgen~\citep[\procgen,][]{procgen19} suite, shown at Figure~\ref{fig:procgen-envs}.
We used the default settings for the environments and we only varied the number of training levels used for learning, which we term \levels.
The tasks are generally partially-observed (POMDPs) specified by:
\begin{enumerate}[noitemsep]
  \item \textbf{Observation space, $\mathcal{O}$:}
    The original $64 \times 64$ RGB pixel-observations, i.e., $o_t \in \left[ 0, 1 \right]^{64 \times 64 \times 3}$.
  \item \textbf{Action space, $\mathcal{A}$:}
    The original 15 discrete actions, i.e., $a_t \in \{0, \ldots, 14\}$.
\end{enumerate}

\clearpage
\subsubsection{MinAtar~\citep{young2019minatar}}
\label{app:minatar}

\begin{figure*}[!h]
  \centering
  \begin{subfigure}[b]{0.18\linewidth}
    \centering
    \includegraphics[width=\linewidth]{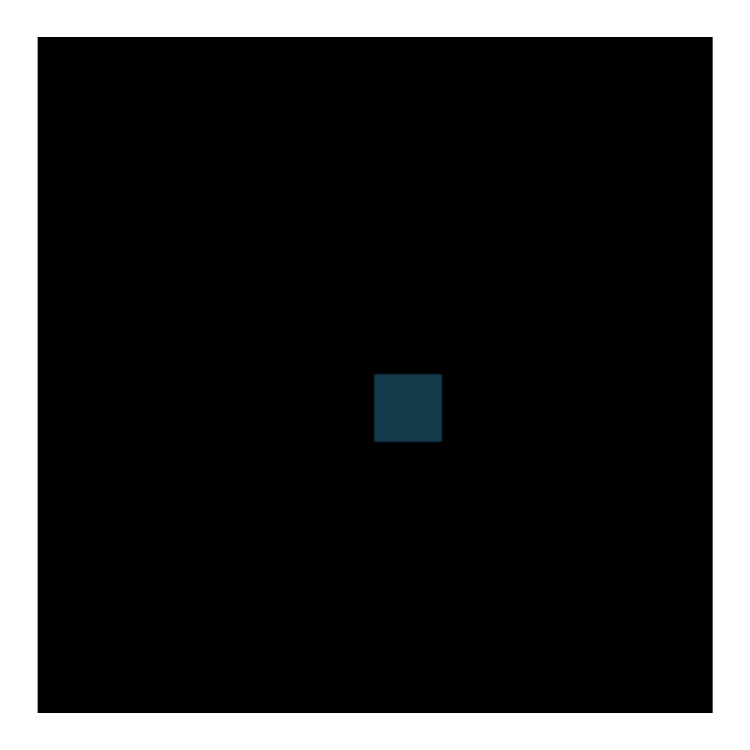}
    \caption{\texttt{asterix}}
    \label{subfig:asterix}
  \end{subfigure}
  ~
  \begin{subfigure}[b]{0.18\linewidth}
    \centering
    \includegraphics[width=\linewidth]{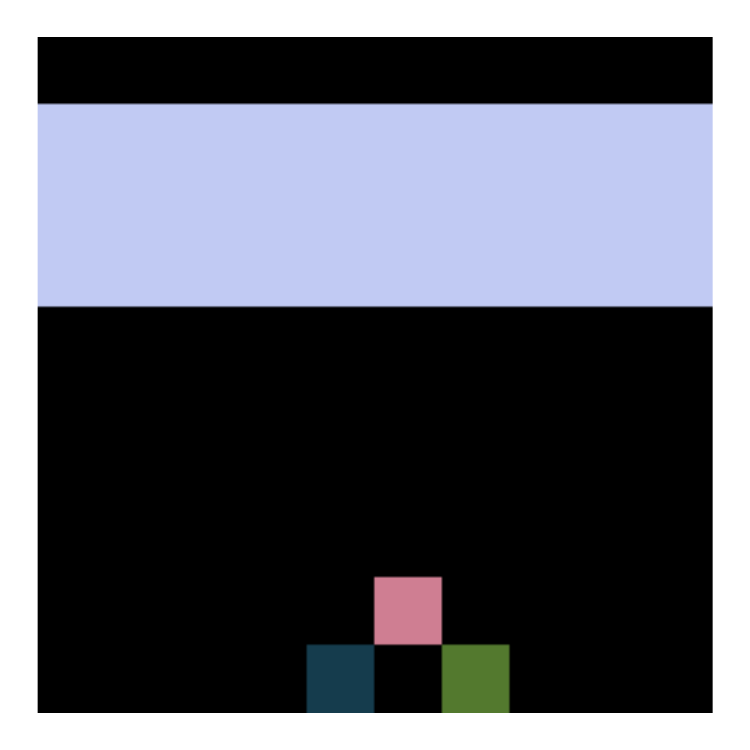}
    \caption{\texttt{breakout}}
    \label{subfig:breakout}
  \end{subfigure}
  ~
  \begin{subfigure}[b]{0.18\linewidth}
    \centering
    \includegraphics[width=\linewidth]{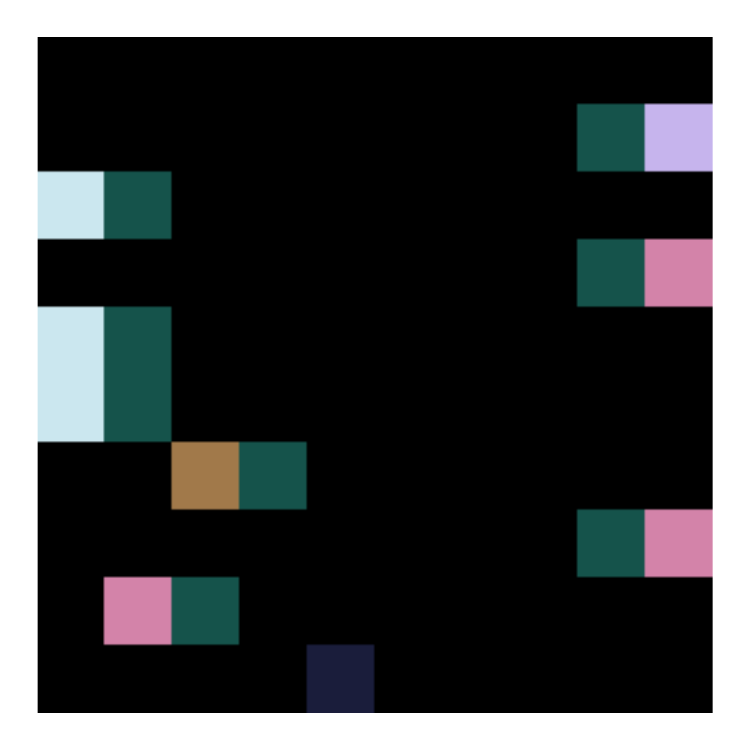}
    \caption{\texttt{freeway}}
    \label{subfig:freeway}
  \end{subfigure}
  ~
  \begin{subfigure}[b]{0.18\linewidth}
    \centering
    \includegraphics[width=\linewidth]{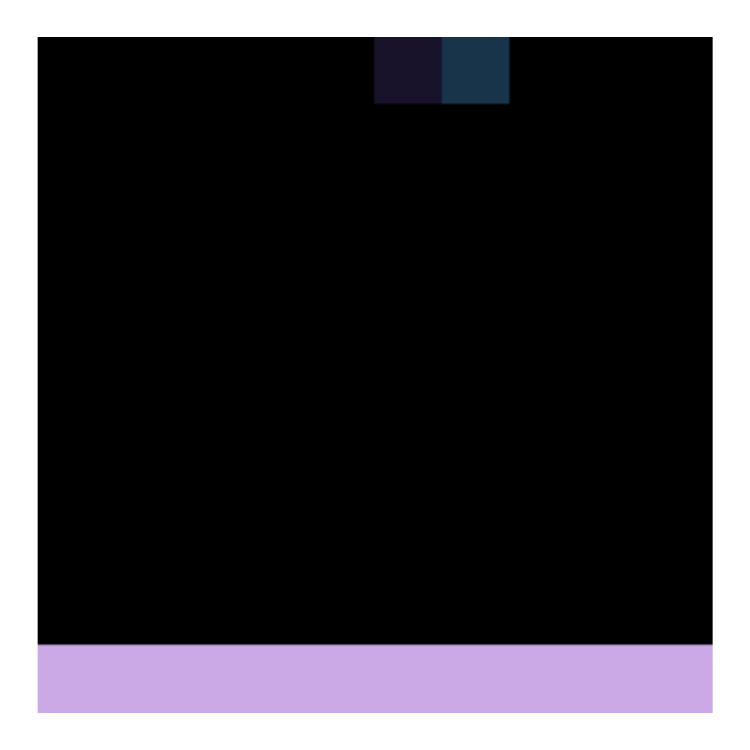}
    \caption{\texttt{seaquest}}
    \label{subfig:seaquest}
  \end{subfigure}
  ~
  \begin{subfigure}[b]{0.18\linewidth}
    \centering
    \includegraphics[width=\linewidth]{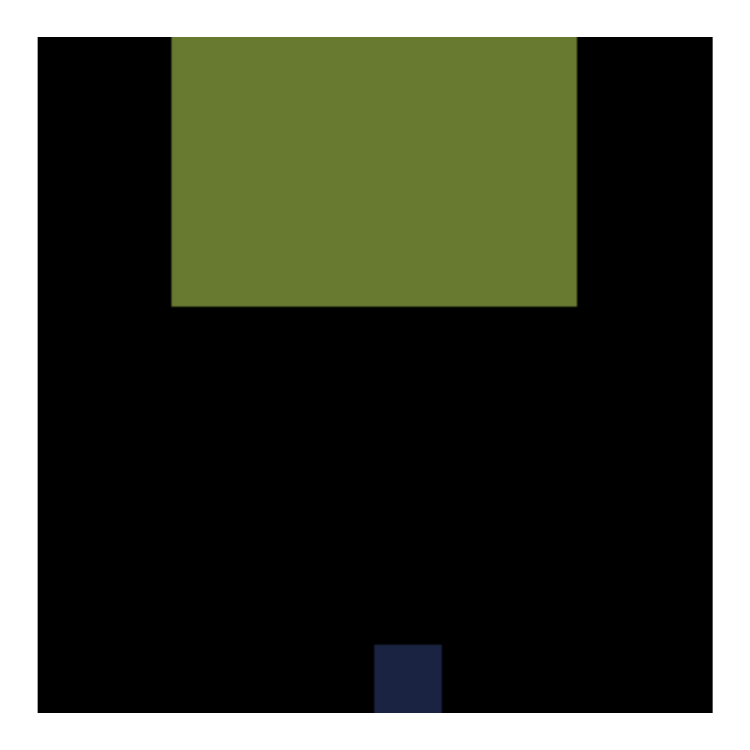}
    \caption{\texttt{space\_invaders}}
    \label{subfig:space_invaders}
  \end{subfigure}
  \caption{
    \minatar~environments.
  }
  \label{fig:minatar-envs}
\end{figure*}

We used all 5 tasks from the MinAtar~\citep[\minatar,][]{young2019minatar} suite, shown in Figure~\ref{fig:minatar-envs}, with the default settings.
The tasks are fully-observed and specified by:
\begin{enumerate}[noitemsep]
  \item \textbf{State space, $\mathcal{S}$:}
    The original $10 \times 10 \times \texttt{n\_channels}$ symbolic observations, i.e., $s_t \in \left[ 0, 1 \right]^{10 \times 10 \times \texttt{n\_channels}}$, where \texttt{n\_channels} varies between tasks, from $4$ to $10$.
  \item \textbf{Action space, $\mathcal{A}$:}
    The original 6 discrete (non-minimal) actions, i.e., $a_t \in \{0, \ldots, 5\}$.
  \item \textbf{Transition dynamics, $p(s' | s, a)$:}
    The default $0.1$ probability for sticky actions is used.
\end{enumerate}

\subsubsection{DeepMind Continuous Control~\citep{tunyasuvunakool2020dm_control}}
\label{app:dm-control}

\begin{minipage}{0.75\textwidth}
  We use the \walker~walk task from the DeepMind Continuous Control~\citep{tunyasuvunakool2020dm_control} suite and modified its reward function.
  Pixel-observations are used, and the problem is generally partially-observed.
  The task is specified by:
  \begin{enumerate}[noitemsep]
    \item \textbf{Observation space, $\mathcal{S}$:}
      A $64 \times 64$ RGB pixel-observation, where the robot body is in the centre of the frame, i.e., $o_t \in \left[ 0, 1 \right]^{64 \times 64 \times 3}$.
    \item \textbf{Action space, $\mathcal{A}$:}
      A six-dimensional continuous action, i.e., $a_{t} \in \left[ -1, +1 \right]^{6}$.
    \item \textbf{Reward function, $r(s, a)$:}
      Originally, the reward is bounded in $[0, 1]$, i.e., $r_{t} \in [0, 1]$, which is computed based on the robot’s torso height and forward velocity.
      We modify the original per-step reward, by setting to zero any reward below a parameter $\eta$, i.e., $\tilde{r}_{t} = \mathcal{H}(r_{t} - \eta)r_{t}$, where $\mathcal{H}$ is the Heaviside step function.
      For $\eta = 0$, we recover the original reward, and for $\eta > 0$ we obtain an increasingly more difficult, in terms of exploration, \walker~task.
  \end{enumerate}
\end{minipage}
\hspace{1.0em}
\begin{minipage}{0.20\textwidth}
  \centering
  \includegraphics[width=\linewidth]{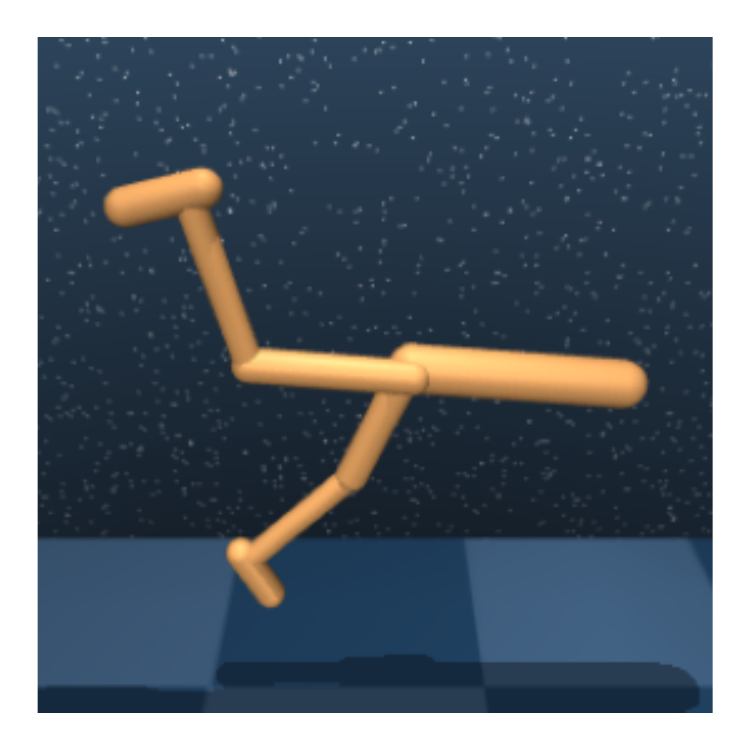}
  \captionof{figure}{\walker}
\end{minipage}

\subsection{Experiments}
\label{subapp:experimens}

In this section, we provide details on the experimental protocol we follow for each experiment.

\subsubsection{Figures~\ref{fig:ive-cg-gp} and~\ref{fig:ive-gp-full}}
\label{subsubapp:ive-gp}

We focus on the prediction problem~\citep{sutton2018reinforcement}, modelled as a Markov reward process (MRP) with an one-dimensional state space, i.e., $s \in \mathcal{S} = [-3, +3]$ and a discount factor $\gamma=0.9$.
We are provided with state-value target pairs, i.e., $\{(s_{i}, \bar{v}_{i})\}_{i=1}^{N}$ with $N=10$ and learn (i) a representation function $\hhat(s; \omega)$, (ii) a value function $\vhat(z; \phi)$ and (iii) a model $\mhat(\cdot, \cdot | z; \theta) \triangleq (\rhat(z; \theta), \phat(z; \theta))$, represented as neural networks with parameters, $\omega$, $\phi$ and $\theta$, respectively.
In particular:
\begin{align}
  \hhat_{\omega}(s)   =   \hhat(s; \omega)    &= \texttt{tanh}(\text{MLP}_{\omega}(s))  \triangleq z        \in [-1, +1]^{32} \label{eq:figure-1-h} \\
  \vhat_{\phi}(z)   =   \vhat(z; \phi)    &= \text{MLP}_{\phi}(z)                 \triangleq v        \in \mathbb{R}\\
  \phat_{\theta}(z) =   \phat(z; \theta)  &= \text{LSTM}_{\theta}(z, \mathbf{0})  \triangleq z^{1}    \in [-1, +1]^{32} \\
  \rhat_{\theta}(z) =   \rhat(z; \theta)  &= \text{MLP}_{\theta}(z)               \triangleq r^{1}        \in \mathbb{R}, \label{eq:figure-1-p}
\end{align}
where all the multi-layer percepetrons (MLPs) have one hidden layer of 32 units with an ELU~\citep{clevert2015fast} non-linearity and $z^{k}$ is the (latent) state after taking $k$ steps with the model $\mhat$, starting from state $z^{0} \triangleq z$~\citep{silver2017predictron}.
\looseness=-1

We make value prediction by repeatedly applying the $\mhat$ model-induced Bellman operator $\T_{\mhat}$ on the value function $\vhat$, i.e., constructing different $k$-steps model predicted values ($k$-MPVs, Eqn.~(\ref{eq:k-mpv})).
In particular, the predictions are given by:
\begin{align}
  \vhat_{\mhat}^{0}(s)  &= (\T_{\mhat_{\theta}})^{0} \vhat_{\phi} (\hhat_{\omega}(s)) = \vhat_{\phi}(\hhat_{\omega}(s)) = \vhat_{\phi} \circ \hhat_{\omega} (s) \label{eq:0-mpv-full} \\
  \vhat_{\mhat}^{1}(s)  &= (\T_{\mhat_{\theta}})^{1} \vhat_{\phi} (\hhat_{\omega}(s)) = \left( \rhat_{\theta} + \gamma  \vhat_{\phi} \right) \circ \phat_{\theta} \circ \hhat_{\omega}(s) \\
  \vhat_{\mhat}^{2}(s)  &= (\T_{\mhat_{\theta}})^{2} \vhat_{\phi} (\hhat_{\omega}(s)) = \left( \rhat_{\theta} + \gamma  (\rhat_{\theta} + \gamma  \vhat_{\phi}) \circ \phat_{\theta} \right) \circ \phat_{\theta} \circ \hhat_{\omega}(s) \\
  &\;\;\vdots \nonumber \\
  \vhat_{\mhat}^{k}(s)  &= (\T_{\mhat_{\theta}})^{k} \vhat_{\phi} (\hhat_{\omega}(s)) = \left( \sum_{j=1}^{k-1} \gamma^{j-1} \rhat_{\theta} \circ \underdescribe{(\phat_{\theta} \circ \cdots \circ \phat_{\theta})}{j\text{-times}} \ + \  \gamma^{k}  \vhat_{\phi} \circ \underdescribe{(\phat_{\theta} \circ \cdots \circ \phat_{\theta})}{k\text{-times}} \right) \circ \hhat_{\omega}(s) \label{eq:k-mpv-full}
\end{align}
where $\circ$ denotes function composition.
Obviously, the $k$-MPVs with different $k$ have different functional forms, as the predictions at initialisation suggest at Figures~\ref{subfig:ive-prior} and~\ref{subfig:ive-prior-appendix}, too.
Note that we do \emph{not} use the \textbf{bold} notation introduced in Eqn.~(\ref{eq:mc-estimator}) to highlight that there is no Monte Carlo sampling---the learned model is deterministic and the policy is implicit.

We learn the neural network parameters $\omega$, $\phi$ and $\theta$ using the ADAM~\citep{kingma2014adam} optimiser with decoupled weight decay~\citep{loshchilov2017decoupled} to minimise the empirical squared value prediction error for all $k \in \{0, \ldots, 10\}$, i.e.,
\begin{align}
  \min_{\omega, \theta, \phi} \ \sum_{i=1}^{N} \sum_{k=0}^{K} \| \vhat_{\mhat}^{k}(s_{i}) - v_{i} \|_{2}^{2}.
\end{align}
The only source of variability between the $k$-MPVs (implicit value ensemble (IVE) members) is their functional form, induced by different compositions of the learned parametric networks $\vhat_{\phi}$, $\phat_{\theta}$ and $\rhat_{\theta}$ as Eqns.~(\ref{eq:0-mpv-full}-\ref{eq:k-mpv-full}) show.

\input{figures/ive-gp}

\subsubsection{Figure~\ref{fig:std-value-ensembles}}

\paragraph{Data.}
We collect experience/data $\B$ by running a uniformly random policy $\pi_{\text{uniform}}$ for $500$ steps (i.e., 25 episodes).
We exclude transitions from and to the top left cell, which we call the \emph{out-of-distribution} (OOD) or unvisited state.

\paragraph{Value learning.}
We learn a tabular action-value function $\hat{q} \approx q^{\pi_{\text{uniform}}}$ using expected SARSA~\citep{van2009theoretical} and then we induce a (state-)value function, i.e., $\hat{v}(s) \triangleq \E_{a \sim \pi_{\text{uniform}}}[ \hat{q}(s, a) ], \forall (s, a) \in \mathcal{S} \times \mathcal{A}$.

\paragraph{Model learning.}
Maximum-likelihood estimation (MLE, see Section~\ref{sec:background}) with data $\B$ is used for learning the tabular model of the environment $\hat{m} \approx m^{*}$.

\paragraph{Visualisations.}
The mean and standard deviations are normalised in $[0, 1]$, i.e., for given quantity $x_{s}$ for state $s$ and $x_{\text{min}}$ ($x_{\text{max}}$ the minimum (maximum) quantity across all states, we plot $\bar{x}_{s} = (x_{s} - x_{\text{min}}) / (x_{\text{max}} - x_{\text{min}})$.
We report the results for a single repetition of the experiment since it is a qualitative observation.

\paragraph{IVE($n$).}
We calculate the MPVs exactly, according to Eqn.~(\ref{eq:k-mpv}).
We vary the parameter $n$, i.e., maximum number of applications of the model-induced Bellman operator $\T_{\mhat}$ on the learned value function $\vhat$.

\paragraph{EVE($n$) and EMVE($n$).}
The explicit value ensemble (EVE, Figure~\ref{subfig:vensemble-net}) and explicit model value ensemble (EMVE, Figure~\ref{subfig:mensemble-net}) are also trained on the same data using the same value and model learning algorithms.
The ensemble components different only in their (random) initialisation and seed used in stochastic gradient descent.

\subsubsection{Table~\ref{tab:dreamer-mvsic-walker-walk}}

We use the \walker~task and train Dreamer~\citep{hafner2019dream} for 1M steps.
An action repeat of $2$ is used thus 0.5M agent-environment interaction steps are made per run.
We repeat each experiment 3 times, varying the random seed in each one.
We report the episodic returns  (rounded to the nearest tenth) at the end of training by setting the agents in ``evaluation'' mode and average their performance across 10 episodes.

\subsubsection{Figure~\ref{fig:procgen-probing}}

We train Muesli (without any modification to its acting strategy or learning algorithm) for $100$M environment frames.
Figure~\ref{fig:procgen-probing} (left) reports the final performance of the agent evaluated on an additional 10M frames on the train and test levels.
Mean episode returns are normalised as: $\tilde{R} = (R-R_{\text{min}})/(R_{\text{max}}-R_{\text{min}})$, using min and max scores for each game~\citep{procgen19}.

The model-value self-inconsistency, reported in Figure~\ref{fig:procgen-probing} (right), is computed by unrolling the model for $5$ steps using actions sampled from the policy and taking the standard deviation over the IVE:
\begin{align}
    k\text{-MVP}(s) &= \mathbf{\vhat_{\mhat}^{k}}(s) \overset{(\ref{eq:mc-estimator})}{=} \sum_{i=1}^{k-1} \gamma^{i-1}\mathbf{r_{\mhat}^{i+1}} + \gamma^{k}\vhat(\mathbf{s_{\mhat}^{k}})  \label{eq:ive_mc} \\
  \sigma\text{-IVE}(s) &= \text{std}_k[k\text{-MVP}(s)], \ \text{for } k=1,\dots, 5
\end{align}
where the ``bold'' notation refers to reward and value predictions given a single action sequence sampled from the policy $\pi$, as described in Eqn.~(\ref{eq:mc-estimator}).

\subsubsection{Figure~\ref{fig:reaching-ood-states}}

For training the values and model and calculating IVE and EVE, we follow the same protocol as in Figure~\ref{fig:std-value-ensembles}.
In this experiment, we use the learned \emph{action}-value functions instead of the state-values, see Section~\ref{subsec:mpv-with-action-value-functions} for a formal discussion.
We denote with $\sigma$-IVE($5$) and $\sigma$-EVE($5$) the standard deviation across the $5$ ensemble members of the implicit and explicit ensembles of the action-values, respectively.
Also, $\sigma$-IVE($5$) $\in \mathbb{R}^{\mathcal{S} \times \mathcal{A}}$ and $\sigma$-IVE($5$)$[s, a]$ is the standard deviation of the implicit value ensemble at the state $s$ for action $a$.
We use the standard deviation across the ensemble of action-values for inducing policies that are novelty- seeking or avoiding:
\begin{itemize}[noitemsep,topsep=0pt, labelwidth=!, labelindent=0pt]
  \item In Figure~\ref{subfig:prob_reaching_ood_state_ofu}, the action that maximises the standard deviation across the value ensemble is selected, per-state, i.e., $\pi_{\text{seeking}}(s) = \argmax_{a \in \mathcal{A}} \sigma\text{-XVE}(5)[s, a]$, where XVE $\in \{$IVE, EVE$\}$.
  These are the novelty-seeking policies that their probability of reaching the novel state is higher than a uniformly random policy.
  \item In Figure~\ref{subfig:prob_reaching_ood_state_pfu}, the action that minimises the standard deviation across the value ensemble is selected, per-state, i.e., $\pi_{\text{avoiding}}(s) = \argmin_{a \in \mathcal{A}} \sigma\text{-XVE}(5)[s, a]$, where XVE $\in \{$IVE, EVE$\}$.
  These are the novelty-avoiding policies that their probability of reaching the novel state is lower than a uniformly random policy.
\end{itemize}
We calculate the probabilities by constructing a Markov chain, induced by the coupling of the policy under consideration $\pi$ and the ``true'' environment model, $\mstar$.
The Markov chain's transition kernel is given by $p_{\mstar}^{\pi}(s' | s) \triangleq \sum_{a \in \mathcal{A}} p_{\mstar}(s' | s, a) \pi(a | s)$.
We can write the transition kernel as a matrix $P_{\mstar}^{\pi} \in \mathbb{R}^{\mathcal{S} \times \mathcal{S}}$, such that $P_{\mstar}^{\pi}[i, j] = p_{\mstar}^{\pi}(j | i)$.
The $(i, j)$ entry of the transition matrix, i.e., $P_{\mstar}^{\pi}[i, j]$ is the probability of reaching the state $j$ after one-step when starting from state $i$ and following policy $\pi$ in the environment with model $\mstar$.
The $(i, j)$ entry of the $l$-th power of the transition matrix, i.e., $(P_{\mstar}^{\pi})^{l}[i, j]$ is the probability of reaching the state $j$ after $l$-steps when starting from state $i$ and following policy $\pi$ in the environment with model $\mstar$.

In Figure~\ref{fig:reaching-ood-states}, we start from the bottom right cell, i.e., $i=\texttt{bottom right}$ and plot the probability of reaching the top left cell, i.e., $j=\texttt{top right}$ after $l$-steps, and we vary $l$ from $1$ to $150$.
We repeat each experiment 100 times, varying the random seed in each one.

\subsection{Table~\ref{tab:vpn-ive-minatar}}
We use the \minatar~tasks and train VPN~\citep{oh2017value} and some variants of it for 2M steps.
The only modification to the original VPN(5) is the way value estimates are constructed:
\begin{itemize}[noitemsep,topsep=0pt, labelwidth=!, labelindent=0pt]
  \item $\vhat_{\mhat}^{1}$ is VPN variant that uses the $1$-MPV for value estimation.
  \item $\vhat_{\mhat}^{5}$ is VPN variant that uses the $5$-MPV for value estimation.
  \item $\mu$-IVE($5$) is the original VPN(5) agent that uses the mean over the implicit value ensemble with $n=5$ for value estimation.
\end{itemize}
The estimated values are used for value-based planning, as discussed in~\citep[Eqn. (1) \& Appendix D.]{oh2017value}.\looseness=-1

%% file: figures/ive-gp.tex
\begin{figure}
  \centering

  \includegraphics[width=0.5\linewidth]{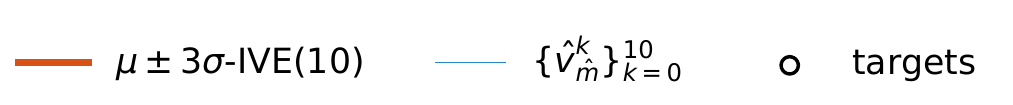}

  \begin{subfigure}[b]{0.30\linewidth}
    \centering
    \includegraphics[width=\linewidth]{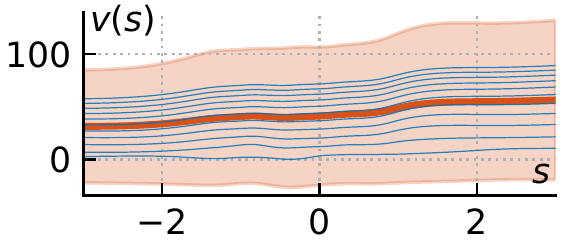}
    \caption{At initialisation (i.e., \emph{prior})}
    \label{subfig:ive-prior-appendix}
  \end{subfigure}
  ~
  \begin{subfigure}[b]{0.30\linewidth}
    \centering
    \includegraphics[width=\linewidth]{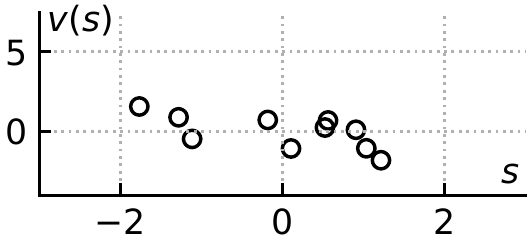}
    \caption{Value targets (i.e., \emph{data})}
    \label{subfig:value-targets-appendix}
  \end{subfigure}
  ~
  \begin{subfigure}[b]{0.30\linewidth}
    \centering
    \includegraphics[width=\linewidth]{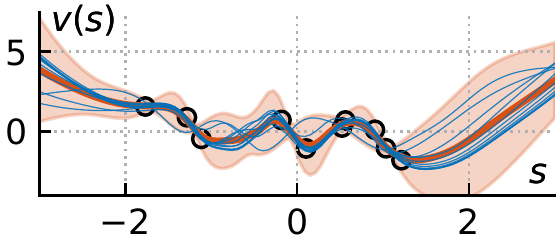}
    \caption{After training (i.e., \emph{posterior})}
    \label{subfig:ive-posterior-appendix}
  \end{subfigure}
  \caption{
    Expanded version of Figure~\ref{fig:ive-cg-gp}.
    A value prediction problem of an implicit policy, modelled as a Markov reward process~\citep[MRP,][]{sutton2018reinforcement} with an one-dimensional state space, i.e., $s \in \mathcal{S}=[-3, 3]$.
    We learn a model $\mhat$ and a value function $\vhat$ and construct a $k$-steps model predicted value ($k$-MPV, Eqn.~(\ref{eq:k-mpv})) by applying the model induced Bellman operator $\T_{\mhat}$ repeatedly $k$ times on the learned value function $\vhat$, i.e., $\vhat_{\mhat}^{k}(s) \triangleq (\T_{\mhat})^{k}\vhat(s)$.
    We visualise the $k$-MPVs, a.k.a components of the \emph{implicit value ensemble} (IVE, Eqn.~(\ref{eq:ive})) for $k \in \{0, \ldots, 10\}$ (in \textcolor{mlblue}{blue}) along with the ensemble mean and standard deviation (in \textcolor{mlorange}{orange}), constructed from a single (point) estimate of the value function and model. 
    (a) The predictions at initialisation, i.e., before training.
    (b) The data, i.e., state and value target pairs.
    (c) The predictions after training every IVE member towards the value targets in (b), i.e., $\min_{m, v} \sum_{i} \sum_{k} \| \vhat_{\mhat}^{k}(s_{i}) - v_{i} \|_{2}^{2}$.
    We observe in (c) that the ensemble components fit the value targets and their standard deviation is zero at and around the observed (in-distribution) data but it is non-zero otherwise (out-of-distribution points).
    Therefore the IVE members' disagreement can be used as a signal for epistemic uncertainty.
    In this example, the variability between the IVE members' predictions is only due to their different functional forms.
    \looseness=-1
  }
  \label{fig:ive-gp-full}
\end{figure}

%% file: sections/B_implementation_details.tex
\section{Implementation Details}
\label{app:implementation-details}

For our experiments we used Python~\citep{van1995python}.
We used JAX~\citep{jax2018github,deepmind2020jax} as the core computational library for implementing Muesli~\citep{hessel2021muesli} and VPN~\citep{oh2017value}.
We used the official TensorFlow~\citep{abadi2016tensorflow} implementation of Dreamer~\citep{hafner2019dream}.
We also used Matplotlib~\citep{hunter2007matplotlib} for the visualisations.

\subsection{Tabular Methods}
\label{subapp:tabular-methods}

We initialise the rewards, transition logits and action-values by sampling from a normal distribution with mean $0$ and standard deviation $1$.
The ADAM~\citep{kingma2014adam} optimiser with learning rate $5e\text{-}5$ is used, and all losses converge after $10,000$ epochs of stochastic gradient descent with batch size $128$.

\subsection{Dreamer~\citep{hafner2019dream}}

We use the Dreamer agent's default hyperparameters, as introduced by~\citep{hafner2019dream}.
For the self-inconsistency-seeking variant, i.e., $\mu+\sigma$-IVE(5), we used a scalar weighting factor $\beta_{\text{IVE}}^{*}=0.1$ to balance the mean and standard deviation across the ensemble members, tuned with grid search in $\{0.05, 0.1, 0.2, 1.0, 10.0 \}$.
The same tuning procedure is used for the baselines.
The reported scores are for $\beta_{\text{EVE}}^{*}=0.2$ and $\beta_{\text{EMVE}}^{*}=0.1$.

\subsection{Muesli~\citep{hessel2021muesli}}

We use the Muesli agent's hyperparameters.
In particular we use the ones from the large-scale Atari experiments by~\citet{hessel2021muesli}.
Nonetheless, we set the fraction of replay data in each batch to $0.8$ (instead of the original $0.95$) to shorten training time.
To encourage diversity in value and reward predictions for unvisited states we have augmented the value and reward prediction heads of the model with untrainable \textit{randomized prior networks} \citep{osband2018randomized}, using a prior scale of $5.0$.
Note that unlike in \citet{osband2018randomized}, we did not introduce additional heads per prediction or modify the training procedure.
\looseness=-1

\subsection{VPN~\citep{oh2017value}}

We use the MinAtar DQN-torso~\citep{young2019minatar} and an LSTM~\citep{hochreiter1997long} with 128 hidden units and otherwise follow the original VPN(5) hyperparameters, as introduced by~\citet{oh2017value}.

%% file: sections/C_extensions.tex
\section{Extensions}
\label{app:extensions}

\subsection{IVE with the Bellman Optimality Operator}
\label{subsec:ive-with-the-bellman-optimality-operator}

In Section~\ref{sec:your-model-based-agent-is-secretly-an-ensemble-of-value-functions}, we defined the $k$-steps model-predicted value ($k$-MPV) in terms of the model-induced Bellman evaluation operator and a value function for a policy $\pi$, and constructed the implicit value ensemble (IVE) accordingly.
In this section. we provide a brief presentation of MPVs and IVE in terms of the model-induced Bellman optimality operator and optimal value functions.

\begin{definition}[Bellman optimality operator]
  Given the model $\mstar$, the one-step Bellman optimality operator $\T^{*}: \mathbb{V} \rightarrow \mathbb{V}$ is induced, and its application on a state-function $v \in \mathbb{V}$, for all $s \in \mathcal{S}$, is given by
  \begin{align}
    \T^{*} v(s) \triangleq \max_{a \in \mathcal{A}} \E_{\mstar} \left[ R_{0} + \gamma v(S_{1}) \mid S_{0}=s. A_{1}=a \right].
  \label{eq:bellman-optimality-one-step}
  \end{align}
\end{definition}

The $k$-times repeated application of an one-step Bellman optimality operator gives rise to the $k$-steps Bellman optimality operator,
\begin{align}
  (\T^{*})^{k} v \triangleq \underdescribe{\T^{*} \ldots \T^{*}}{k\text{-times}} v .
\label{eq:bellman-optimality-k-steps}
\end{align}

The Bellman optimality operator, $\T^{*}$, is a contraction mapping~\citep{puterman2014markov}, and its fixed point is the value of the optimal policy $\pi^{*}$, i.e., $\lim_{n \rightarrow \infty} (\T^{*})^{n} v = v^{\pi^{*}} \triangleq v^{*}$, for any state-function $v \in \mathbb{V} \triangleq \{ f: \mathcal{S} \rightarrow \mathbb{R} \}$.

\paragraph{Model-induced Bellman optimality operator.}
A model $\mhat$ induces a Bellman optimality operator $\T_{\mhat}^{*}$ with a fixed point $v_{\mhat}^{*}$, i.e., the value of the optimal policy \emph{under the model} (a.k.a. the solution of the model.
Similar to Eqn.~(\ref{eq:bellman-optimality-k-steps}), a $k$-steps model-induced Bellman optimality operator is given by $(\T_{\mhat}^{*})^{k} v = \underdescribe{\T_{\mhat}^{*}\ldots\T_{\mhat}^{*}}{k\text{-times}} v$.

\paragraph{Model-predicted values.}
The $k$-steps MPV, using the model-induced Bellman optimality operator is given by
\begin{align}
  \vhat_{\mhat}^{k} \triangleq (\T_{\mhat}^{*})^{k} \vhat
\end{align}

\paragraph{Implicit value ensemble.}
An ensemble of $k$-MPV predictions can be made by varying $k$, giving rise to
\begin{align}
  \{ \vhat_{\mhat}^{i} \}_{i=0}^{n} \triangleq \underdescribe{ \{ \vhat, \T_{\mhat}^{*} \vhat, \ldots, (\T_{\mhat}^{*})^{n} \vhat \} }{n+1 \text{ value estimates}}.
\end{align}

The IVE with the Bellman optimality operator can be used for values learned with, e.g., Q-learning~\citep{watkins1992q}, or with other value-based agents, e.g., VPN~\citep{oh2017value}.
We use this idea in Appendix~\ref{app:ablations}.

\subsection{MPV~with Action-Value Functions}
\label{subsec:mpv-with-action-value-functions}

In order to be able to modulate action selection using the self-inconsistency signal, we have computed the k-MPV conditioned on both state and action:
\begin{align}
     k\text{-MVP}(s, a) &= \mathbf{\hat{q}_{\mhat}^{k}}(s, a) = \sum_{i=0}^{k-1} \gamma^{i}\mathbf{r_{\mhat}^{i+1}} + \gamma^{k}\vhat(\mathbf{s_{\mhat}^{k}})
     \label{eq:iveq_mc},
\end{align}
where now reward and value predictions are computed after unrolling the model using action $a$ for one step, and actions sampled from the policy for the remaining $k-1$ steps.

%% file: sections/D_ablations.tex
\section{Additional Experiments}
\label{app:ablations}

\subsection{Measuring Self-Inconsistency in OOD States}
\label{sec:ood_100}

\begin{figure}[!h]
    \centering
    \includegraphics[width=0.6\linewidth]{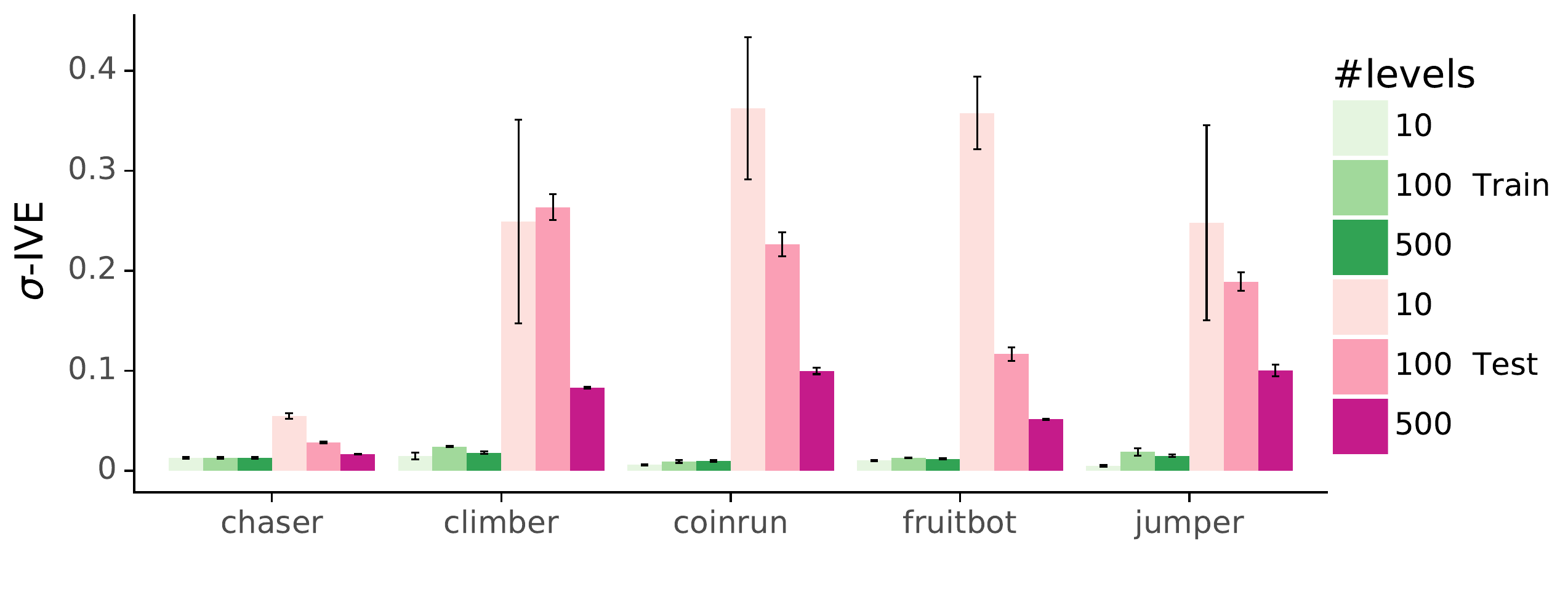}
    \caption{ $\sigma$-IVE($5$) computed using the model of the Muesli agent while evaluating on both training and unseen test levels, for different numbers of unique levels seen during training. To estimate the IVE, we used 100 action sequences from the policy.  Bars, error-bars show mean and standard error across 3 seeds, respectively.}
    \label{fig:procgen_ive100}
\end{figure}

To complement our results in Figure \ref{fig:procgen-probing}, we have also evaluated self-inconsistency by computing the IVE as an average over 100 action sequences sampled from the policy, see Figure \ref{fig:procgen_ive100}. We observed only minor quantitative differences compared to the results presented in Figure \ref{fig:procgen-probing} (where we were using a single action sequence to estimate the IVE). 

\subsection{Measuring Explicit Ensemble (EVE) Variance in OOD States}
\begin{figure}[!h]
    \centering
    \includegraphics[height=3.5cm]{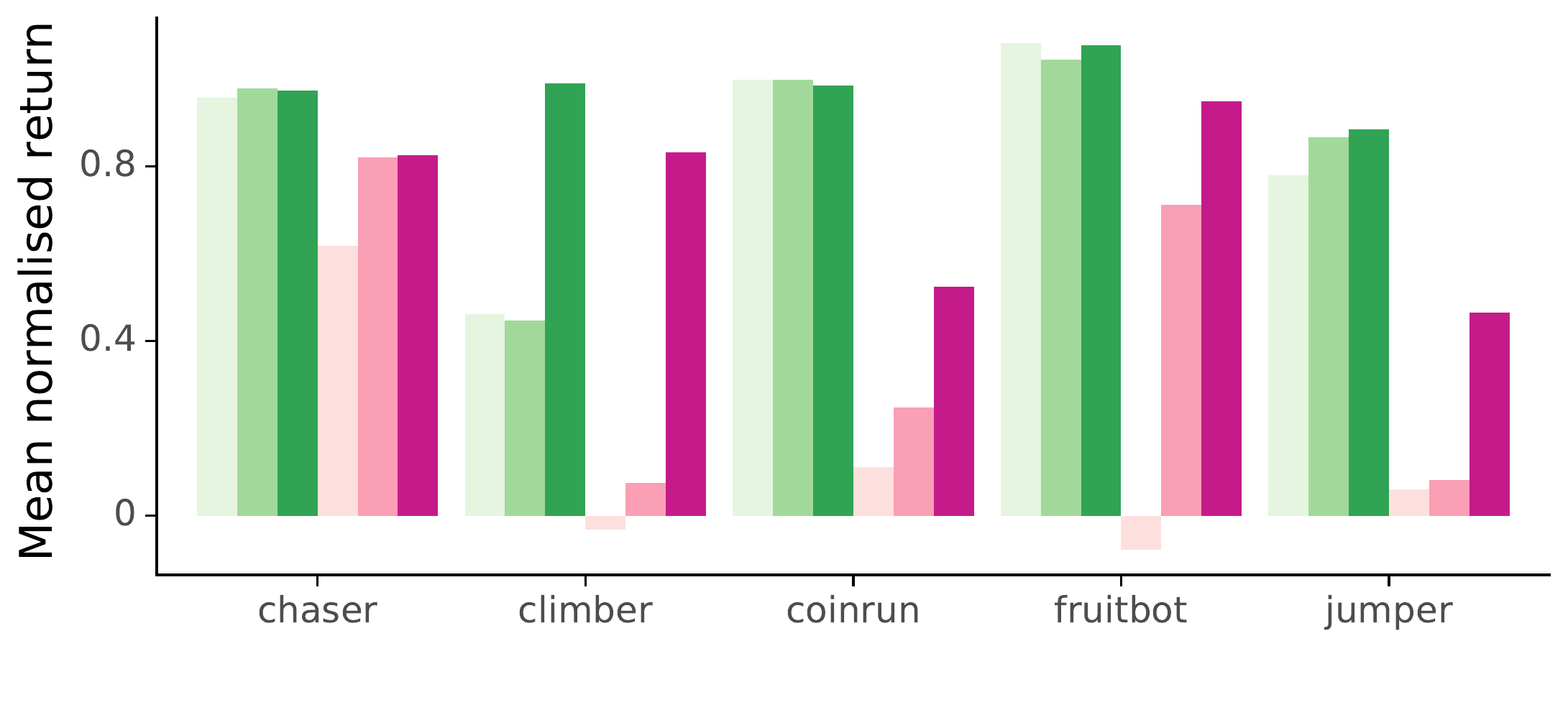}
    \includegraphics[height=3.5cm]{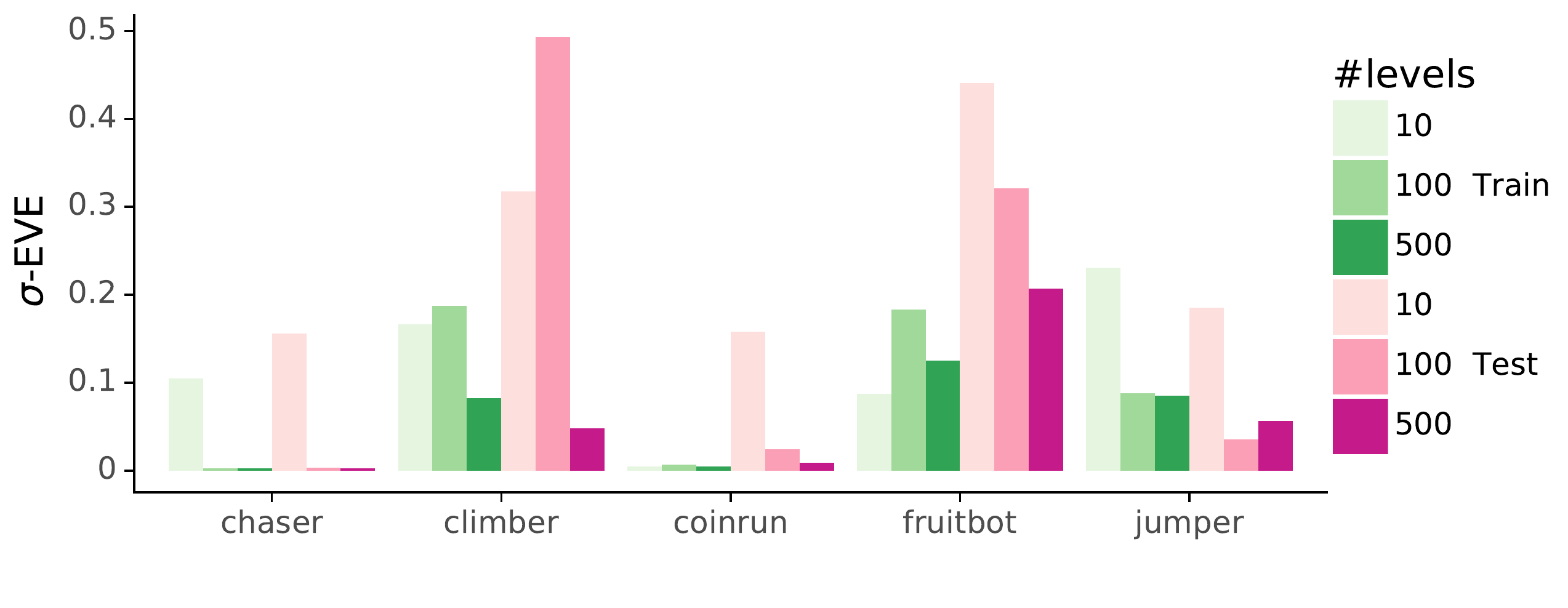}
    \caption{ $\sigma$-EVE($5$) computed using the Muesli agent augmented with an ensemble of 5 value heads (different random initialisation) while evaluating on both training and unseen test levels, for different numbers of unique levels seen during training. (Left:) Performance for training (green) and test (pink) for varying number of levels. (Right:) Explicit value ensemble inconsistency measured by standard deviation of the 5 different heads. Results are from a single seed.}
    \label{fig:procgen_eve5}
\end{figure}

To complement our results in Figure \ref{fig:procgen-probing} for IVE, we provide the EVE results in Figure~\ref{fig:procgen_eve5}.
We observe that EVE behaves similar to IVE in terms of ensemble variance as a function of \levels~for both training and testing levels.

\subsection{How to Use the IVE(5) Signal?}
\label{sec:howto}
In the following experiments we consider the self-inconsistency signal as an optimistic bonus to encourage better exploration during training, hence generalising better during evaluation. We test variants of the $+\sigma$-IVE(5) signal by mixing the policy with the self-inconsistency in probability space $+\sigma$-IVE(5)$\triangleq (1-\beta)\pi+\beta \cdot \sigma$-IVE(5), and by mixing the signal with the policy logits:  $z+\sigma$-IVE(5)$\triangleq \text{softmax}(z_\pi+\beta \cdot \sigma$-IVE(5)$)$. We vary the number of MPV~in the ensemble for $n=5, 10$.
Use further test using a different metric for measuring the disagreement across the n{MPV}s that considers different weighting averages over $k$:

\begin{align}
    d_{JS} = \textsc{JSD}_{\mathbf{w}}(\text{IVE}(n)) = \textsc{H}\left(\sum_k w_k \hat{v}^k_{\hat{m}}\right) - \sum_k w_k \textsc{H}\left(\hat{v}^k_{\hat{m}}\right)
\end{align}
with three weighting schemes: a decreasing weight $dec_{JS}: w_k = r^{k}/(\sum_{j} r^j)$ such that the weight decreases to $1/3$ over $n$, an increasing weight $inc_{JS}$ with the inverse trend, and a uniform weight $uni_{JS}$ that corresponds to the uniform mixing over n $w_k=1/n$.
\input{figures/using-self-inconsistency}

In Figure~\ref{subfig:climbtest} we observe that learning with an optimistic bonus helps with generalisation at evaluation time. Figure\ref{subfig:variants} we observe that mixing over probability space is less sensitive to re-scaling $\beta$, but yields higher variance. We notice a trade-off between the weighting scheme used vs. the size of the IVE, for higher ns the best performing metric has less weight on the larger k-{MPV}s. For the decreasing metric the results remain more robust, suggesting that the inconsistencies are higher for larger ks. We used $\beta=0.1$ for mixing in probability and $\beta=1$ for the logit case.\looseness=-1

\subsection{Ablation on Pessimism for Evaluation}
\label{sec:pefu_beta}
We evaluate in Figure~\ref{fig:pefu_beta} how sensitive the self-inconsistency signal is to different re-scaling parameters $\beta$ when acting pessimistically at test time $z-\beta\ d$-$\sigma$-IVE(5) with an increasing weight. We trained a vanilla Muesli agent using 10/100 levels over 150M frames and evaluated with a pessimistic bonus for the consecutive 20M frames over 3 seeds.
\input{figures/pefu_eval}

\subsection{Dreamer Variants}

\begin{table}[!h]
\begin{minipage}[t]{0.47\textwidth}
  In Section~\ref{subsec:self-inconsistency-as-a-signal-for-exploration}, we modified the Dreamer~\citep{hafner2019dream} agent to improve its exploration without having to learn an explicit ensemble of value functions.
  We modify the behavioural policy used for collecting data, using the mean and standard deviation of the implicit value ensemble, i.e., $\mu$-IVE($5$) and $\sigma$-IVE($5$), respectively.
  We use the original Dreamer setup otherwise.
  \\[0.5em]
  In particular, for each time-step $t$, we sample action $\mathbf{a}_{\pi}^{t}$ from the learned policy $\pi$ and then calculate the IVE($5$), similar to Eqn.~(\ref{eq:mc-estimator}).
  Then, we can form the utility function
  \begin{align}
    \mathcal{U} = \mu\text{-IVE}(5) + \beta \cdot \sigma\text{-IVE}(5).
  \end{align}
  We use online gradient-based or sample-based planning, a.k.a. model-predictive control~\citep[MPC,][]{garcia1989model} for selecting an action.
  \\[0.5em]
  We used $\beta = 0.1$, 10 gradient steps or 10 samples from the learned policy for guiding the search in all of our experiments, shown in Table~\ref{tab:dreamer-walker-walk-full}.
\end{minipage}
\hspace{1.0em}
\begin{minipage}[t]{0.5\textwidth}
  \input{tables/dreamer-walker-walk-full}
\end{minipage}
\end{table}

\clearpage
\subsection{Qualitative Analysis of Different Value Ensembles}

In Figure~\ref{fig:all-value-ensembles}, we plot the standard deviation across different types of value ensembles, as illustrated in Figure~\ref{fig:nets}.

\input{figures/all_value_ensembles}

%% file: figures/using-self-inconsistency.tex
\begin{figure*}[!h]
  \centering
  \begin{subfigure}[b]{0.64\linewidth}
    \centering
    \includegraphics[width=\linewidth]{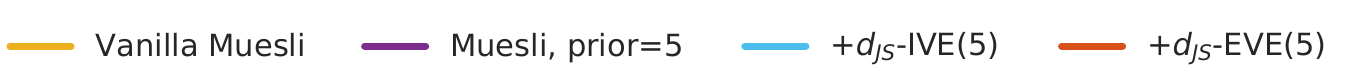}
  \end{subfigure}
  ~
  \begin{subfigure}[b]{0.32\linewidth}
    \centering
    \includegraphics[width=\linewidth]{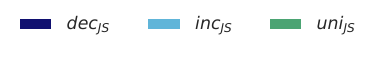}
  \end{subfigure}

  \begin{subfigure}[b]{0.25\linewidth}
    \centering
    \includegraphics[width=\linewidth]{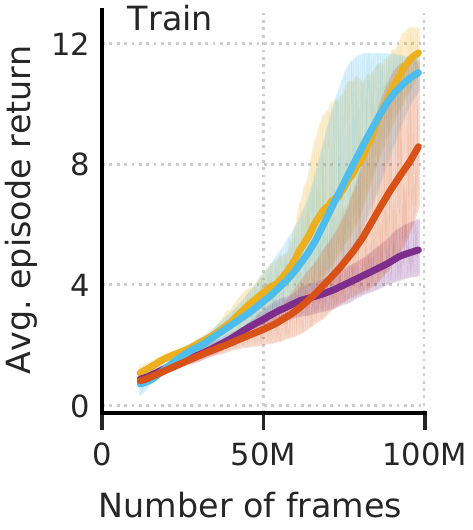}
    \caption{Train}
    \label{subfig:climbtrain}
  \end{subfigure}
  ~
  \begin{subfigure}[b]{0.25\linewidth}
    \centering
    \includegraphics[width=\linewidth]{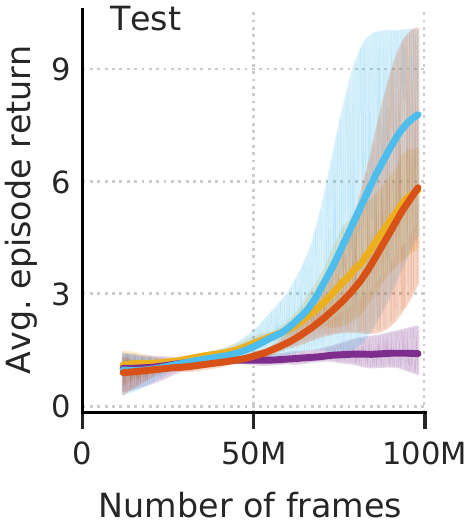}
    \caption{Test}
    \label{subfig:climbtest}
  \end{subfigure}
  ~
  \begin{subfigure}[b]{0.32\linewidth}
    \centering
    \includegraphics[width=\linewidth]{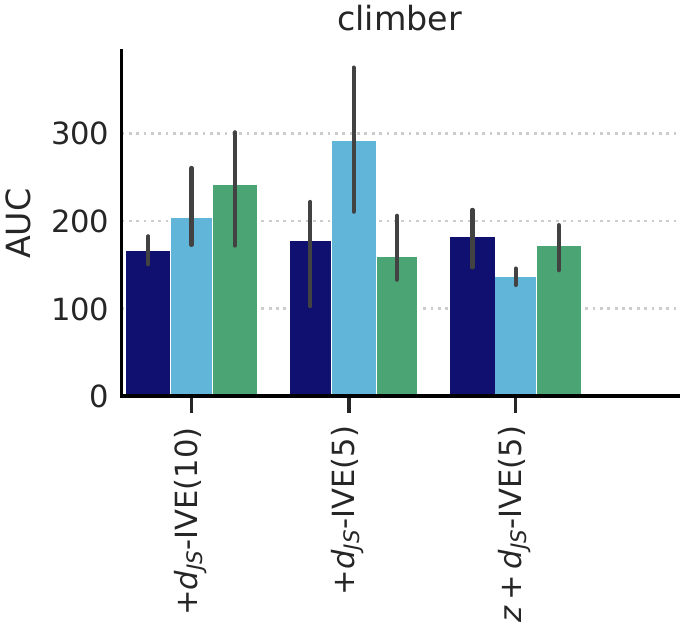}
    \caption{Variations}
    \label{subfig:variants}
  \end{subfigure}
  \caption{
    Model-value inconsistency (see Section~\ref{subsec:model-value-inconsistency}) as the Jensen-Shannon divergence of the implicit value ensemble (see Section~\ref{subsec:implicit-value-esemble}) for different numbers of ensemble components $n$, trained across 100 levels error bars show SE over 3 seeds.
    (a) Mean episode return during training with 100 levels, for Muesli baselines and for an agent trained with optimistic divergence over an explicit ensemble $d_{JS}$-EVE(5) and over IVE(5), both with an increasing Jensen-Shannon disagreement.
    (b) Mean episode return for evaluation without the optimistic disagreement for the same methods. (c) Ablation study over $d_{JS}$-IVE~ of varying length $n=5, 10$ and by mixing in logit space $z+d$-IVE~vs. mixing in probability space $+d$-IVE.
  }
  \label{fig:using-self-inconsistency}
\vspace{-1em}
\end{figure*}

%% file: figures/pefu_eval.tex
\begin{figure*}[!h]
  \centering
  \includegraphics[width=0.6\linewidth]{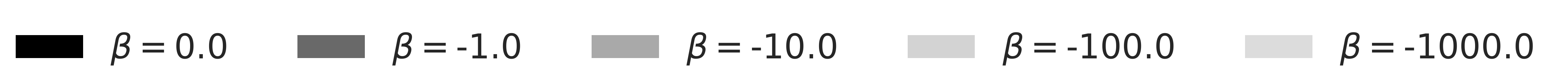}
  
  \includegraphics[width=0.7\linewidth, trim={0, 0, 0, 0.cm}, clip]{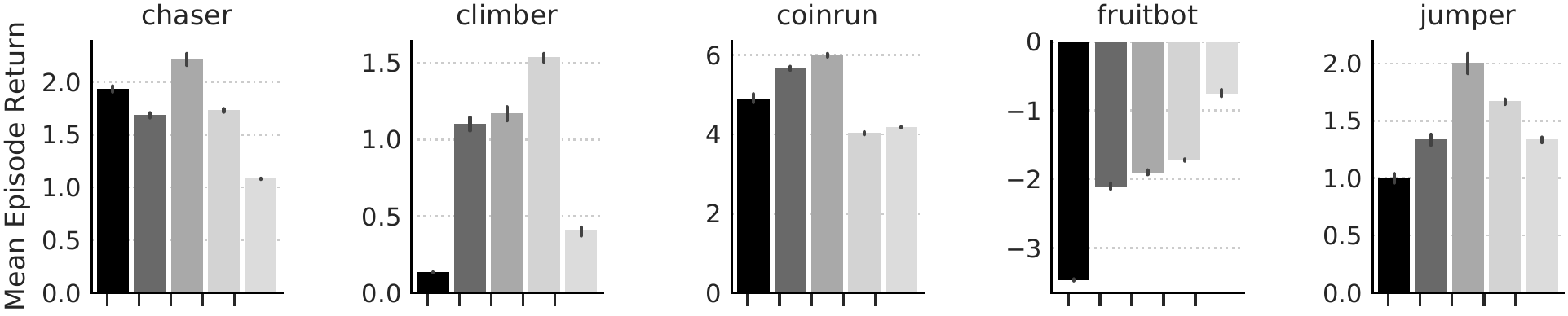}
  \includegraphics[width=0.7\linewidth, trim={0, 0, 0, 0.cm}, clip]{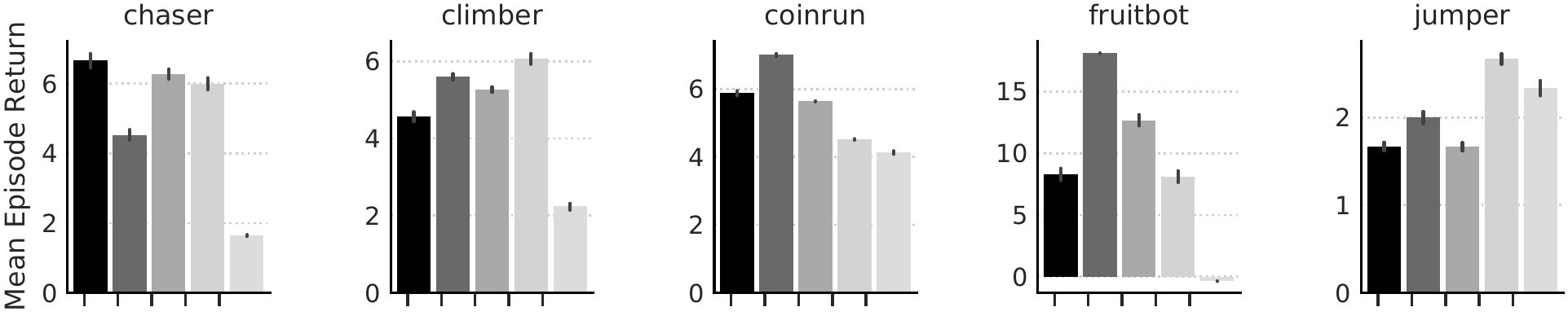}
  \caption{
    Mean episode return evaluated with pessimism bonus $-d_{JS}$-IVE with increasing weights for each procgen environment on a trained vanilla Muesli using 10 levels (top) and 100 levels (bottom). Error bars show 95\% CI.
  }
  \label{fig:pefu_beta}
\vspace{-1em}
\end{figure*}

%% file: tables/dreamer-walker-walk-full.tex
  \caption{
    Results for the Dreamer agent and IVE variants on a modified version of the \walker~task with varying degrees of reward sparsity controlled by $\eta$, where higher $\eta$ corresponds to harder exploration.
    A ``$\diamondsuit$'' indicates methods that use gradient-based trajectory optimisation, while ``$\clubsuit$'' indicates methods that use sample-based trajectory optimisation.
    We report mean and standard error of episodic returns (rounded to the nearest tenth) over 3 runs after 1M steps.
    Higher-is-better and the performance is upper bounded by 1000.
    The \textbf{best performing} method, per-task, is in bold.
  }
  \label{tab:dreamer-walker-walk-full}
  \resizebox{\linewidth}{!}{
  \begin{tabular}{lrrrrr}
  \toprule
  \textbf{Methods} &
  $\eta = 0.0$     &
  $\eta = 0.2$     &
  $\eta = 0.3$     &
  $\eta = 0.5$     \\
  \midrule
  Dreamer                               &
  \bftab 1000${\color{black!50}\pm00}$  &
  720${\color{black!50}\pm10}$          &
  570${\color{black!50}\pm60}$          &
  80${\color{black!50}\pm50}$           \\
  \midrule
  Dreamer$^{\diamondsuit}$              &
  \bftab 1000${\color{black!50}\pm00}$  &
  540${\color{black!50}\pm30}$          &
  240${\color{black!50}\pm50}$          &
  40${\color{black!50}\pm30}$           \\

  \rowcolor{ourmethod} $\mu$-IVE(5)$^{\diamondsuit}$ &
  \bftab 1000${\color{black!50}\pm00}$      &
  860${\color{black!50}\pm40}$              &
  690${\color{black!50}\pm70}$              &
  210${\color{black!50}\pm60}$              \\
  \midrule
  $\mu+\sigma$-EVE(5)$^{\diamondsuit}$      &
  \bftab 1000${\color{black!50}\pm00}$      &
  \bftab 1000${\color{black!50}\pm00}$      &
  \bftab 980${\color{black!50}\pm10}$       &
  \bftab 280${\color{black!50}\pm50}$       \\
  $\mu+\sigma$-EMVE(5)$^{\diamondsuit}$     &
  \bftab 1000${\color{black!50}\pm00}$      &
  910${\color{black!50}\pm20}$              &
  730${\color{black!50}\pm40}$              &
  210${\color{black!50}\pm60}$              \\
  \rowcolor{ourmethod} $\mu+\sigma$-IVE(5)$^{\diamondsuit}$ &
  \bftab 1000${\color{black!50}\pm00}$      &
  \bftab 1000${\color{black!50}\pm00}$      &
  \bftab 1000${\color{black!50}\pm00}$      &
  \bftab 330${\color{black!50}\pm70}$       \\
  \rowcolor{ourmethod} $\mu+\sigma$-IVE(5)$^{\clubsuit}$ &
  \bftab 1000${\color{black!50}\pm00}$      &
  \bftab 1000${\color{black!50}\pm00}$      &
  \bftab 1000${\color{black!50}\pm00}$      &
  \bftab 280${\color{black!50}\pm40}$       \\
  \bottomrule
  \end{tabular}
  }

%% file: figures/all_value_ensembles.tex
\begin{figure*}[!h]
  \centering
  \includegraphics[width=0.18\linewidth]{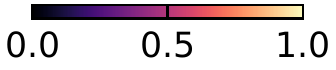}
  \\[1em]
  \begin{subfigure}[b]{0.18\linewidth}
    \centering
    \includegraphics[width=\linewidth]{assets/tabular/ive-stats/sigma-mvSIC1.pdf}
    \caption{$\sigma$-IVE(1)}
  \end{subfigure}
  ~
  \begin{subfigure}[b]{0.18\linewidth}
    \centering
    \includegraphics[width=\linewidth]{assets/tabular/ive-stats/sigma-mvSIC2.pdf}
    \caption{$\sigma$-IVE(2)}
  \end{subfigure}
  ~
  \begin{subfigure}[b]{0.18\linewidth}
    \centering
    \includegraphics[width=\linewidth]{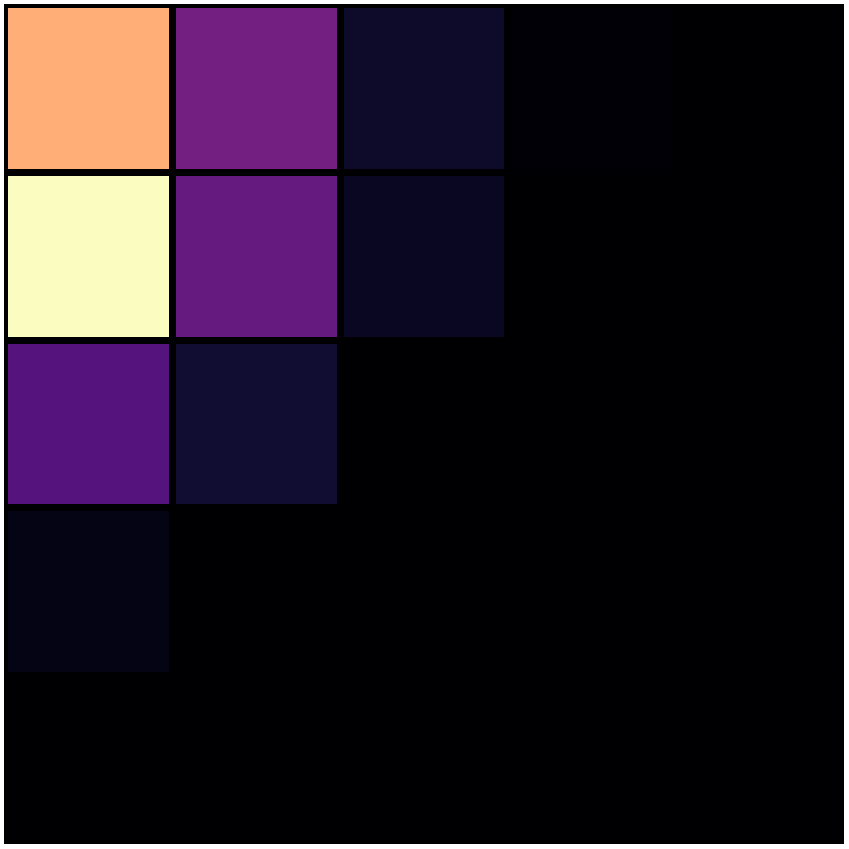}
    \caption{$\sigma$-IVE(3)}
  \end{subfigure}
  ~
  \begin{subfigure}[b]{0.18\linewidth}
    \centering
    \includegraphics[width=\linewidth]{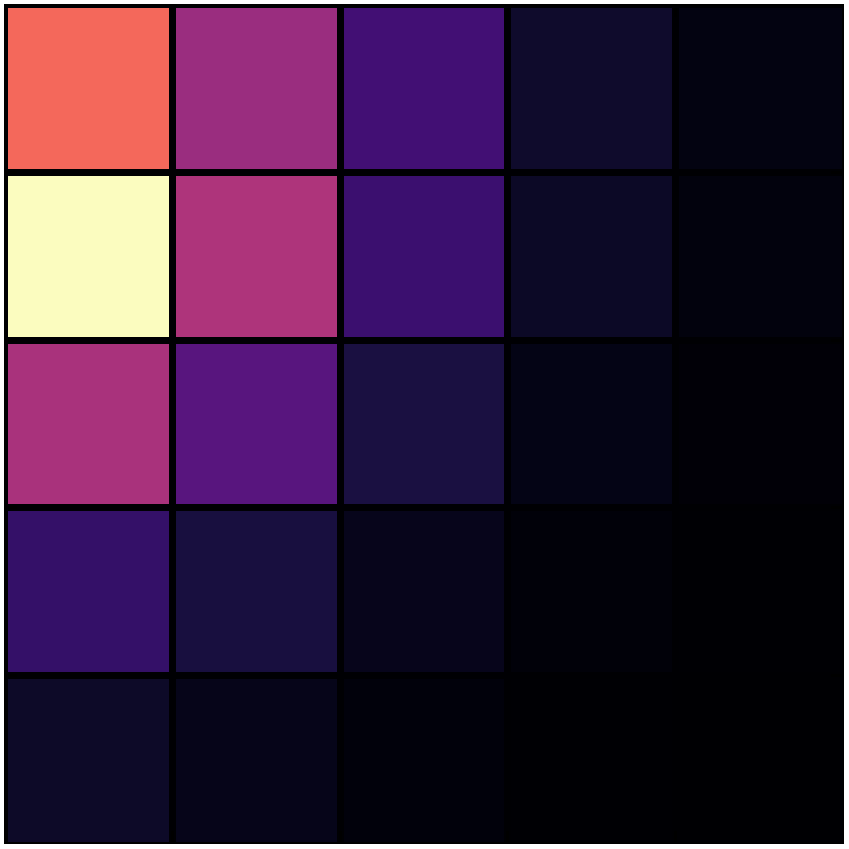}
    \caption{$\sigma$-IVE(10)}
  \end{subfigure}
  ~
  \begin{subfigure}[b]{0.18\linewidth}
    \centering
    \includegraphics[width=\linewidth]{assets/tabular/ive-stats/sigma-mvSIC20.pdf}
    \caption{$\sigma$-IVE(20)}
  \end{subfigure}
  \\[1em]
  \begin{subfigure}[b]{0.18\linewidth}
    \centering
    \includegraphics[width=\linewidth]{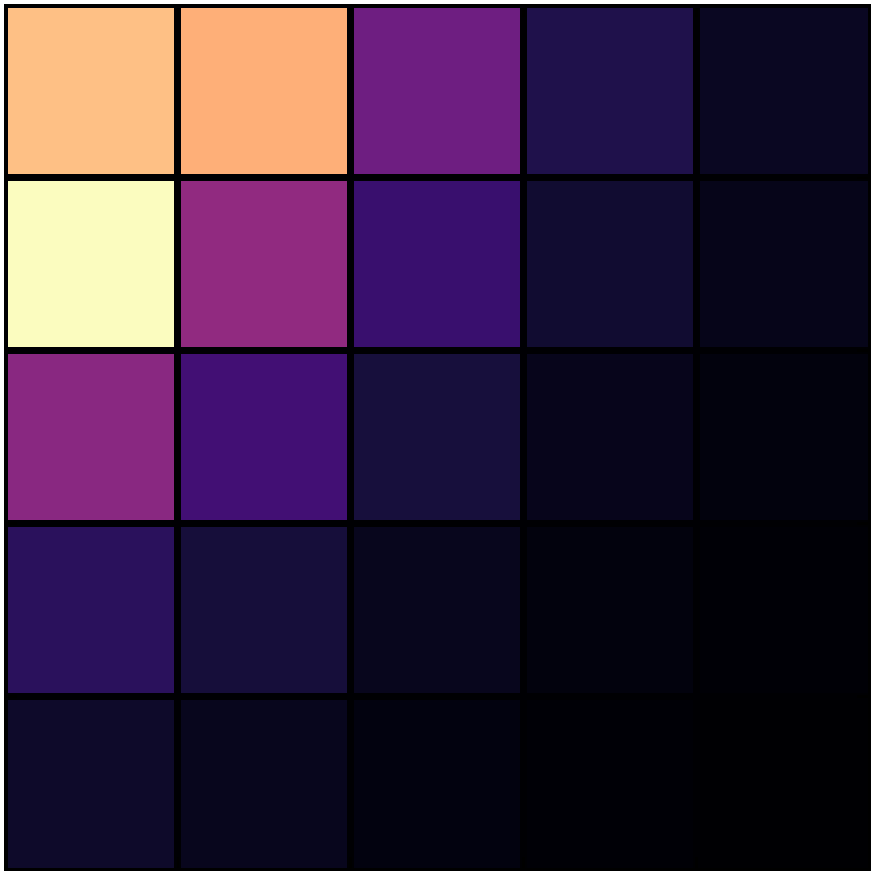}
    \caption{$\sigma$-EVE(2)}
  \end{subfigure}
  ~
  \begin{subfigure}[b]{0.18\linewidth}
    \centering
    \includegraphics[width=\linewidth]{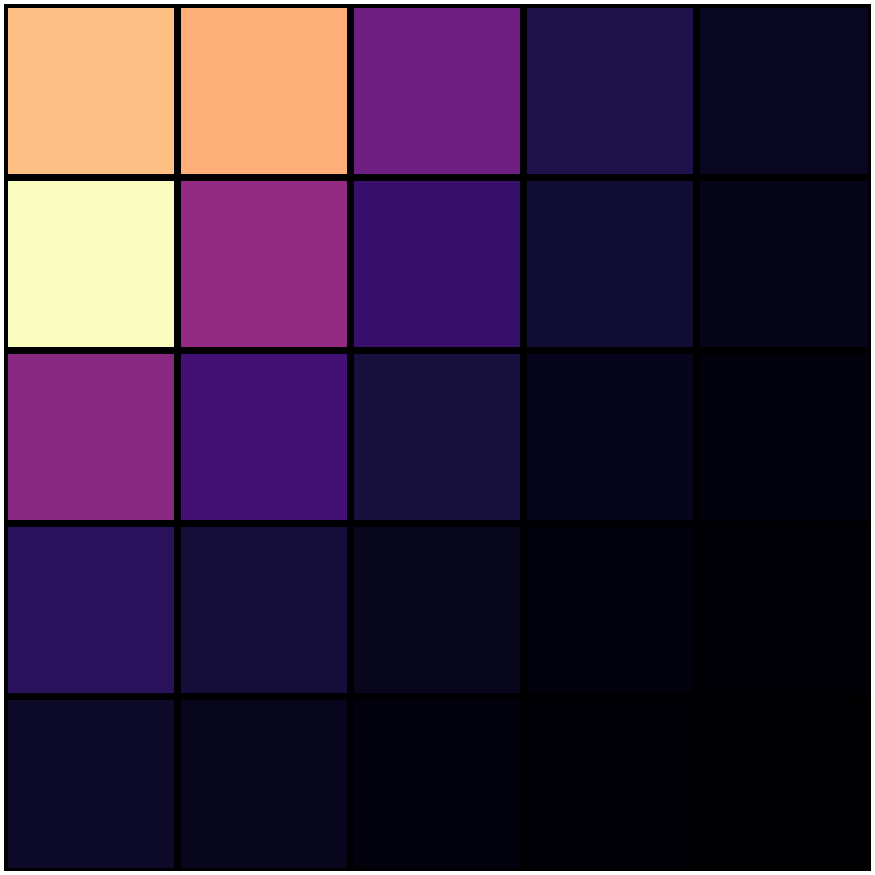}
    \caption{$\sigma$-EVE(3)}
  \end{subfigure}
  ~
  \begin{subfigure}[b]{0.18\linewidth}
    \centering
    \includegraphics[width=\linewidth]{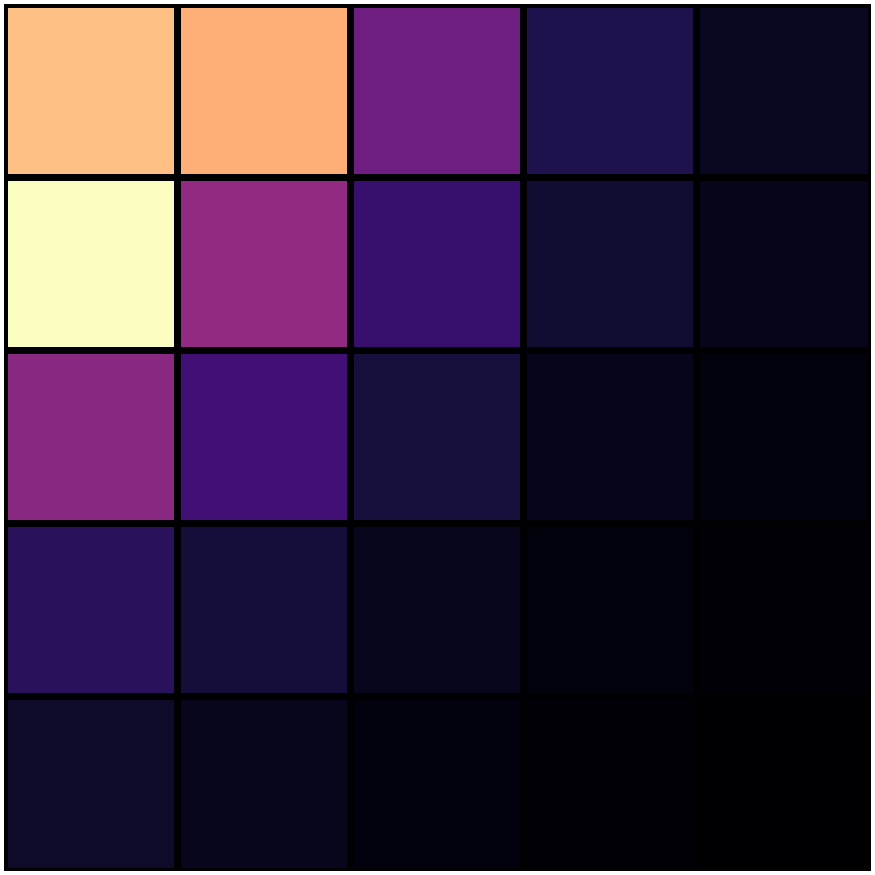}
    \caption{$\sigma$-EVE(4)}
  \end{subfigure}
  ~
  \begin{subfigure}[b]{0.18\linewidth}
    \centering
    \includegraphics[width=\linewidth]{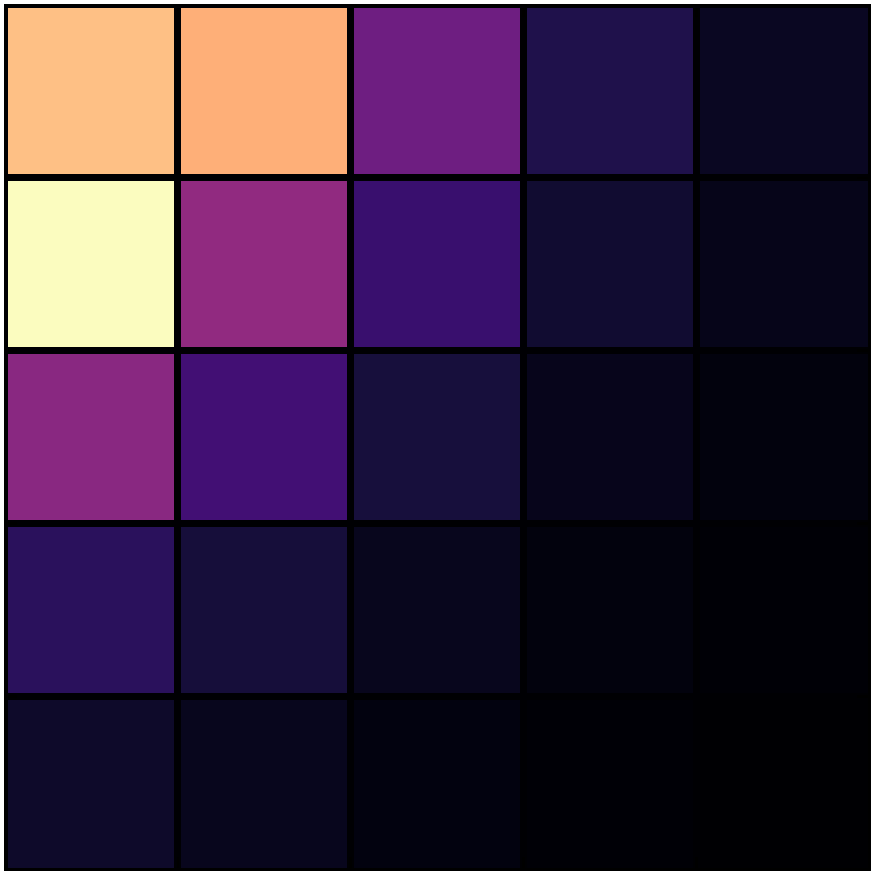}
    \caption{$\sigma$-EVE(10)}
  \end{subfigure}
  ~
  \begin{subfigure}[b]{0.18\linewidth}
    \centering
    \includegraphics[width=\linewidth]{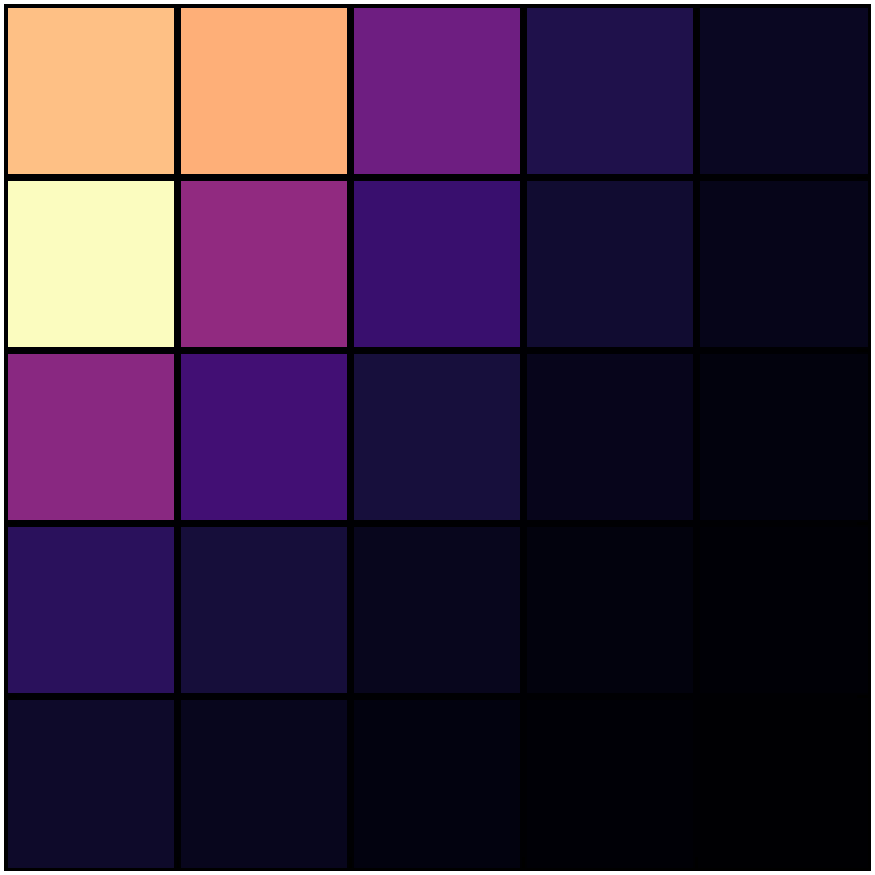}
    \caption{$\sigma$-EVE(20)}
  \end{subfigure}
  \\[1em]
  \begin{subfigure}[b]{0.18\linewidth}
    \centering
    \includegraphics[width=\linewidth]{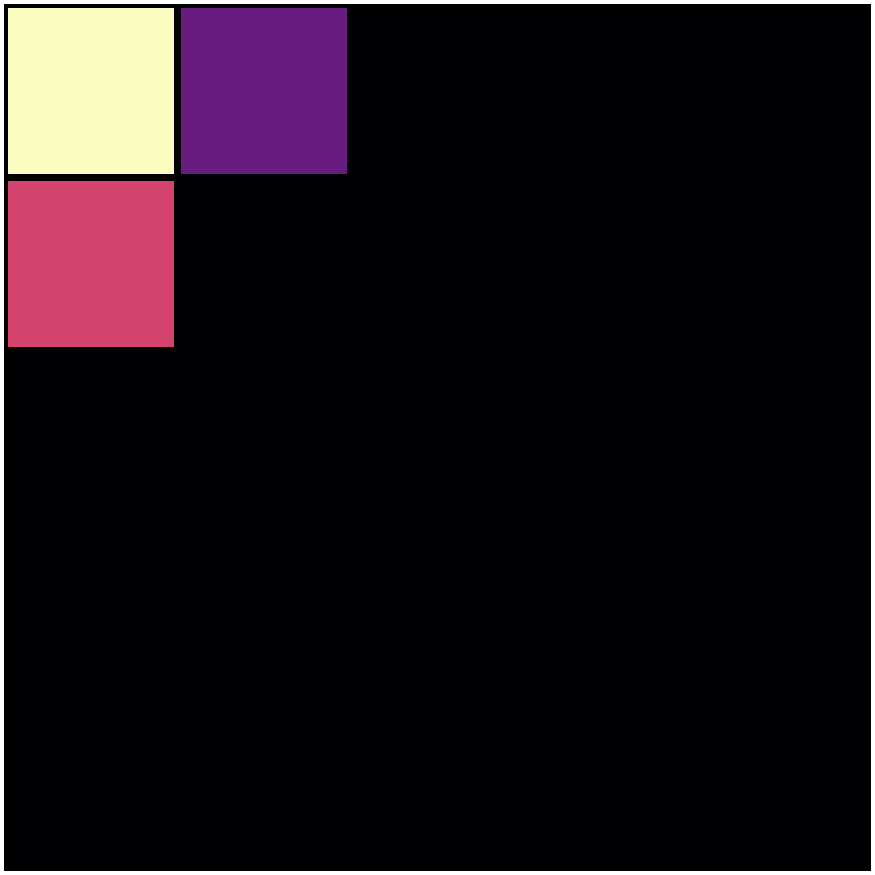}
    \caption{$\sigma$-EMVE(2)}
  \end{subfigure}
  ~
  \begin{subfigure}[b]{0.18\linewidth}
    \centering
    \includegraphics[width=\linewidth]{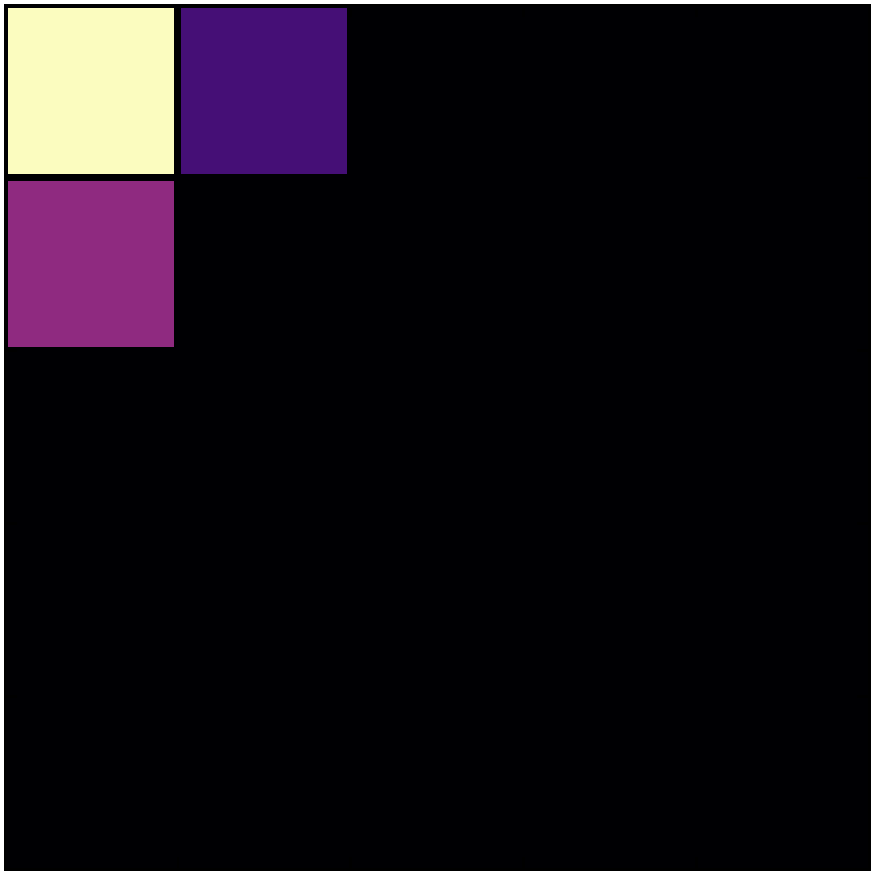}
    \caption{$\sigma$-EMVE(3)}
  \end{subfigure}
  ~
  \begin{subfigure}[b]{0.18\linewidth}
    \centering
    \includegraphics[width=\linewidth]{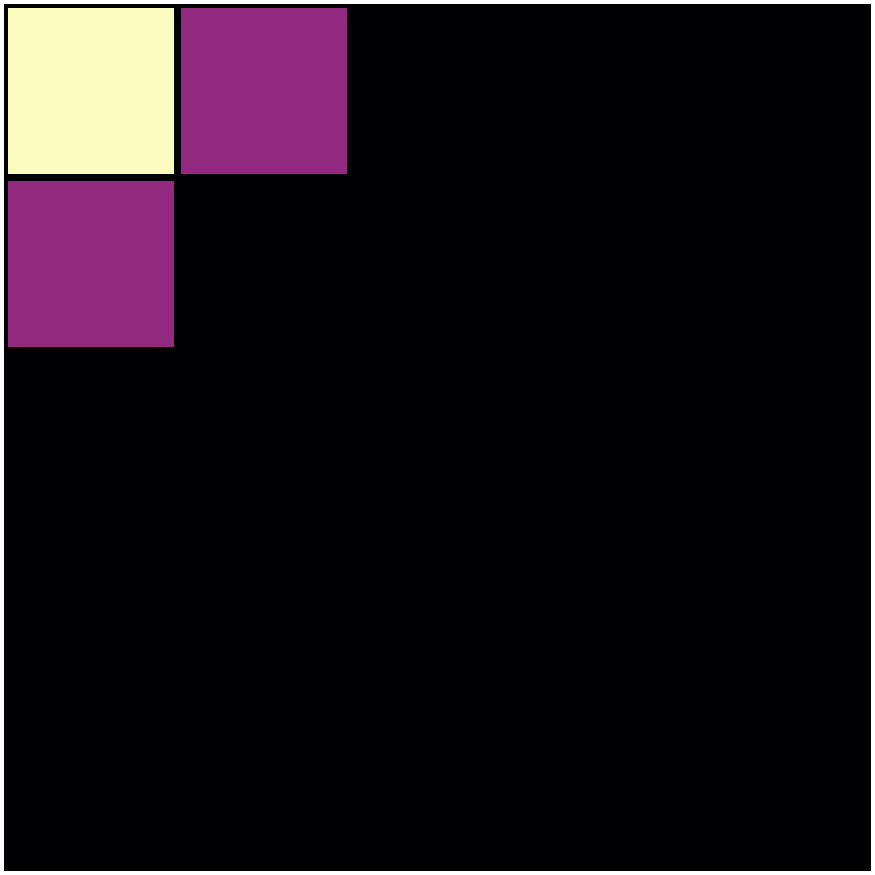}
    \caption{$\sigma$-EMVE(4)}
  \end{subfigure}
  ~
  \begin{subfigure}[b]{0.18\linewidth}
    \centering
    \includegraphics[width=\linewidth]{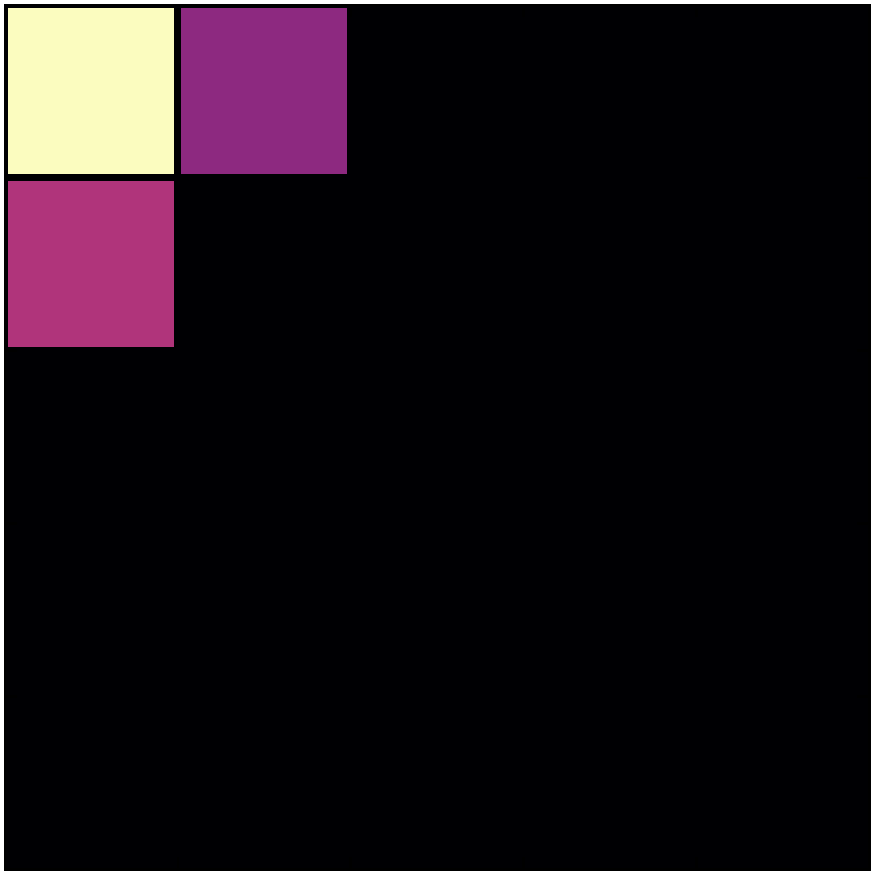}
    \caption{$\sigma$-EMVE(10)}
  \end{subfigure}
  ~
  \begin{subfigure}[b]{0.18\linewidth}
    \centering
    \includegraphics[width=\linewidth]{assets/tabular/value_ensembles/sigma-emve19.pdf}
    \caption{$\sigma$-EMVE(20)}
  \end{subfigure}

  \caption{
    Standard deviation across value ensembles.
    (i) Explicit value ensembles (EVE), as illustrated in Figure~\ref{subfig:vensemble-net};
    (ii) explicit model (value) ensembles (EMVE), as illustrated in Figure~\ref{subfig:mensemble-net} and
    (iii) implicit value ensembles (IVE), as illustrated in Figure~\ref{subfig:ive-net}.
    All values are normalised \emph{per-figure} in range $[0, 1]$.
  }
  \label{fig:all-value-ensembles}
\end{figure*}

%% file: sections/E_muesli_and_ive.tex
\section{Muesli and its Implicit Value Ensemble}
\label{app:muesli-and-ive}

In this section, we provide an exposition of how the (i) functional form and (ii) learning algorithm of the Muesli~\citep{hessel2021muesli} and MuZero~\citep{schrittwieser2020mastering} agents contribute to the diversification of their implicit value ensemble (IVE) members.
This is to complement our analysis in Section~\ref{subsec:heterogeneous-ensemble-of-value-functions-from-a-learned-model}.

\subsection{Functional Form}
The Muesli agent~\citep[see Figure 10,][]{hessel2021muesli} is comprised of (i) an representation function $\hhat(s; \omega)$, (ii) a (latent) state-value function $\vhat(z; \phi)$ and (iii) an action-conditioned model $\mhat(\cdot, \cdot | z, a; \theta) \triangleq (\rhat(z, a; \theta), \phat(z, a; \theta))$, represented as neural networks with parameters $\omega$, $\phi$ and $\theta$, respectively.
We omit the parametrisation and learning of the policy head $\pihat(\cdot | z)$ since it does not impact our analysis.
In summary, the neural network functions are given by:
\begin{align}
  \hhat_{\omega}(s)     =   \hhat(s; \omega)    &\triangleq \mathbf{z}       \in \mathcal{Z} \\
  \vhat_{\phi}(z)       =   \vhat(z; \phi)      &\triangleq \mathbf{v}       \in \mathbb{R}\\
  \rhat_{\theta}(z, a)  =   \rhat(z, a; \theta) &\triangleq \mathbf{r^{1}}   \in \mathbb{R} \\
  \phat_{\theta}(z, a)  =   \phat(z, a; \theta) &\triangleq \mathbf{z^{1}}   \in \mathcal{Z}. \label{eq:muesli-p}
\end{align}
Note the similarity with Eqn.~(\ref{eq:figure-1-h}-\ref{eq:figure-1-p}).
The main difference is that the Muesli's transition model and reward function, i.e., $\phat_{\theta}$ and $\rhat_{\theta}$, are action-conditioned.
We also use the \textbf{bold} notation introduced in Eqn.~(\ref{eq:mc-estimator}).
To ease the analysis, we define the state-to-state transition function and state-to-reward function by coupling the policy $\pihat$ with the transition function $\phat_{\theta}$ and reward function $\rhat_{\theta}$, respectively, giving rise to a transition and reward kernel i.e.,
\looseness=-1
\begin{align}
  \mathbf{z^{1}} \sim  \phat_{\theta}^{\pihat}(z) &\triangleq \phat_{\theta}(z, \pihat(\cdot | z)) \\
  \mathbf{r^{1}} \sim \rhat_{\theta}^{\pihat}(z) &\triangleq \rhat_{\theta}(z, \pihat(\cdot | z)).
  \label{eq:rphat-pi}
\end{align}

We construct the implicit value ensemble (IVE) by repeatedly applying the policy $\pihat$ and $\mhat_{\theta}$ model-induced Bellman operator $\T_{\mhat_{\theta}}^{\pihat}$ on the value function $\vhat_{\phi}$, i.e., constructing different $k$-steps model predicted values ($k$-MPVs, Eqn.~(\ref{eq:k-mpv})), given by
\begin{align}
  \mathbf{\vhat_{\mhat}^{0}}(s)  &= (\T_{\mhat_{\theta}}^{\pihat})^{0} \vhat_{\phi} (\hhat_{\omega}(s)) = \vhat_{\phi}(\hhat_{\omega}(s)) = \vhat_{\phi} \circ \hhat_{\omega} (s) \\
  \mathbf{\vhat_{\mhat}^{1}}(s)  &= (\T_{\mhat_{\theta}}^{\pihat})^{1} \vhat_{\phi} (\hhat_{\omega}(s)) = \left( \rhat_{\theta}^{\pihat} + \gamma  \vhat_{\phi} \right) \circ \phat_{\theta}^{\pihat} \circ \hhat_{\omega}(s) \\
  \mathbf{\vhat_{\mhat}^{2}}(s)  &= (\T_{\mhat_{\theta}}^{\pihat})^{2} \vhat_{\phi} (\hhat_{\omega}(s)) = \left( \rhat_{\theta} + \gamma  (\rhat_{\theta}^{\pihat} + \gamma  \vhat_{\phi}) \circ \phat_{\theta}^{\pihat} \right) \circ \phat_{\theta}^{\pihat} \circ \hhat_{\omega}(s) \\
  &\;\;\vdots \nonumber \\
  \mathbf{\vhat_{\mhat}^{k}}(s)  &= (\T_{\mhat_{\theta}}^{\pihat})^{k} \vhat_{\phi} (\hhat_{\omega}(s)) = \left( \sum_{j=1}^{k-1} \gamma^{j-1} \rhat_{\theta}^{\pihat} \circ \underdescribe{(\phat_{\theta}^{\pihat} \circ \cdots \circ \phat_{\theta}^{\pihat})}{j\text{-times}} \ + \  \gamma^{k}  \vhat_{\phi} \circ \underdescribe{(\phat_{\theta}^{\pihat} \circ \cdots \circ \phat_{\theta}^{\pihat})}{k\text{-times}} \right) \circ \hhat_{\omega}(s) \label{eq:muesli-k-mpv}
\end{align}
where $\circ$ denotes function composition.
Obviously, the functional form of the $k$-MPVs with different $k$ is different since they compose differently $\mhat_{\theta}$ and $\vhat_{\phi}$.
For instance, $\mathbf{\vhat_{\mhat}^{0}}(s)$ uses only $\vhat_{\phi}$, while $\mathbf{\vhat_{\mhat}^{1}}(s)$ and $\mathbf{\vhat_{\mhat}^{k}}(s)$ for $k>1$ use both $\mhat_{\theta}$ and $\vhat_{\phi}$ but not in the same way.
Instead, for $k \rightarrow \infty$, we obtain a purely $\mhat_{\theta}$-based prediction (a.k.a. the fixed point of $\T_{\mhat_{\theta}}^{\pihat}$).

For a stochastic policy in Eqn.~(\ref{eq:rphat-pi}) we sample from the policy $\pihat(\cdot | z)$.
Hence we obtain stochastic estimates of the $k$-MPV in Eqn.~(\ref{eq:muesli-k-mpv}), similar to Eqn.~(\ref{eq:mc-estimator}).
Nonetheless, note that the Muesli model is a \emph{deterministic expectation} model and hence we do not have to sample from it.
Empirically, we found that the impact of using a stochastic policy on the estimation of the IVE members was negligible, see Appendix~\ref{app:ablations}.
We illustrate the computational graph for the $k$-MPV in Figure~\ref{fig:ive-muesli-cg}.

\input{figures/ive-muesli-cg}

\subsection{Learning Algorithm}
The value and model learning algorithms of Muesli further diversify the IVE member predictions.
In our analysis, we consider two distinct cases: (i) when the model and value are trained with on-policy trajectories from $\pihat$ and (ii) when off-policy trajectories from behavioural policies $\pi_{\beta}$ are used.
Note that the learning algorithm of the Muesli agent uses a mix of on- and off-policy trajectories, similar to the LASER agent~\citep{schmitt2020off}.

\paragraph{On-policy trajectories.}
Provided a sequence of states, actions and rewards, collected from running policy $\pihat$ in the \emph{true} environment $\mstar$, i.e., $(s_{t:t+T}, a_{t:t+T}, r_{t:t+T}) \sim \mstar_{\pihat}$, we present the targets used for the IVE members for the first state, i.e., $\{\mathbf{v_{\mhat}^{k}}(s_{t})\}_{k=0}^{K}$, where in practice $K=5$~\citep{hessel2021muesli}.
We use the notation from Figure~\ref{fig:ive-muesli-cg}.
$n$-step bootstrap value estimates~\citep{sutton1988learning} are constructed using $\vhat_{\phi}$ and used as \emph{value targets} $v_{t:t+T-1}^{\text{target}}$, where
\begin{align}
  v_{t+i}^{\text{target}} = \sum_{j=1}^{n-1} \gamma^{j-1} r_{t+i+j} + \gamma^{n} \vhat_{\phi}(s_{t+i+n})
\end{align}
and $n=5$.
Alternative methods for constructing value targets, e.g., TD($\lambda$)~\citep{sutton2018reinforcement} could be used, too.
The ``prediction-target'' pairs for the different IVE members for state $s_{t}$ and actions $a_{t:t+T}$ are then given by\footnote{We assume that $T-1>K+n$.}
\looseness=-1
\begin{equation}
\begin{array}{rll}
  \mathbf{\vhat_{\mhat}^{0}}(s_{t})= &\mathbf{v^{0}} &\longleftarrow v_{t}^{\text{target}} \\
  \mathbf{\vhat_{\mhat}^{1}}(s_{t})=  &\mathbf{r^{1}} + \gamma\mathbf{v^{1}} &\longleftarrow r_{t+1} + \gamma v_{t+1}^{\text{target}} \\
  \mathbf{\vhat_{\mhat}^{k}}(s_{t})= &\sum_{i=1}^{k-1} \gamma^{i-1} \mathbf{r^{j}} + \gamma^{k} \mathbf{v^{k}} &\longleftarrow \sum_{i=1}^{k-1} \gamma^{i-1} r_{t+j} + \gamma^{k} v_{t+k}^{\text{target}},
\end{array}
\label{eq:muesli-mpv-targets}
\end{equation}
where the $\longleftarrow$ indicates that an objective function (e.g., L2 loss) is minimised that makes the two sides of the arrow approximately equal.
Importantly, Muesli and MuZero \emph{ground} reward and value predictions independently, or in other words, the $k$-MPV is trained by minimising the following loss:
\begin{align}
  \mathcal{L} \triangleq
  (\mathbf{r^{1}} - r_{t+1})^{2} +
  \cdots +
  (\mathbf{r^{k-1}} - r_{t+k-1})^{2} +
  (\mathbf{v^{k}} - v_{t+k}^{\text{target}})^{2}
  =
  \sum_{i=1}^{k-1} (\mathbf{r^{i}} - r_{t+i})^{2} +
  (\mathbf{v^{k}} - v_{t+k}^{\text{target}})^{2}
  .
\end{align}
Eqn.~(\ref{eq:muesli-mpv-targets}) highlight that while all the IVE members are trained to approximate the value of the policy $\pihat$, i.e., $\mathbf{\vhat_{\mhat}^{0}} \approx \mathbf{\vhat_{\mhat}^{1}} \approx \cdots \approx \mathbf{\vhat_{\mhat}^{k}} \approx v^{\pihat}$, each one is regressed against a different target/estimate of the value, further diversifying the ensemble.

\paragraph{Off-policy trajectories.}
Provided a sequence of states, actions and rewards, collected from a behavioural policy $\pi_{\beta}$ interacting with the \emph{true} environment $\mstar$, i.e., $(s_{t:t+T}, a_{t:t+T}, r_{t:t+T}) \sim \mstar_{\pi_{\beta}}$, we construct off-policy corrected $n$-step bootstrap \emph{value target} $v_{t:t+T-1}^{\text{target}}$ with the Retrace~\citep{munos2016safe} algorithm, using $\vhat_{\phi}$.

Next, we define the multi-step action-conditioned model-based reward and value estimates, i.e.,
\begin{align}
  \mathbf{\rhat_{\mhat}}(s_{t}, a_{t}, \ldots, a_{t+k-1}) \triangleq \mathbf{\rhat_{\mhat}}(s_{t}, a_{t:t+k-1})
  \quad \text{and} \quad
  \mathbf{\vhat_{\mhat}}(s_{t}, a_{t}, \ldots, a_{t+k-1}) \triangleq \mathbf{\vhat_{\mhat}}(s_{t}, a_{t:t+k-1}),
\label{eq:muesli-multi-step-ac-rv}
\end{align}
where for on-policy actions, i.e., $a_{t:t+k-1} \sim \pihat$, we note that $\mathbf{\vhat_{\mhat}}(s_{t}, a_{t:t+k-1})$ is the $k$-MPV of Eqn.~(\ref{eq:muesli-k-mpv}).
As illustrated in Figure~\ref{fig:ive-muesli-cg},
the multi-step action-conditioned reward and value estimates in Eqn.~(\ref{eq:muesli-multi-step-ac-rv}) are computed by setting the value of the nodes $\mathbf{a^{0:k-1}}$ to the replayed action sequence $a_{t:t+k-1}$ from the behavioural policy $\pi_{\beta}$.
The learning of the model and value function proceeds as in the on-policy case, obtaining the following ``prediction-target'' pairs, analogous to Eqn.~(\ref{eq:muesli-mpv-targets})
\looseness=-1
\begin{equation}
\begin{array}{ll}
  \mathbf{\vhat_{\mhat}}(s_{t}) &\longleftarrow v_{t}^{\text{target}} \\
  \mathbf{\rhat_{\mhat}}(s_{t}, a_{t}) + \gamma\mathbf{\vhat_{\mhat}}(s_{t}, a_{t}) &\longleftarrow r_{t+1} + \gamma v_{t+1}^{\text{target}} \\
  \sum_{i=1}^{k-1} \gamma^{i-1} \mathbf{\rhat_{\mhat}}(s_{t}, a_{t:t+j-1}) + \gamma^{k} \mathbf{\vhat_{\mhat}}(s_{t}, a_{t:t+k-1}) &\longleftarrow \sum_{i=1}^{k-1} \gamma^{i-1} r_{t+j} + \gamma^{k} v_{t+k}^{\text{target}},
\end{array}
\label{eq:muesli-multi-step-ac-rv-targets}
\end{equation}

Eqn.~(\ref{eq:muesli-multi-step-ac-rv-targets}) suggests that in the off-policy case, $k$-MPVs are not trained directly since the sampled sequence of actions that conditions the reward and value predictions does not (necessarily) come from $\pihat$.
Nonetheless, the multi-step action-conditioned reward and value predictors are the building blocks for constructing the $k$-MPVs and hence we conjecture that the diversity induced by training these with different targets will lead to diversity in the IVE members too.

Overall, the Muesli agent is trained with both on- and off-policy trajectories.
In the case of on-policy trajectories, we showed in Eqn.~(\ref{eq:muesli-mpv-targets}) that that different targets are used for training each IVE member.
When off-policy trajectories are used, instead of $k$-MPVs, model-based multi-step action-conditioned reward and value predictors are trained to regress different reward and value targets for different $k$.
These predictors are used for constructing the IVE members at acting time and hence this diversity at training time can impact the diversity of the IVE members too.

%% file: figures/ive-muesli-cg.tex
\begin{figure}[t]
  \centering
  \input{assets/tikz/ive-muesli-cg}
  \caption{
    The computational graph of the \emph{implicit value ensemble} (IVE) members for the Muesli~\citep{hessel2021muesli} agent.
    The action nodes $\{\mathbf{a^{i}}\}_{i=0}^{k-1}$ are stochastic nodes from which we sample from the (latent-)state-conditioned policy $\mathbf{a^{i}} \sim \pihat(\cdot | \mathbf{z^{i}})$.
    \looseness=-1
  }
  \label{fig:ive-muesli-cg}
\end{figure}
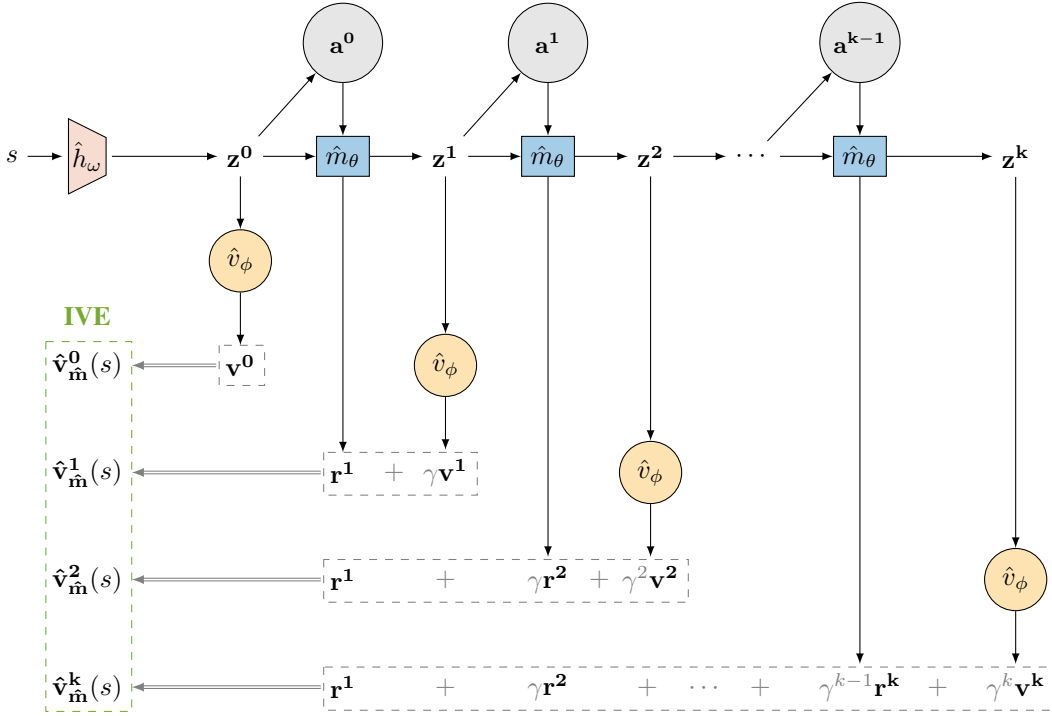

%% file: assets/tikz/ive-muesli-cg.tex
\begin{tikzpicture}
    \tikzstyle{empty}=[]
    
    \node (s) at (1,0) {$s$};
    
    \draw [draw,fill=mlorange!20] plot coordinates {(1.75, +0.5) (1.75, -0.5) (2.25, -0.25) (2.25, +0.25)} -- cycle;
    \node (h) at (2,0) {$\hhat_{\omega}$};
    \path (s) [>=latex, ->] edge (h);
    
    \node[right = 4em of h] (z0) {$\mathbf{z^{0}}$};
    \path (h) [>=latex, ->] edge (z0);
    
    \node[draw,rectangle,right = 2em of z0, fill=mlblue!35] (mh1) {$\hat{m}_{\theta}$};
    \path (z0) [>=latex, ->] edge (mh1);
    \node[right = 2em of mh1] (z1) {$\mathbf{z^{1}}$};
    \path (mh1) [>=latex, ->] edge (z1);
    \node[draw,rectangle,right = 2em of z1, fill=mlblue!35] (mh2) {$\hat{m}_{\theta}$};
    \path (z1) [>=latex, ->] edge (mh2);
    \node[right = 2em of mh2] (z2) {$\mathbf{z^{2}}$};
    \path (mh2) [>=latex, ->] edge (z2);
    \node[right = 2em of z2] (z3) {$\cdots$};
    \path (z2) [>=latex, ->] edge (z3);
    \node[draw,rectangle,right = 2em of z3, fill=mlblue!35] (mh3) {$\hat{m}_{\theta}$};
    \path (z3) [>=latex, ->] edge (mh3);
    \node[right = 4em of mh3] (z4) {$\mathbf{z^{k}}$};
    \path (mh3) [>=latex, ->] edge (z4);
    
    \node[draw,circle,above = 2em of mh1,fill=gray!20, minimum size=3em] (a1) {\footnotesize$\mathbf{a^{0}}$};
    \path (z0) [>=latex, ->] edge (a1);
    \path (a1) edge [>=latex, ->] (mh1);
    \node[draw,circle,above = 2em of mh2,fill=gray!20, minimum size=3em] (a2) {\footnotesize$\mathbf{a^{1}}$};
    \path (z1) [>=latex, ->] edge (a2);
    \path (a2) [>=latex, ->] edge (mh2);
    \node[draw,circle,above = 2em of mh3,fill=gray!20, minimum size=3em] (a4) {\footnotesize$\mathbf{a^{k-1}}$};
    \path (z3) [>=latex, ->] edge (a4);
    \path (a4) [>=latex, ->] edge (mh3);

    \node[draw,circle,below = 2em of z0, fill=mlyellow!35] (vh0) {$\hat{v}_{\phi}$};
    \path (z0) [>=latex, ->] edge (vh0);
    \node[below = 2em of vh0] (v0) {$\mathbf{v^{0}}$};
    \path (vh0) [>=latex, ->] edge (v0);
    
    \node[draw,circle, fill=mlyellow!35] (vh1) at (v0 -| z1) {$\hat{v}_{\phi}$};
    \path (z1) [>=latex, ->] edge (vh1);
    \node[below = 2em of vh1] (v1) {${\color{gray} \gamma} \mathbf{v^{1}}$};
    \path (vh1) [>=latex, ->] edge (v1);
    \node[draw,circle,fill=mlyellow!35] (vh2) at (v1 -| z2) {$\hat{v}_{\phi}$};
    \path (z2) [>=latex, ->] edge (vh2);
    \node[below = 2em of vh2] (v2) {${\color{gray} \gamma^{2}} \mathbf{v^{2}}$};
    \path (vh2) [>=latex, ->] edge (v2);
    \node[draw,circle,fill=mlyellow!35] (vh4) at (v2 -| z4) {$\hat{v}_{\phi}$};
    \path (z4) [>=latex, ->] edge (vh4);
    \node[below = 2em of vh4] (v4) {${\color{gray} \gamma^{k}} \mathbf{v^{k}}$};
    \path (vh4) [>=latex, ->] edge (v4);

    \node (r1) at (v1 -| mh1) {$\mathbf{r^{1}}$};
    \path (mh1) [>=latex, ->] edge (r1);
    \node (r2) at (v2 -| mh2) {${\color{gray} \gamma} \mathbf{r^{2}}$};
    \path (mh2) [>=latex, ->] edge (r2);
    \node (r1_) at (r2 -| r1) {$\mathbf{r^{1}}$};
    \node (r4) at (v4 -| mh3) {${\color{gray} \gamma^{k-1}} \mathbf{r^{k}}$};
    \path (mh3) [>=latex, ->] edge (r4);
    \node (r1__) at (r4 -| r1) {$\mathbf{r^{1}}$};
    \node (r2_) at (r4 -| r2) {${\color{gray} \gamma} \mathbf{r^{2}}$};
    
    \node (plus) at ($(r1)!0.5!(v1)$) {\color{gray}$+$};
    \node (plus) at ($(r2)!0.5!(v2)$) {\color{gray}$+$};
    \node (plus) at ($(r1_)!0.5!(r2)$) {\color{gray}$+$};
    \node (plus) at ($(r4)!0.5!(v4)$) {\color{gray}$+$};
    \node (plus) at ($(r1__)!0.5!(r2_)$) {\color{gray}$+$};
    \node (plus) at ($(r2_)!0.5!(r4)$) {\color{gray}$+\quad \cdots \quad +$};
    
    \node (ive0) at (v0 -| h) {$\mathbf{\vhat_{\mhat}^{0}}(s)$};
    \node (ive1) at (r1 -| h) {$\mathbf{\vhat_{\mhat}^{1}}(s)$};
    \node (ive2) at (r1_ -| h) {$\mathbf{\vhat_{\mhat}^{2}}(s)$};
    \node (ive4) at (r1__ -| h) {$\mathbf{\vhat_{\mhat}^{k}}(s)$};
    {\color{mlgreen}
    \draw[dashed,draw] ($(ive0.north west)+(+0.04,0.00)$) rectangle ($(ive4.south east)+(0.00,-0.00)$);
    \node[above = 0.25em of ive0] (ivelabel) {\textbf{IVE}};
    }

    {\color{gray}
    \draw[dashed,draw] ($(v0.north west)+(+0.04,0.00)$) rectangle ($(v0.south east)+(0.00,-0.00)$);
    \path (v0) edge [>=latex, double, ->] (ive0);
    \draw[dashed,draw] ($(r1.north west)+(+0.04,0.00)$) rectangle ($(v1.south east)+(0.00,-0.00)$);
    \path (r1) edge [>=latex, double, ->] (ive1);
    \draw[dashed,draw] ($(r1_.north west)+(+0.04,0.00)$) rectangle ($(v2.south east)+(0.00,-0.00)$);
    \path (r1_) edge [>=latex, double, ->] (ive2);
    \draw[dashed,draw] ($(r1__.north west)+(+0.04,0.00)$) rectangle ($(v4.south east)+(0.00,-0.00)$);
    \path (r1__) edge [>=latex, double, ->] (ive4);
    }

\end{tikzpicture}

%% file: main.bbl
\begin{thebibliography}{97}
\providecommand{\natexlab}[1]{#1}
\providecommand{\url}[1]{\texttt{#1}}
\expandafter\ifx\csname urlstyle\endcsname\relax
  \providecommand{\doi}[1]{doi: #1}\else
  \providecommand{\doi}{doi: \begingroup \urlstyle{rm}\Url}\fi

\bibitem[Abadi et~al.(2016)Abadi, Barham, Chen, Chen, Davis, Dean, Devin,
  Ghemawat, Irving, Isard, et~al.]{abadi2016tensorflow}
Abadi, M., Barham, P., Chen, J., Chen, Z., Davis, A., Dean, J., Devin, M.,
  Ghemawat, S., Irving, G., Isard, M., et~al.
\newblock
  \href{https://www.usenix.org/system/files/conference/osdi16/osdi16-abadi.pdf}{Tensorflow:
  A system for large-scale machine learning}.
\newblock In \emph{12th $\{$USENIX$\}$ symposium on operating systems design
  and implementation ($\{$OSDI$\}$ 16)}, pp.\  265--283, 2016.

\bibitem[Agarwal et~al.(2020)Agarwal, Schuurmans, and
  Norouzi]{agarwal2020optimistic}
Agarwal, R., Schuurmans, D., and Norouzi, M.
\newblock \href{https://arxiv.org/abs/1907.04543}{An optimistic perspective on
  offline reinforcement learning}.
\newblock In \emph{International Conference on Machine Learning}, pp.\
  104--114. PMLR, 2020.

\bibitem[Anschel et~al.(2017)Anschel, Baram, and Shimkin]{anschel2017averaged}
Anschel, O., Baram, N., and Shimkin, N.
\newblock \href{https://arxiv.org/abs/1611.01929}{Averaged-dqn: Variance
  reduction and stabilization for deep reinforcement learning}.
\newblock In \emph{International conference on machine learning}, pp.\
  176--185. PMLR, 2017.

\bibitem[Antor{\'a}n et~al.(2020)Antor{\'a}n, Allingham, and
  Hern{\'a}ndez-Lobato]{antoran2020depth}
Antor{\'a}n, J., Allingham, J.~U., and Hern{\'a}ndez-Lobato, J.~M.
\newblock \href{https://arxiv.org/abs/2006.08437}{Depth uncertainty in neural
  networks}.
\newblock \emph{arXiv preprint arXiv:2006.08437}, 2020.

\bibitem[Babuschkin et~al.(2020)Babuschkin, Baumli, Bell, Bhupatiraju, Bruce,
  Buchlovsky, Budden, Cai, Clark, Danihelka, Fantacci, Godwin, Jones, Hennigan,
  Hessel, Kapturowski, Keck, Kemaev, King, Martens, Mikulik, Norman, Quan,
  Papamakarios, Ring, Ruiz, Sanchez, Schneider, Sezener, Spencer, Srinivasan,
  Stokowiec, and Viola]{deepmind2020jax}
Babuschkin, I., Baumli, K., Bell, A., Bhupatiraju, S., Bruce, J., Buchlovsky,
  P., Budden, D., Cai, T., Clark, A., Danihelka, I., Fantacci, C., Godwin, J.,
  Jones, C., Hennigan, T., Hessel, M., Kapturowski, S., Keck, T., Kemaev, I.,
  King, M., Martens, L., Mikulik, V., Norman, T., Quan, J., Papamakarios, G.,
  Ring, R., Ruiz, F., Sanchez, A., Schneider, R., Sezener, E., Spencer, S.,
  Srinivasan, S., Stokowiec, W., and Viola, F.
\newblock \href{http://github.com/deepmind}{The {D}eep{M}ind {JAX}
  {E}cosystem}, 2020.

\bibitem[Ball et~al.(2020)Ball, Parker-Holder, Pacchiano, Choromanski, and
  Roberts]{ball2020ready}
Ball, P., Parker-Holder, J., Pacchiano, A., Choromanski, K., and Roberts, S.
\newblock \href{https://arxiv.org/abs/2002.02693}{Ready policy one: World
  building through active learning}.
\newblock In \emph{International Conference on Machine Learning}, pp.\
  591--601. PMLR, 2020.

\bibitem[Bellemare et~al.(2016)Bellemare, Srinivasan, Ostrovski, Schaul,
  Saxton, and Munos]{bellemare2016unifying}
Bellemare, M., Srinivasan, S., Ostrovski, G., Schaul, T., Saxton, D., and
  Munos, R.
\newblock \href{https://arxiv.org/abs/1606.01868}{Unifying count-based
  exploration and intrinsic motivation}.
\newblock \emph{Advances in neural information processing systems},
  29:\penalty0 1471--1479, 2016.

\bibitem[Bellemare et~al.(2017)Bellemare, Dabney, and
  Munos]{bellemare2017distributional}
Bellemare, M.~G., Dabney, W., and Munos, R.
\newblock \href{https://arxiv.org/abs/1707.06887}{A distributional perspective
  on reinforcement learning}.
\newblock In \emph{International Conference on Machine Learning}, pp.\
  449--458. PMLR, 2017.

\bibitem[Bellman(1957)]{bellman1957markovian}
Bellman, R.
\newblock \href{https://www.jstor.org/stable/24900506}{A Markovian decision
  process}.
\newblock \emph{Journal of mathematics and mechanics}, 6\penalty0 (5):\penalty0
  679--684, 1957.

\bibitem[Blundell et~al.(2015)Blundell, Cornebise, Kavukcuoglu, and
  Wierstra]{blundell2015weight}
Blundell, C., Cornebise, J., Kavukcuoglu, K., and Wierstra, D.
\newblock \href{https://arxiv.org/abs/1505.05424}{Weight uncertainty in neural
  network}.
\newblock In \emph{International Conference on Machine Learning}, pp.\
  1613--1622. PMLR, 2015.

\bibitem[Bojar \& Tamchyna(2011)Bojar and Tamchyna]{bojar2011improving}
Bojar, O. and Tamchyna, A.
\newblock \href{https://aclanthology.org/W11-2138/}{Improving translation model
  by monolingual data}.
\newblock In \emph{Proceedings of the Sixth Workshop on Statistical Machine
  Translation}, pp.\  330--336, 2011.

\bibitem[Bradbury et~al.(2018)Bradbury, Frostig, Hawkins, Johnson, Leary,
  Maclaurin, Necula, Paszke, Vander{P}las, Wanderman-{M}ilne, and
  Zhang]{jax2018github}
Bradbury, J., Frostig, R., Hawkins, P., Johnson, M.~J., Leary, C., Maclaurin,
  D., Necula, G., Paszke, A., Vander{P}las, J., Wanderman-{M}ilne, S., and
  Zhang, Q.
\newblock \href{https://github.com/google/jax}{{JAX}: composable
  transformations of {P}ython+{N}um{P}y programs}, 2018.

\bibitem[Buckman et~al.(2018)Buckman, Hafner, Tucker, Brevdo, and
  Lee]{buckman2018sample}
Buckman, J., Hafner, D., Tucker, G., Brevdo, E., and Lee, H.
\newblock \href{https://arxiv.org/abs/1807.01675}{Sample-efficient
  reinforcement learning with stochastic ensemble value expansion}.
\newblock \emph{arXiv preprint arXiv:1807.01675}, 2018.

\bibitem[Byravan et~al.(2020)Byravan, Springenberg, Abdolmaleki, Hafner,
  Neunert, Lampe, Siegel, Heess, and Riedmiller]{byravan2020imagined}
Byravan, A., Springenberg, J.~T., Abdolmaleki, A., Hafner, R., Neunert, M.,
  Lampe, T., Siegel, N., Heess, N., and Riedmiller, M.
\newblock \href{https://arxiv.org/abs/1910.04142}{Imagined value gradients:
  Model-based policy optimization with tranferable latent dynamics models}.
\newblock In \emph{Conference on Robot Learning}, pp.\  566--589. PMLR, 2020.

\bibitem[Chua et~al.(2018)Chua, Calandra, McAllister, and Levine]{chua2018deep}
Chua, K., Calandra, R., McAllister, R., and Levine, S.
\newblock \href{https://arxiv.org/abs/1805.12114}{Deep reinforcement learning
  in a handful of trials using probabilistic dynamics models}.
\newblock \emph{arXiv preprint arXiv:1805.12114}, 2018.

\bibitem[Clevert et~al.(2015)Clevert, Unterthiner, and
  Hochreiter]{clevert2015fast}
Clevert, D.-A., Unterthiner, T., and Hochreiter, S.
\newblock \href{https://arxiv.org/abs/1511.07289}{Fast and accurate deep
  network learning by exponential linear units (elus)}.
\newblock \emph{arXiv preprint arXiv:1511.07289}, 2015.

\bibitem[Cobbe et~al.(2019)Cobbe, Hesse, Hilton, and Schulman]{procgen19}
Cobbe, K., Hesse, C., Hilton, J., and Schulman, J.
\newblock \href{http://arxiv.org/abs/1912.01588}{Leveraging Procedural
  Generation to Benchmark Reinforcement Learning}.
\newblock \emph{CoRR}, abs/1912.01588, 2019.

\bibitem[Coulom(2006)]{coulom2006efficient}
Coulom, R.
\newblock \href{https://hal.inria.fr/inria-00116992/document}{Efficient
  selectivity and backup operators in Monte-Carlo tree search}.
\newblock In \emph{International conference on computers and games}, pp.\
  72--83. Springer, 2006.

\bibitem[Dabney et~al.(2020)Dabney, Barreto, Rowland, Dadashi, Quan, Bellemare,
  and Silver]{dabney2020value}
Dabney, W., Barreto, A., Rowland, M., Dadashi, R., Quan, J., Bellemare, M.~G.,
  and Silver, D.
\newblock \href{https://arxiv.org/abs/2006.02243}{The value-improvement path:
  Towards better representations for reinforcement learning}.
\newblock \emph{arXiv preprint arXiv:2006.02243}, 2020.

\bibitem[Dearden et~al.(1998)Dearden, Friedman, and
  Russell]{dearden1998bayesian}
Dearden, R., Friedman, N., and Russell, S.
\newblock \href{https://www.aaai.org/Papers/AAAI/1998/AAAI98-108.pdf}{Bayesian
  Q-learning}.
\newblock In \emph{Aaai/iaai}, pp.\  761--768, 1998.

\bibitem[Dearden et~al.(1999)Dearden, Friedman, and Andre]{dearden2013model}
Dearden, R., Friedman, N., and Andre, D.
\newblock \href{https://arxiv.org/abs/1301.6690}{Model-based Bayesian
  exploration}.
\newblock \emph{arXiv preprint arXiv:1301.6690}, 1999.

\bibitem[Dusenberry et~al.(2020)Dusenberry, Jerfel, Wen, Ma, Snoek, Heller,
  Lakshminarayanan, and Tran]{dusenberry2020efficient}
Dusenberry, M., Jerfel, G., Wen, Y., Ma, Y., Snoek, J., Heller, K.,
  Lakshminarayanan, B., and Tran, D.
\newblock \href{https://arxiv.org/abs/2005.07186}{Efficient and scalable
  bayesian neural nets with rank-1 factors}.
\newblock In \emph{International conference on machine learning}, pp.\
  2782--2792. PMLR, 2020.

\bibitem[Edunov et~al.(2018)Edunov, Ott, Auli, and
  Grangier]{edunov2018understanding}
Edunov, S., Ott, M., Auli, M., and Grangier, D.
\newblock \href{https://arxiv.org/abs/1808.09381}{Understanding
  back-translation at scale}.
\newblock \emph{arXiv preprint arXiv:1808.09381}, 2018.

\bibitem[Farquhar et~al.(2017)Farquhar, Rockt{\"a}schel, Igl, and
  Whiteson]{farquhar2017treeqn}
Farquhar, G., Rockt{\"a}schel, T., Igl, M., and Whiteson, S.
\newblock \href{https://arxiv.org/abs/1710.11417}{Treeqn and atreec:
  Differentiable tree-structured models for deep reinforcement learning}.
\newblock \emph{arXiv preprint arXiv:1710.11417}, 2017.

\bibitem[Farquhar et~al.(2021)Farquhar, Baumli, Marinho, Filos, Hessel, van
  Hasselt, and Silver]{farquhar2021self}
Farquhar, G., Baumli, K., Marinho, Z., Filos, A., Hessel, M., van Hasselt, H.,
  and Silver, D.
\newblock \href{https://arxiv.org/abs/2110.12840}{Self-consistent models and
  values}.
\newblock \emph{arXiv preprint arXiv:2110.12840}, 2021.

\bibitem[Fau{\ss}er \& Schwenker(2015)Fau{\ss}er and
  Schwenker]{fausser2015neural}
Fau{\ss}er, S. and Schwenker, F.
\newblock
  \href{https://link.springer.com/content/pdf/10.1007/s11063-013-9334-5.pdf}{Neural
  network ensembles in reinforcement learning}.
\newblock \emph{Neural Processing Letters}, 41\penalty0 (1):\penalty0 55--69,
  2015.

\bibitem[Fedus et~al.(2019)Fedus, Gelada, Bengio, Bellemare, and
  Larochelle]{fedus2019hyperbolic}
Fedus, W., Gelada, C., Bengio, Y., Bellemare, M.~G., and Larochelle, H.
\newblock \href{https://arxiv.org/abs/1902.06865}{Hyperbolic discounting and
  learning over multiple horizons}.
\newblock \emph{arXiv preprint arXiv:1902.06865}, 2019.

\bibitem[Feinberg et~al.(2018)Feinberg, Wan, Stoica, Jordan, Gonzalez, and
  Levine]{feinberg2018model}
Feinberg, V., Wan, A., Stoica, I., Jordan, M.~I., Gonzalez, J.~E., and Levine,
  S.
\newblock \href{https://arxiv.org/abs/1803.00101}{Model-based value expansion
  for efficient model-free reinforcement learning}.
\newblock In \emph{Proceedings of the 35th International Conference on Machine
  Learning (ICML 2018)}, 2018.

\bibitem[Filos et~al.(2020)Filos, Tigkas, McAllister, Rhinehart, Levine, and
  Gal]{filos2020can}
Filos, A., Tigkas, P., McAllister, R., Rhinehart, N., Levine, S., and Gal, Y.
\newblock \href{https://arxiv.org/abs/2006.14911}{Can autonomous vehicles
  identify, recover from, and adapt to distribution shifts?}
\newblock In \emph{International Conference on Machine Learning}, pp.\
  3145--3153. PMLR, 2020.

\bibitem[Filos et~al.(2021)Filos, Lyle, Gal, Levine, Jaques, and
  Farquhar]{filos2021psiphi}
Filos, A., Lyle, C., Gal, Y., Levine, S., Jaques, N., and Farquhar, G.
\newblock \href{https://arxiv.org/abs/2102.12560}{PsiPhi-Learning:
  Reinforcement Learning with Demonstrations using Successor Features and
  Inverse Temporal Difference Learning}.
\newblock \emph{arXiv preprint arXiv:2102.12560}, 2021.

\bibitem[Flennerhag et~al.(2020)Flennerhag, Wang, Sprechmann, Visin, Galashov,
  Kapturowski, Borsa, Heess, Barreto, and Pascanu]{flennerhag2020temporal}
Flennerhag, S., Wang, J.~X., Sprechmann, P., Visin, F., Galashov, A.,
  Kapturowski, S., Borsa, D.~L., Heess, N., Barreto, A., and Pascanu, R.
\newblock \href{https://arxiv.org/abs/2010.02255}{Temporal Difference
  Uncertainties as a Signal for Exploration}.
\newblock \emph{arXiv preprint arXiv:2010.02255}, 2020.

\bibitem[Garcia et~al.(1989)Garcia, Prett, and Morari]{garcia1989model}
Garcia, C.~E., Prett, D.~M., and Morari, M.
\newblock
  \href{https://www.researchgate.net/profile/Mohamed-Mourad-Lafifi/post/Is_there_any_good_and_easy_to_understand_set_of_lectrues_or_a_good_reference_for_someone_who_wants_to_learn_optimal_control_theory/attachment/5a758d464cde266d58886cc7/AS\%3A589894925692933\%401517653318888/download/Model+Predictive+Control+_+Theory+and+Practice+a+Survey+Garca1989.pdf}{Model
  predictive control: Theory and practice—A survey}.
\newblock \emph{Automatica}, 25\penalty0 (3):\penalty0 335--348, 1989.

\bibitem[Gregor et~al.(2019)Gregor, Rezende, Besse, Wu, Merzic, and
  Oord]{gregor2019shaping}
Gregor, K., Rezende, D.~J., Besse, F., Wu, Y., Merzic, H., and Oord, A. v.~d.
\newblock \href{https://arxiv.org/abs/1906.09237}{Shaping belief states with
  generative environment models for rl}.
\newblock \emph{arXiv preprint arXiv:1906.09237}, 2019.

\bibitem[Grimm et~al.(2021)Grimm, Barreto, Farquhar, Silver, and
  Singh]{grimm2021proper}
Grimm, C., Barreto, A., Farquhar, G., Silver, D., and Singh, S.
\newblock \href{https://arxiv.org/abs/2106.10316}{Proper Value Equivalence}.
\newblock \emph{arXiv preprint arXiv:2106.10316}, 2021.

\bibitem[Guez et~al.(2020)Guez, Viola, Weber, Buesing, Kapturowski, Precup,
  Silver, and Heess]{guez2020value}
Guez, A., Viola, F., Weber, T., Buesing, L., Kapturowski, S., Precup, D.,
  Silver, D., and Heess, N.
\newblock \href{https://arxiv.org/abs/2002.08329}{Value-driven hindsight
  modelling}.
\newblock \emph{arXiv preprint arXiv:2002.08329}, 2020.

\bibitem[Hafner et~al.(2019{\natexlab{a}})Hafner, Lillicrap, Ba, and
  Norouzi]{hafner2019dream}
Hafner, D., Lillicrap, T., Ba, J., and Norouzi, M.
\newblock \href{https://arxiv.org/abs/1912.01603}{Dream to control: Learning
  behaviors by latent imagination}.
\newblock \emph{arXiv preprint arXiv:1912.01603}, 2019{\natexlab{a}}.

\bibitem[Hafner et~al.(2019{\natexlab{b}})Hafner, Lillicrap, Fischer, Villegas,
  Ha, Lee, and Davidson]{hafner2019learning}
Hafner, D., Lillicrap, T., Fischer, I., Villegas, R., Ha, D., Lee, H., and
  Davidson, J.
\newblock \href{https://arxiv.org/abs/1811.04551}{Learning latent dynamics for
  planning from pixels}.
\newblock In \emph{International Conference on Machine Learning}, pp.\
  2555--2565. PMLR, 2019{\natexlab{b}}.

\bibitem[Hessel et~al.(2021)Hessel, Danihelka, Viola, Guez, Schmitt, Sifre,
  Weber, Silver, and van Hasselt]{hessel2021muesli}
Hessel, M., Danihelka, I., Viola, F., Guez, A., Schmitt, S., Sifre, L., Weber,
  T., Silver, D., and van Hasselt, H.
\newblock \href{https://arxiv.org/abs/2104.06159}{Muesli: Combining
  improvements in policy optimization}.
\newblock \emph{arXiv preprint arXiv:2104.06159}, 2021.

\bibitem[Hochreiter \& Schmidhuber(1997)Hochreiter and
  Schmidhuber]{hochreiter1997long}
Hochreiter, S. and Schmidhuber, J.
\newblock \href{https://www.bioinf.jku.at/publications/older/2604.pdf}{Long
  short-term memory}.
\newblock \emph{Neural computation}, 9\penalty0 (8):\penalty0 1735--1780, 1997.

\bibitem[Huang et~al.(2016)Huang, Sun, Liu, Sedra, and
  Weinberger]{huang2016deep}
Huang, G., Sun, Y., Liu, Z., Sedra, D., and Weinberger, K.~Q.
\newblock \href{https://arxiv.org/abs/1603.09382}{Deep networks with stochastic
  depth}.
\newblock In \emph{European conference on computer vision}, pp.\  646--661.
  Springer, 2016.

\bibitem[Huang et~al.(2017)Huang, Li, Pleiss, Liu, Hopcroft, and
  Weinberger]{huang2017snapshot}
Huang, G., Li, Y., Pleiss, G., Liu, Z., Hopcroft, J.~E., and Weinberger, K.~Q.
\newblock \href{https://arxiv.org/abs/1704.00109}{Snapshot ensembles: Train 1,
  get m for free}.
\newblock \emph{arXiv preprint arXiv:1704.00109}, 2017.

\bibitem[Hunter(2007)]{hunter2007matplotlib}
Hunter, J.~D.
\newblock
  \href{https://ieeexplore.ieee.org/iel5/4160243/4160244/04160265.pdf?casa_token=z4juTDjMUu4AAAAA:4wX75cc11iYBafTwM2zz5VFZ5zkvtqZdtCefxxJrstv5ltdiEOREcMh_MxDUjXvxxEJxpVodi6kh}{Matplotlib:
  A 2D graphics environment}.
\newblock \emph{IEEE Annals of the History of Computing}, 9\penalty0
  (03):\penalty0 90--95, 2007.

\bibitem[Hutter(2004)]{hutter2004universal}
Hutter, M.
\newblock \emph{\href{http://www.hutter1.net/ai/suaibook.pdf}{Universal
  artificial intelligence: Sequential decisions based on algorithmic
  probability}}.
\newblock Springer Science \& Business Media, 2004.

\bibitem[Jaderberg et~al.(2016)Jaderberg, Mnih, Czarnecki, Schaul, Leibo,
  Silver, and Kavukcuoglu]{jaderberg2016reinforcement}
Jaderberg, M., Mnih, V., Czarnecki, W.~M., Schaul, T., Leibo, J.~Z., Silver,
  D., and Kavukcuoglu, K.
\newblock \href{https://arxiv.org/abs/1611.05397}{Reinforcement learning with
  unsupervised auxiliary tasks}.
\newblock \emph{arXiv preprint arXiv:1611.05397}, 2016.

\bibitem[Kalweit \& Boedecker(2017)Kalweit and
  Boedecker]{kalweit2017uncertainty}
Kalweit, G. and Boedecker, J.
\newblock
  \href{http://proceedings.mlr.press/v78/kalweit17a/kalweit17a.pdf}{Uncertainty-driven
  imagination for continuous deep reinforcement learning}.
\newblock In \emph{Conference on Robot Learning}, pp.\  195--206. PMLR, 2017.

\bibitem[Kenton et~al.(2019)Kenton, Filos, Evans, and
  Gal]{kenton2019generalizing}
Kenton, Z., Filos, A., Evans, O., and Gal, Y.
\newblock \href{https://arxiv.org/abs/1907.01475}{Generalizing from a few
  environments in safety-critical reinforcement learning}.
\newblock \emph{arXiv preprint arXiv:1907.01475}, 2019.

\bibitem[Kingma \& Ba(2014)Kingma and Ba]{kingma2014adam}
Kingma, D.~P. and Ba, J.
\newblock \href{https://arxiv.org/abs/1412.6980}{Adam: A method for stochastic
  optimization}.
\newblock \emph{arXiv preprint arXiv:1412.6980}, 2014.

\bibitem[Kumar \& Varaiya(2015)Kumar and Varaiya]{kumar2015stochastic}
Kumar, P.~R. and Varaiya, P.
\newblock
  \emph{\href{https://epubs.siam.org/doi/book/10.1137/1.9781611974263}{Stochastic
  systems: Estimation, identification, and adaptive control}}.
\newblock SIAM, 2015.

\bibitem[Kurutach et~al.(2018)Kurutach, Clavera, Duan, Tamar, and
  Abbeel]{kurutach2018model}
Kurutach, T., Clavera, I., Duan, Y., Tamar, A., and Abbeel, P.
\newblock \href{https://arxiv.org/abs/1802.10592}{Model-ensemble trust-region
  policy optimization}.
\newblock \emph{arXiv preprint arXiv:1802.10592}, 2018.

\bibitem[Lakshminarayanan et~al.(2016)Lakshminarayanan, Pritzel, and
  Blundell]{lakshminarayanan2016simple}
Lakshminarayanan, B., Pritzel, A., and Blundell, C.
\newblock \href{https://arxiv.org/abs/1612.01474}{Simple and scalable
  predictive uncertainty estimation using deep ensembles}.
\newblock \emph{arXiv preprint arXiv:1612.01474}, 2016.

\bibitem[Lee et~al.(2019)Lee, Nagabandi, Abbeel, and Levine]{lee2019stochastic}
Lee, A.~X., Nagabandi, A., Abbeel, P., and Levine, S.
\newblock \href{https://arxiv.org/abs/1907.00953}{Stochastic latent
  actor-critic: Deep reinforcement learning with a latent variable model}.
\newblock \emph{arXiv preprint arXiv:1907.00953}, 2019.

\bibitem[Lopes et~al.(2012)Lopes, Lang, Toussaint, and
  Oudeyer]{lopes2012exploration}
Lopes, M., Lang, T., Toussaint, M., and Oudeyer, P.-Y.
\newblock
  \href{https://hal.inria.fr/file/index/docid/755248/filename/nips.pdf}{Exploration
  in model-based reinforcement learning by empirically estimating learning
  progress}.
\newblock In \emph{Neural Information Processing Systems (NIPS)}, 2012.

\bibitem[Loshchilov \& Hutter(2017)Loshchilov and
  Hutter]{loshchilov2017decoupled}
Loshchilov, I. and Hutter, F.
\newblock \href{https://arxiv.org/abs/1711.05101}{Decoupled weight decay
  regularization}.
\newblock \emph{arXiv preprint arXiv:1711.05101}, 2017.

\bibitem[Lowrey et~al.(2018)Lowrey, Rajeswaran, Kakade, Todorov, and
  Mordatch]{lowrey2018plan}
Lowrey, K., Rajeswaran, A., Kakade, S., Todorov, E., and Mordatch, I.
\newblock \href{https://arxiv.org/abs/1811.01848}{Plan online, learn offline:
  Efficient learning and exploration via model-based control}.
\newblock \emph{arXiv preprint arXiv:1811.01848}, 2018.

\bibitem[Lu et~al.(2021)Lu, Van~Roy, Dwaracherla, Ibrahimi, Osband, and
  Wen]{lu2021reinforcement}
Lu, X., Van~Roy, B., Dwaracherla, V., Ibrahimi, M., Osband, I., and Wen, Z.
\newblock \href{https://arxiv.org/abs/2103.04047}{Reinforcement Learning, Bit
  by Bit}.
\newblock \emph{arXiv preprint arXiv:2103.04047}, 2021.

\bibitem[Lyle et~al.(2021)Lyle, Rowland, Ostrovski, and Dabney]{lyle2021effect}
Lyle, C., Rowland, M., Ostrovski, G., and Dabney, W.
\newblock \href{http://proceedings.mlr.press/v130/lyle21a/lyle21a.pdf}{On The
  Effect of Auxiliary Tasks on Representation Dynamics}.
\newblock In \emph{International Conference on Artificial Intelligence and
  Statistics}, pp.\  1--9. PMLR, 2021.

\bibitem[Ma et~al.(2020)Ma, Chen, Hsu, and Lee]{ma2020contrastive}
Ma, X., Chen, S., Hsu, D., and Lee, W.~S.
\newblock \href{https://arxiv.org/abs/2008.02430}{Contrastive variational
  model-based reinforcement learning for complex observations}.
\newblock \emph{arXiv e-prints}, pp.\  arXiv--2008, 2020.

\bibitem[Maddox et~al.(2019)Maddox, Izmailov, Garipov, Vetrov, and
  Wilson]{maddox2019simple}
Maddox, W.~J., Izmailov, P., Garipov, T., Vetrov, D.~P., and Wilson, A.~G.
\newblock
  \href{https://proceedings.neurips.cc/paper/2019/file/118921efba23fc329e6560b27861f0c2-Paper.pdf}{A
  simple baseline for bayesian uncertainty in deep learning}.
\newblock \emph{Advances in Neural Information Processing Systems},
  32:\penalty0 13153--13164, 2019.

\bibitem[Milnor(1951)]{milnor1951games}
Milnor, J.
\newblock \href{https://apps.dtic.mil/sti/pdfs/ADA596133.pdf}{Games against
  nature}.
\newblock Technical report, RAND Project Air Force Santa Monica CA, 1951.

\bibitem[Mnih et~al.(2013)Mnih, Kavukcuoglu, Silver, Graves, Antonoglou,
  Wierstra, and Riedmiller]{mnih2013playing}
Mnih, V., Kavukcuoglu, K., Silver, D., Graves, A., Antonoglou, I., Wierstra,
  D., and Riedmiller, M.
\newblock \href{https://arxiv.org/abs/1312.5602}{Playing atari with deep
  reinforcement learning}.
\newblock \emph{arXiv preprint arXiv:1312.5602}, 2013.

\bibitem[Munos et~al.(2016)Munos, Stepleton, Harutyunyan, and
  Bellemare]{munos2016safe}
Munos, R., Stepleton, T., Harutyunyan, A., and Bellemare, M.~G.
\newblock \href{https://arxiv.org/abs/1606.02647}{Safe and efficient off-policy
  reinforcement learning}.
\newblock \emph{arXiv preprint arXiv:1606.02647}, 2016.

\bibitem[Nikishin et~al.(2021)Nikishin, Abachi, Agarwal, and
  Bacon]{nikishin2021control}
Nikishin, E., Abachi, R., Agarwal, R., and Bacon, P.-L.
\newblock \href{https://arxiv.org/abs/2106.03273}{Control-Oriented Model-Based
  Reinforcement Learning with Implicit Differentiation}.
\newblock \emph{arXiv preprint arXiv:2106.03273}, 2021.

\bibitem[Oh et~al.(2017)Oh, Singh, and Lee]{oh2017value}
Oh, J., Singh, S., and Lee, H.
\newblock \href{https://arxiv.org/abs/1707.03497}{Value prediction network}.
\newblock \emph{arXiv preprint arXiv:1707.03497}, 2017.

\bibitem[Osband et~al.(2016)Osband, Blundell, Pritzel, and
  Van~Roy]{osband2016deep}
Osband, I., Blundell, C., Pritzel, A., and Van~Roy, B.
\newblock \href{https://arxiv.org/abs/1602.04621}{Deep exploration via
  bootstrapped DQN}.
\newblock \emph{Advances in neural information processing systems},
  29:\penalty0 4026--4034, 2016.

\bibitem[Osband et~al.(2018)Osband, Aslanides, and
  Cassirer]{osband2018randomized}
Osband, I., Aslanides, J., and Cassirer, A.
\newblock \href{https://arxiv.org/abs/1806.03335}{Randomized prior functions
  for deep reinforcement learning}.
\newblock \emph{arXiv preprint arXiv:1806.03335}, 2018.

\bibitem[Pathak et~al.(2017)Pathak, Agrawal, Efros, and
  Darrell]{pathak2017curiosity}
Pathak, D., Agrawal, P., Efros, A.~A., and Darrell, T.
\newblock \href{https://arxiv.org/abs/1705.05363}{Curiosity-driven exploration
  by self-supervised prediction}.
\newblock In \emph{International conference on machine learning}, pp.\
  2778--2787. PMLR, 2017.

\bibitem[Pathak et~al.(2019)Pathak, Gandhi, and Gupta]{pathak2019self}
Pathak, D., Gandhi, D., and Gupta, A.
\newblock \href{http://arxiv.org/abs/1906.04161}{Self-supervised exploration
  via disagreement}.
\newblock In \emph{International conference on machine learning}, pp.\
  5062--5071. PMLR, 2019.

\bibitem[Pearce et~al.(2020)Pearce, Leibfried, and
  Brintrup]{pearce2020uncertainty}
Pearce, T., Leibfried, F., and Brintrup, A.
\newblock
  \href{http://proceedings.mlr.press/v108/pearce20a/pearce20a.pdf}{Uncertainty
  in neural networks: Approximately bayesian ensembling}.
\newblock In \emph{International conference on artificial intelligence and
  statistics}, pp.\  234--244. PMLR, 2020.

\bibitem[Puterman(2014)]{puterman2014markov}
Puterman, M.~L.
\newblock
  \emph{\href{https://onlinelibrary.wiley.com/doi/book/10.1002/9780470316887}{Markov
  decision processes: discrete stochastic dynamic programming}}.
\newblock John Wiley \& Sons, 2014.

\bibitem[Raileanu \& Rockt{\"a}schel(2020)Raileanu and
  Rockt{\"a}schel]{raileanu2020ride}
Raileanu, R. and Rockt{\"a}schel, T.
\newblock \href{https://arxiv.org/abs/2002.12292}{RIDE: Rewarding impact-driven
  exploration for procedurally-generated environments}.
\newblock \emph{arXiv preprint arXiv:2002.12292}, 2020.

\bibitem[Richalet et~al.(1978)Richalet, Rault, Testud, and
  Papon]{richalet1978model}
Richalet, J., Rault, A., Testud, J., and Papon, J.
\newblock
  \href{https://www.sciencedirect.com/science/article/abs/pii/0005109878900018}{Model
  predictive heuristic control}.
\newblock \emph{Automatica (journal of IFAC)}, 14\penalty0 (5):\penalty0
  413--428, 1978.

\bibitem[Savage(1972)]{savage1972foundations}
Savage, L.~J.
\newblock \emph{\href{https://philpapers.org/rec/SAVTFO-2}{The foundations of
  statistics}}.
\newblock Courier Corporation, 1972.

\bibitem[Schmidhuber(1990)]{schmidhuber1990line}
Schmidhuber, J.
\newblock \href{https://mediatum.ub.tum.de/doc/814960/file.pdf}{An on-line
  algorithm for dynamic reinforcement learning and planning in reactive
  environments}.
\newblock In \emph{1990 IJCNN international joint conference on neural
  networks}, pp.\  253--258. IEEE, 1990.

\bibitem[Schmidhuber(2010)]{schmidhuber2010formal}
Schmidhuber, J.
\newblock \href{https://people.idsia.ch/~juergen/ieeecreative.pdf}{Formal
  theory of creativity, fun, and intrinsic motivation (1990--2010)}.
\newblock \emph{IEEE Transactions on Autonomous Mental Development}, 2\penalty0
  (3):\penalty0 230--247, 2010.

\bibitem[Schmitt et~al.(2020)Schmitt, Hessel, and Simonyan]{schmitt2020off}
Schmitt, S., Hessel, M., and Simonyan, K.
\newblock
  \href{http://proceedings.mlr.press/v119/schmitt20a/schmitt20a.pdf}{Off-policy
  actor-critic with shared experience replay}.
\newblock In \emph{International Conference on Machine Learning}, pp.\
  8545--8554. PMLR, 2020.

\bibitem[Schrittwieser et~al.(2020)Schrittwieser, Antonoglou, Hubert, Simonyan,
  Sifre, Schmitt, Guez, Lockhart, Hassabis, Graepel,
  et~al.]{schrittwieser2020mastering}
Schrittwieser, J., Antonoglou, I., Hubert, T., Simonyan, K., Sifre, L.,
  Schmitt, S., Guez, A., Lockhart, E., Hassabis, D., Graepel, T., et~al.
\newblock \href{https://www.nature.com/articles/s41586-020-03051-4}{Mastering
  atari, go, chess and shogi by planning with a learned model}.
\newblock \emph{Nature}, 588\penalty0 (7839):\penalty0 604--609, 2020.

\bibitem[Sekar et~al.(2020)Sekar, Rybkin, Daniilidis, Abbeel, Hafner, and
  Pathak]{sekar2020planning}
Sekar, R., Rybkin, O., Daniilidis, K., Abbeel, P., Hafner, D., and Pathak, D.
\newblock \href{https://arxiv.org/abs/2005.05960}{Planning to explore via
  self-supervised world models}.
\newblock In \emph{International Conference on Machine Learning}, pp.\
  8583--8592. PMLR, 2020.

\bibitem[Shyam et~al.(2019)Shyam, Ja{\'s}kowski, and Gomez]{shyam2019model}
Shyam, P., Ja{\'s}kowski, W., and Gomez, F.
\newblock \href{https://arxiv.org/abs/1810.12162}{Model-based active
  exploration}.
\newblock In \emph{International conference on machine learning}, pp.\
  5779--5788. PMLR, 2019.

\bibitem[Silver et~al.(2017)Silver, Hasselt, Hessel, Schaul, Guez, Harley,
  Dulac-Arnold, Reichert, Rabinowitz, Barreto, et~al.]{silver2017predictron}
Silver, D., Hasselt, H., Hessel, M., Schaul, T., Guez, A., Harley, T.,
  Dulac-Arnold, G., Reichert, D., Rabinowitz, N., Barreto, A., et~al.
\newblock \href{https://arxiv.org/abs/1612.08810}{The predictron: End-to-end
  learning and planning}.
\newblock In \emph{International Conference on Machine Learning}, pp.\
  3191--3199. PMLR, 2017.

\bibitem[Stadie et~al.(2015)Stadie, Levine, and
  Abbeel]{stadie2015incentivizing}
Stadie, B.~C., Levine, S., and Abbeel, P.
\newblock \href{https://arxiv.org/abs/1507.00814}{Incentivizing exploration in
  reinforcement learning with deep predictive models}.
\newblock \emph{arXiv preprint arXiv:1507.00814}, 2015.

\bibitem[Strens(2000)]{strens2000bayesian}
Strens, M.
\newblock
  \href{https://www.ece.uvic.ca/~bctill/papers/learning/Strens_2000.pdf}{A
  Bayesian framework for reinforcement learning}.
\newblock In \emph{ICML}, volume 2000, pp.\  943--950, 2000.

\bibitem[Sutton(1988)]{sutton1988learning}
Sutton, R.~S.
\newblock
  \href{https://link.springer.com/content/pdf/10.1007/BF00115009.pdf}{Learning
  to predict by the methods of temporal differences}.
\newblock \emph{Machine learning}, 3\penalty0 (1):\penalty0 9--44, 1988.

\bibitem[Sutton(1991)]{sutton1991dyna}
Sutton, R.~S.
\newblock
  \href{https://citeseerx.ist.psu.edu/viewdoc/download?doi=10.1.1.48.6005&rep=rep1&type=pdf}{Dyna,
  an integrated architecture for learning, planning, and reacting}.
\newblock \emph{ACM Sigart Bulletin}, 2\penalty0 (4):\penalty0 160--163, 1991.

\bibitem[Sutton \& Barto(2018)Sutton and Barto]{sutton2018reinforcement}
Sutton, R.~S. and Barto, A.~G.
\newblock
  \emph{\href{http://incompleteideas.net/book/the-book-2nd.html}{Reinforcement
  learning: An introduction}}.
\newblock MIT press, 2018.

\bibitem[Tibshirani(1996)]{tibshirani1996comparison}
Tibshirani, R.
\newblock
  \href{https://citeseerx.ist.psu.edu/viewdoc/download?doi=10.1.1.36.1022&rep=rep1&type=pdf}{A
  comparison of some error estimates for neural network models}.
\newblock \emph{Neural Computation}, 8\penalty0 (1):\penalty0 152--163, 1996.

\bibitem[Tunyasuvunakool et~al.(2020)Tunyasuvunakool, Muldal, Doron, Liu,
  Bohez, Merel, Erez, Lillicrap, Heess, and
  Tassa]{tunyasuvunakool2020dm_control}
Tunyasuvunakool, S., Muldal, A., Doron, Y., Liu, S., Bohez, S., Merel, J.,
  Erez, T., Lillicrap, T., Heess, N., and Tassa, Y.
\newblock \href{https://arxiv.org/abs/2006.12983}{dm\_control: Software and
  tasks for continuous control}.
\newblock \emph{Software Impacts}, 6:\penalty0 100022, 2020.

\bibitem[Van~Rossum \& Drake~Jr(1995)Van~Rossum and Drake~Jr]{van1995python}
Van~Rossum, G. and Drake~Jr, F.~L.
\newblock
  \emph{\href{http://www.cs.cmu.edu/afs/cs.cmu.edu/project/gwydion-1/OldFiles/OldFiles/python/Doc/ref.ps}{Python
  reference manual}}.
\newblock Centrum voor Wiskunde en Informatica Amsterdam, 1995.

\bibitem[Van~Seijen et~al.(2009)Van~Seijen, Van~Hasselt, Whiteson, and
  Wiering]{van2009theoretical}
Van~Seijen, H., Van~Hasselt, H., Whiteson, S., and Wiering, M.
\newblock \href{https://ieeexplore.ieee.org/document/4927542}{A theoretical and
  empirical analysis of Expected Sarsa}.
\newblock In \emph{2009 ieee symposium on adaptive dynamic programming and
  reinforcement learning}, pp.\  177--184. IEEE, 2009.

\bibitem[Watkins \& Dayan(1992)Watkins and Dayan]{watkins1992q}
Watkins, C.~J. and Dayan, P.
\newblock
  \href{https://link.springer.com/article/10.1007/BF00992698}{Q-learning}.
\newblock \emph{Machine learning}, 8\penalty0 (3-4):\penalty0 279--292, 1992.

\bibitem[Watter et~al.(2015)Watter, Springenberg, Boedecker, and
  Riedmiller]{watter2015embed}
Watter, M., Springenberg, J.~T., Boedecker, J., and Riedmiller, M.
\newblock \href{https://arxiv.org/abs/1506.07365}{Embed to control: A locally
  linear latent dynamics model for control from raw images}.
\newblock \emph{arXiv preprint arXiv:1506.07365}, 2015.

\bibitem[Wenzel et~al.(2020)Wenzel, Snoek, Tran, and
  Jenatton]{wenzel2020hyperparameter}
Wenzel, F., Snoek, J., Tran, D., and Jenatton, R.
\newblock \href{https://arxiv.org/abs/2006.13570}{Hyperparameter ensembles for
  robustness and uncertainty quantification}.
\newblock \emph{arXiv preprint arXiv:2006.13570}, 2020.

\bibitem[Werbos(1987)]{werbos1987learning}
Werbos, P.~J.
\newblock
  \href{https://bibbase.org/network/publication/werbos-learninghowtheworldworksspecificationsforpredictivenetworksinrobotsandbrains-1987}{Learning
  how the world works: Specifications for predictive networks in robots and
  brains}.
\newblock In \emph{Proceedings of IEEE International Conference on Systems, Man
  and Cybernetics, NY}, 1987.

\bibitem[Wichard et~al.(2003)Wichard, Merkwirth, and
  Ogorzalek]{wichard2003building}
Wichard, J., Merkwirth, C., and Ogorzalek, M.
\newblock
  \href{http://www.j-wichard.de/publications/salerno_lncs_2003.pdf}{Building
  ensembles with heterogeneous models}, 2003.

\bibitem[Wilson \& Izmailov(2020)Wilson and Izmailov]{wilson2020bayesian}
Wilson, A.~G. and Izmailov, P.
\newblock \href{https://arxiv.org/abs/2002.08791}{Bayesian deep learning and a
  probabilistic perspective of generalization}.
\newblock \emph{arXiv preprint arXiv:2002.08791}, 2020.

\bibitem[Young \& Tian(2019)Young and Tian]{young2019minatar}
Young, K. and Tian, T.
\newblock \href{https://arxiv.org/abs/1903.03176}{Minatar: An atari-inspired
  testbed for thorough and reproducible reinforcement learning experiments}.
\newblock \emph{arXiv preprint arXiv:1903.03176}, 2019.

\bibitem[Yu et~al.(2021)Yu, Lan, Zeng, Feng, Zhang, and
  Chen]{yu2021playvirtual}
Yu, T., Lan, C., Zeng, W., Feng, M., Zhang, Z., and Chen, Z.
\newblock \href{https://arxiv.org/abs/2106.04152}{Playvirtual: Augmenting
  cycle-consistent virtual trajectories for reinforcement learning}.
\newblock \emph{Advances in Neural Information Processing Systems}, 34, 2021.

\bibitem[Zhu et~al.(2017)Zhu, Park, Isola, and Efros]{zhu2017unpaired}
Zhu, J.-Y., Park, T., Isola, P., and Efros, A.~A.
\newblock \href{https://arxiv.org/abs/1703.10593}{Unpaired image-to-image
  translation using cycle-consistent adversarial networks}.
\newblock In \emph{Proceedings of the IEEE international conference on computer
  vision}, pp.\  2223--2232, 2017.

\end{thebibliography}
